\documentclass[10pt,twocolumn,letterpaper]{article}

\usepackage{cvpr}              % To produce the CAMERA-READY version
\usepackage{graphicx}
\usepackage{booktabs}
\usepackage{bm}
\usepackage{wrapfig}
\usepackage{algorithm}
\usepackage{algorithmic}
\usepackage{eurosym}
\usepackage{multirow}
\usepackage{colortbl}
\usepackage{tikz}
\usepackage[accsupp]{axessibility}  % Improves PDF readability for those with disabilities.

\definecolor{cvprblue}{rgb}{0.21,0.49,0.74}
\usepackage[pagebackref,breaklinks,colorlinks,allcolors=cvprblue]{hyperref}

\def\paperID{2732} % *** Enter the Paper ID here
\def\confName{CVPR}
\def\confYear{2025}

\title{\textsc{Helvipad}: A Real-World Dataset for Omnidirectional Stereo Depth Estimation}

\author{
  Mehdi Zayene$^1$ \quad Jannik Endres$^{1,2}$ \quad Albias Havolli$^1$ \quad Charles Corbière$^1$\thanks{Project lead, contact: \href{mailto:charles.corbiere@gmail.com}{charles.corbiere@gmail.com}.} \\ 
  \quad Salim Cherkaoui$^1$ \quad Alexandre Kontouli$^1$ \quad Alexandre Alahi$^1$ \\
  {\normalsize $^1$EPFL, Switzerland \quad $^2$TU Darmstadt, Germany} \\
  {\normalsize \url{https://vita-epfl.github.io/Helvipad/}}
}

\begin{document}

\makeatletter
\g@addto@macro\@maketitle{
  \vspace{-1.0cm}
  \begin{figure}[H]
  \setlength{\linewidth}{\textwidth}
  \setlength{\hsize}{\textwidth}
  \centering
  \includegraphics[width=\linewidth]{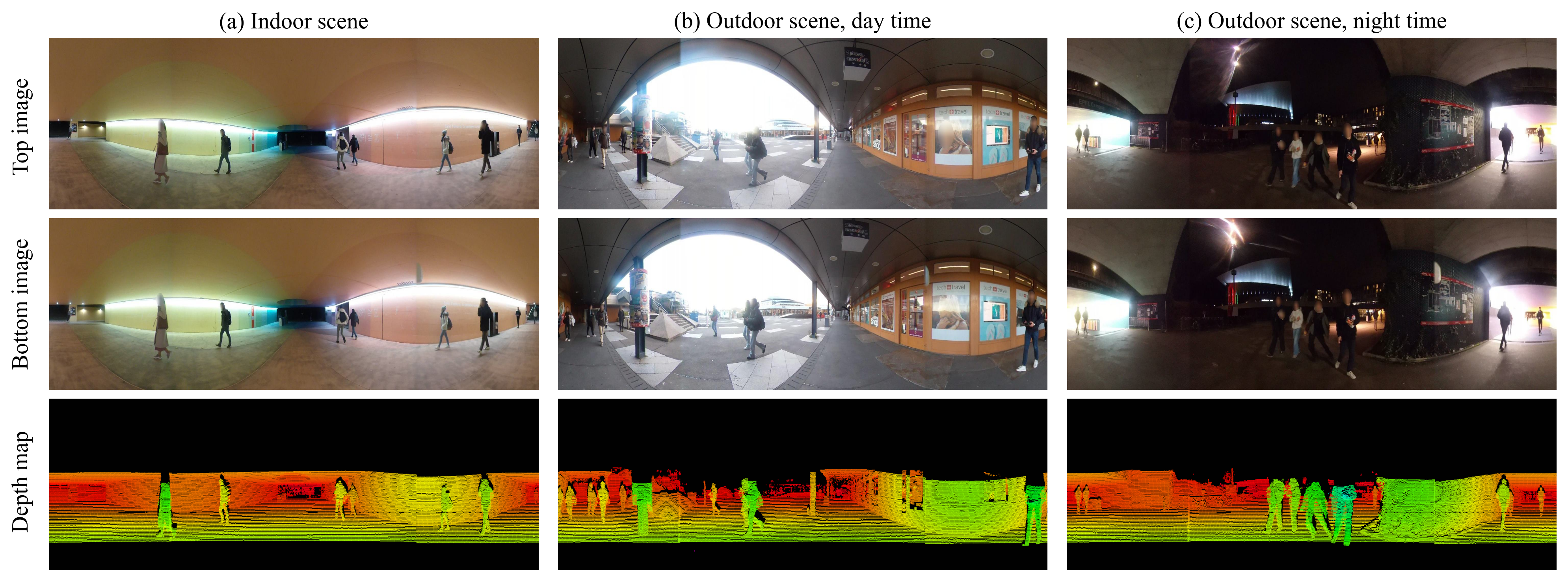}
  \caption{\textbf{Examples from the \textsc{Helvipad} dataset} in diverse settings (indoor, outdoor), under varied lighting (day, night). The dataset consists of paired top-bottom images with corresponding depth maps, captured from a robot navigating in dynamic human environments.}
  \label{fig:examples_Helvipad}
  \vspace{0.3cm}
  \end{figure}
}
\makeatother
\maketitle

\begin{abstract}
Despite progress in stereo depth estimation, omnidirectional imaging remains underexplored, mainly due to the lack of appropriate data. We introduce \textsc{Helvipad}, a real-world dataset for omnidirectional stereo depth estimation, featuring 40K video frames from video sequences across diverse environments, including crowded indoor and outdoor scenes with various lighting conditions. Collected using two 360° cameras in a top-bottom setup and a LiDAR sensor, the dataset includes accurate depth and disparity labels by projecting 3D point clouds onto equirectangular images. Additionally, we provide an augmented training set with an increased label density by using depth completion. We benchmark leading stereo depth estimation models for both standard and omnidirectional images. The results show that while recent stereo methods perform decently, a challenge persists in accurately estimating depth in omnidirectional imaging. To address this, we introduce necessary adaptations to stereo models, leading to improved performance. \vspace{-1.3cm}
\end{abstract}

%=================================================================================%
\section{Introduction}
\label{sec:intro}
%=================================================================================%

Mobile robots are increasingly used in real-world applications such as autonomous driving~\citep{janai2017computer}, healthcare~\citep{technologies9010008}, or agriculture~\citep{Droukas_2023}, where accurate 3D representations of dynamic human environments are crucial for navigation and interaction. Depth estimation~\citep{xu2023iterative,Geiger2012CVPR,depthanything2024,wang20icra} aims to help build these representations by estimating the distance of surrounding objects from the robot's viewpoint.

Historically, LiDAR sensors have been the gold standard for capturing high-quality depth information due to their precision and their 360° coverage. They come, however, with significant limitations~\citep{drivingstereo}, such as providing point clouds that are sparse, especially for objects in far range, and with high deployment costs. This has driven growing interest in estimating depth from more accessible imaging devices, such as stereo cameras. In recent years, deep learning approaches~\citep{zbontar2016stereo,kendall2017end,xu2023iterative} and the release of larger and more challenging benchmarks~\citep{drivingstereo,Apolloscape} have greatly advanced the field of depth estimation. Yet, existing datasets collected with stereo systems typically offer a limited field of view, capturing frontal images or panoramic views that do not fully encompass the surrounding environment.

With the increasing availability of consumer-grade 360° cameras, omnidirectional imaging~\citep{Ai2022DeepLF,Won2019SweepNetWO,wang20icra,won2020end} has gained traction in computer vision. By offering a complete field of view, rich geometric information, and multiple projection types (e.g., equirectangular, cubemap), it provides an attractive option for applications that require a comprehensive spatial awareness, such as robotic navigation and exploration in complex indoor environments~\citep{JRDB}. Despite its potential, the application of deep learning to omnidirectional depth estimation has been hindered by two obstacles: (1) the scarcity of real-world datasets from omnidirectional cameras with pixel-wise depth labels, and (2) the spherical geometry of omnidirectional images introducing significant distortions, which complicates the application of conventional stereo models designed for rectilinear images.

We introduce \textsc{Helvipad}, a comprehensive real-world dataset for stereo depth estimation from 360° images. It consists of 29 video sequences captured in dynamic indoor and outdoor scenes across a university campus (\cref{fig:examples_Helvipad}), under diverse weather and lighting conditions. Data were collected using a custom-built rig equipped with two omnidirectional cameras in a vertical arrangement, which prevents occlusion, and paired with a synchronized LiDAR sensor to provide high-quality depth measurements. Applying depth completion, we augment training data with a significantly increased label density, which proves to be an effective data augmentation technique for multiple models in our experiment. With a total of 39,553 labeled frames, \textsc{Helvipad} serves as a foundational resource for developing and benchmarking omnidirectional depth estimation models capable of navigating in human environments.

When evaluating both standard and omnidirectional depth estimation methods with \textsc{Helvipad}, our results show that recent stereo matching models outperform omnidirectional approaches based on older architectures, but they struggle due to the severe distortions of the equirectangular projection. To cope with the spherical geometry of omnidirectional images, we propose to enhance stereo matching models by incorporating a polar angle map as input, and we make use of the 360° view by applying circular padding during inference. These adaptations, when integrated into a recent state-of-the-art model, lead to performance gains that surpass all prior approaches on our dataset. Additionally, we conduct a detailed scene-wise analysis to study each method's ability to generalize to unseen or underrepresented situations, such as outdoor scenes at night. 

\noindent
Our main contributions are as follows:
\begin{itemize}
    \item We construct \textsc{Helvipad}, a real-world omnidirectional stereo dataset, featuring 40K frames from indoor and outdoor video sequences collected in various conditions, along with high-quality depth and disparity labels, and including further augmented data via depth completion which increases label density for training; 
    \item We benchmark modern stereo depth estimation approaches on our dataset, providing also a detailed analysis by type of scene and evaluating the impact of training on augmented data;
    \item We propose and evaluate key adaptations for stereo matching models to better handle spherical image geometry, resulting in performance gains.
\end{itemize}

%=================================================================================%
\section{Related Work}
\label{sec:related_work}
%=================================================================================%

\begin{table*}[t]
    \centering
    \setlength{\tabcolsep}{6pt} 
    \renewcommand{\arraystretch}{0.9}
    \small % Reduce font size
    \resizebox{\linewidth}{!}{%
    \begin{tabular}{lccccccrr}
        \toprule
        \textbf{Dataset} & \textbf{Real} & \textbf{Images} & \textbf{Scenes}   & \textbf{Depth} & \textbf{Night} & \textbf{Humans} & \textbf{Size} & \textbf{Resolution} \\
        \midrule
        SceneFlow~\citep{Sceneflow}     &  & stereo      & outdoor           & \checkmark &  &  & 39K  & 960 x 540   \\
        KITTI~\citep{Geiger2012CVPR,Menze2015CVPR}         & \checkmark & stereo      & driving           & \checkmark &  & $\bm{+}$ & 400  & 1242 x 375  \\
        ApolloScape~\citep{Apolloscape}   & \checkmark & stereo      & driving           & \checkmark & \checkmark & $\bm{+}$ & 140K & 3384 × 2710 \\
        DrivingStereo~\citep{drivingstereo} & \checkmark & stereo      & driving           & \checkmark &  & $\bm{+}$ & 181K & 1762 x 1080 \\
        \midrule
        nuScenes~\citep{nuscenes}      & \checkmark & multi-view  & driving           &  & \checkmark & $\bm{+}$ & 1.4M & 1600 x 1200 \\
        Waymo~\citep{waymo}         & \checkmark & multi-view  & driving           &  & \checkmark & $\bm{+}$ & 1M   & 1920 x 1080 \\
        OmniHouse~\citep{won2020end} &  & multi-view & indoor & \checkmark & & & 10K & 768 x 800 \\
        OmniThings~\citep{won2020end} &  & multi-view & indoor & \checkmark & & & 10K & 768 x 800 \\
        \midrule
        Stereo-MP3D~\citep{Matterport3D, wang20icra}          &  & 360° stereo & indoor            & \checkmark &  &  & 2K   & 1024 x 512  \\
        Stereo-SF3D~\citep{SF3D, wang20icra}          &  & 360° stereo & indoor            & \checkmark &  &  & 1K   & 1024 x 512  \\
        Pano3D~\citep{Albanis_2021_CVPR} & & 360° mono  & indoor & \checkmark & & & 21K & 1014 x 512 \\
        JRDB~\citep{JRDB}          & \checkmark & 360° stereo & outdoor, indoor &  & \checkmark & $\bm{++}$ & 28K  & 3760 x 480  \\
        \midrule
        \rowcolor{gray!25} \textbf{Helvipad} (Ours)  & \checkmark & 360° stereo   & outdoor, indoor & \checkmark             & \checkmark            & $\bm{++}$         & 40K          & 1920 x 512         \\
        \bottomrule
    \end{tabular}%
    }
    \caption{\textbf{Comparison between \textsc{Helvipad} and popular datasets for stereo depth estimation}. \textsc{Helvipad} is the first real-world stereo dataset for omnidirectional images with pixel-wise labels collected in indoor and outdoor scenes under varying lighting conditions.}
    \label{tab:dataset_comparison}
\end{table*}

\paragraph{Stereo datasets for depth estimation.}
Following early contributions from synthetic datasets like SceneFlow~\citep{NYUv2} and real-world benchmarks like KITTI~\citep{Geiger2012CVPR,Menze2015CVPR}, large-scale driving datasets, such as Apolloscape~\citep{Apolloscape}, nuScenes~\citep{nuscenes}, and Waymo~\citep{waymo} expanded to more complex scenarios, but often lack pixel-wise depth labels or are limited to frontal or multi-view configurations (see \cref{tab:dataset_comparison}). In omnidirectional imaging, public datasets with 360° images have enabled deep learning methods to improve performance in tasks like 2D/3D object detection~\citep{SF3D,JRDB, zheng2020structured3d}, semantic segmentation~\citep{SF3D, zhang2019orientation}, tracking~\citep{JRDB}, and monocular depth estimation~\citep{song2016ssc}). OmniHouse and OmniThings~\citep{won2020end} provide synthetic 360° indoor scenes captured using a cross-configuration of four fisheye cameras. Although this setup offers a wide field of view, it is more challenging and costly to implement than the simpler top-bottom setup of two 360° cameras in \textsc{Helvipad}. MP3D~\citep{Matterport3D} and SF3D~\citep{SF3D} offer 360° images but are limited to synthetic indoor scenes with single view. \citet{wang20icra} built two stereo datasets with top-bottom setup based on these two synthetic databases. The JRDB dataset~\citep{JRDB} spans a variety of tasks in real-world outdoor and indoor environments, but lacks pixel-wise depth annotations. In this context, \textsc{Helvipad} emerges as a comprehensive real-world stereo dataset of 360° images, covering indoor and outdoors scenes with varying lighting conditions and pixel-wise annotations.

\paragraph{Stereo matching.} Deep learning has increasingly dominated stereo matching since \citet{zbontar2016stereo} introduced the use of convolutional neural networks to describe image patches for stereo matching. Subsequent approaches have incorporated 3D convolutional kernels to regularize 4D cost volumes~\citep{kendall2017end,guo2019group} and enhanced architectures with spatial pyramidal pooling, like PSMNet~\citep{chang2018pyramid}. 
Recently, iterative optimization-based methods~\citep{raftstereo,xu2023iterative, zeng2024temporally} refine disparity fields recurrently using local cost values.
For instance, IGEV-Stereo~\citep{xu2023iterative} refines disparity maps iteratively using ConvGRUs \citep{cho2014properties} and constructs a combined geometrical encoding volume that encodes both non-local geometry and fine-grained local matching details.
Other methods use cross-view completion for self-supervised pre-training~\citep{weinzaepfel2023croco}, a Markov Random Field with neural networks for potential functions and message passing~\citep{guan2024neural}, or an adapter to a pre-trained vision transformer for feature extraction~\citep{li2024roadformer}.

\paragraph{Deep learning for omnidirectional depth estimation.} Unlike conventional stereo matching, research in omnidirectional stereo matching remains limited. Early models like OmniMVS~\citep{won2020end} and SweepNet~\citep{Won2019SweepNetWO} introduced wide-baseline omnidirectional setups with fisheye images from multiple cameras, with SweepNet using spherical sweeping~\citep{Meuleman_2021_CVPR} to generate dense cost volumes. Recent works~\citep{reyarea2021360monodepth,shen2022panoformer} address spherical geometry distortions of panoramic images, while circular padding~\citep{cylinpainting2023,kailun2020} has been used to better maintain edge continuity in predictions. Unsupervised OmniMVS~\citep{Chen2023UnsupervisedOE} leverages photometric consistency constraints to alleviate the need of omnidirectional stereo data, but may face challenges under low-light conditions and with non-Lambertian surfaces. Being the only approach to tackle omnidirectional stereo matching from top-bottom camera setup, 360SD-Net~\citep{wang20icra} is an end-to-end deep learning method that mitigates distortions in equirectangular images thanks to a polar angle coordinate input and a learnable cost volume. But its architecture with spatial pyramid pooling does not integrate more recent advances in stereo matching presented in the previous paragraph. In this paper, we enhance a recent leading stereo matching models with adaptations to better handle 360° images.

%====================================================================================%
\section{The \textsc{Helvipad} Dataset}
\label{sec:dataset}
%====================================================================================%

This section details the collection process of the dataset (\cref{subsec:data_acquisition}). We address the challenge of producing accurate depth labels by projecting LiDAR point clouds onto equirectangular images (\cref{subsec:labelling}) and applying depth completion techniques to increase label density for an augmented training set (\cref{subsec:depth_completion}). Finally, we present data split statistics and depth information for the dataset (\cref{subsec:statistics}).

%=--------------------------------------------------------------------------------%
\subsection{Data Acquisition}
\label{subsec:data_acquisition}
%=--------------------------------------------------------------------------------%

To assemble a dataset capturing human pedestrian activity, we built a custom rig and collected data across distinct scenes, encompassing indoor (\textit{e.g.}, corridors, halls) and outdoor (\textit{e.g.},  squares, external parking, footpaths) settings in various lighting conditions on a university campus. The rig features two Ricoh Theta V 360° cameras in a top-bottom setup with a 19.1 cm baseline, capturing equirectangular images at 30 fps, paired with a Ouster OS1-64 LiDAR sensor with a vertical field of view of 42.4°, operating at 10 fps, and mounted 45.0 cm below the bottom camera. A central processor manages data capture and ensures device synchronization.
In total, we collected 29 video sequences. More details on data collection are available in \cref{appx_sec:details_dataset}.

\begin{figure}[t]
  \centering
    \includegraphics[width=\linewidth]{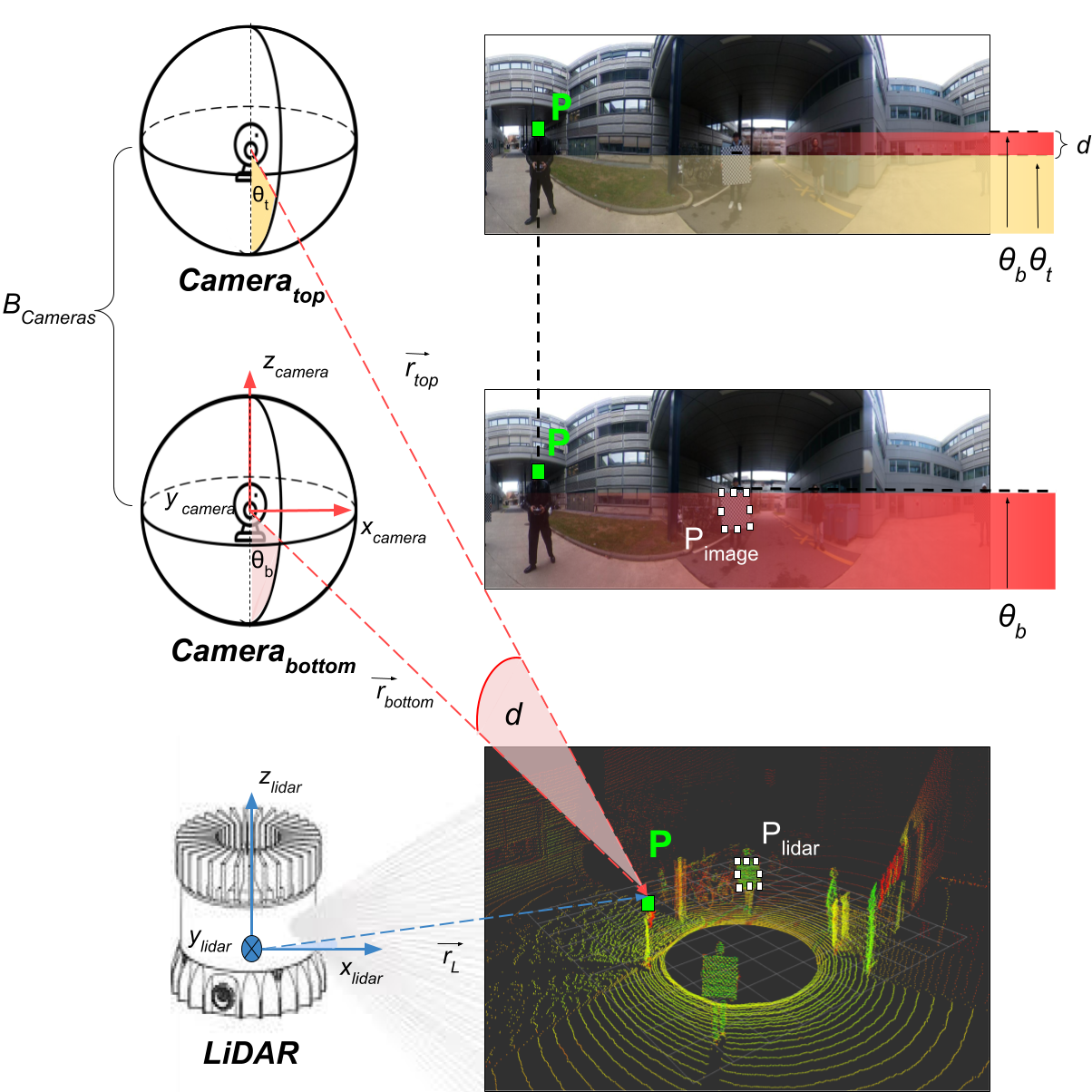}
    \caption{\textbf{LiDAR to 360° image mapping and spherical disparity.} 
    The top and bottom cameras, separated by a baseline \( B_{\text{cameras}} \), capture a shared projection point \( P \), mapped to both image and LiDAR coordinates. Depth vectors \(\vec{r}_{\text{top}}\) and \(\vec{r}_{\text{bottom}}\) represent distances in each coordinate frame.
    The polar angles \(\theta_b\) and \(\theta_t\) represent the angles from the bottom and top cameras to \( P \), respectively, while the angular disparity \(d\) quantifies the angular difference between corresponding points in the two camera views. \vspace{-1.0em}
    \label{fig:spherical_disparity}
}
\end{figure}

%=--------------------------------------------------------------------------------%
\subsection{Mapping LiDAR to 360° Images}
\label{subsec:labelling}
%=--------------------------------------------------------------------------------%

The main challenge in processing our raw data is to ensure a robust and accurate projection of the LiDAR point clouds on the 360° camera images. The following is a summary of the algorithm's approach, presented in ~\cref{alg:lidarmapping}. At the beginning of each recorded session, we select a set of key 3D points in the point cloud obtained by the LiDAR scans on a 19x19 chessboard, focusing on corners and edge midpoints to obtain their coordinates. The corresponding image pixel positions are selected in the equirectangular bottom image. We apply initial rotation and translation transformations $R_{\text{init}}, T_{\text{init}}$ to make the camera lens the coordinate system's center, ensuring alignment with the camera's viewpoint: $p^{\text{cam}}_i = R_{\text{init}} p_{\text{LiDAR},i} + T_{\text{init}}$.
The transformed points are then converted into spherical coordinates $(r, \theta, \phi)$ to match the structure of equirectangular images, and projected onto the equirectangular plane.
The core step of the mapping process is the optimization of the rotation and translation parameters, achieved by minimizing the error between the projected points and their corresponding location on the equirectangular images. Subsequently, these optimized rotation matrix and translation vector compose the best-fit transformation to align LiDAR point clouds on the bottom equirectangular images.

To assess alignment accuracy, we calculated the pixel error by measuring Euclidean distances between 200 randomly selected LiDAR points and their image positions~\citep{Shen2023AnEP}, resulting in an average error of 1.7 pixels. When selecting the same number of points from visually challenging areas (e.g., object edges), the error remains relatively low at 8.0 pixels, further confirming calibration quality.  More details on the evaluation of the LiDAR point cloud mapping are available in \cref{appx_subsec:calib_quality}. Occlusions were not explicitly handled, as the small LiDAR-camera baseline minimizes their impact.

\begin{figure}[t]
  \centering
\begin{algorithm}[H]
\caption{LiDAR-to-Image Mapping Optimization}
\label{alg:lidarmapping}
\small
\begin{algorithmic}[1]
\STATE {\bfseries Input:} Set of LiDAR points $p^{\text{L}} \in \mathbb{R}^{n \times 3}$, set of image points $p^{\text{img}} \in \mathbb{R}^{n \times 2}$, image size $W \times H$, initial rotation and translation matrices $R_{\text{init}} \in \mathbb{R}^{3 \times 3}$, $T_{\text{init}} \in \mathbb{R}^{3}$.
\STATE {\bfseries Output:} Optimized matrices $R_{\text{opt}}, T_{\text{opt}}$.
\STATE Select $S$ LiDAR points and their corresponding image points.
\FOR{$i = 1$ \TO $S$}
    \STATE Transform $p^{\text{L}}_i$ to camera coordinates $p^{\text{cam}}_i$ from $R_{\text{init}}$, $T_{\text{init}}$.
    \STATE Convert $p^{\text{cam}}_i$ to spherical coordinates $(r_i, \theta_i, \phi_i)$.
    \STATE Project to equirectangular: $(x^{\text{eq}}_i, y^{\text{eq}}_i) = \left( \frac{\phi_i + \pi}{2\pi} W, \frac{\theta_i}{\pi} H \right)$.
    \STATE Compute error $e_i = \left\| p^{\text{img}}_i - (x^{\text{eq}}_i, y^{\text{eq}}_i) \right\|$.
\ENDFOR
\STATE Minimize the total error $E = \sum_{i=1}^{n} e_i^2$ using BFGS~\citep{Flet87}.
\end{algorithmic}
\end{algorithm}
\end{figure}

In omnidirectional stereo imaging, spherical disparity is defined as the angular difference between 360° cameras-point rays (see \cref{fig:spherical_disparity}). After mapping, we obtain disparity values from depth values to disparity via the following equation:
\begin{equation}
d = \arctan\left(\frac{\sin(\theta_b)}{r_{\text{bottom}} / B_{\text{camera}} - \cos(\theta_b)}\right),
\end{equation}
where \(\theta_b\) denotes the polar angle, \(r_{\text{bottom}}\) is the depth, \(B_{\text{camera}}\) the camera baseline, and \( d \) is the disparity in radians, converted to degrees in our dataset.

\begin{figure*}[t]
    \centering
    \begin{subfigure}[t]{0.30\linewidth}
        \includegraphics[width=\linewidth]{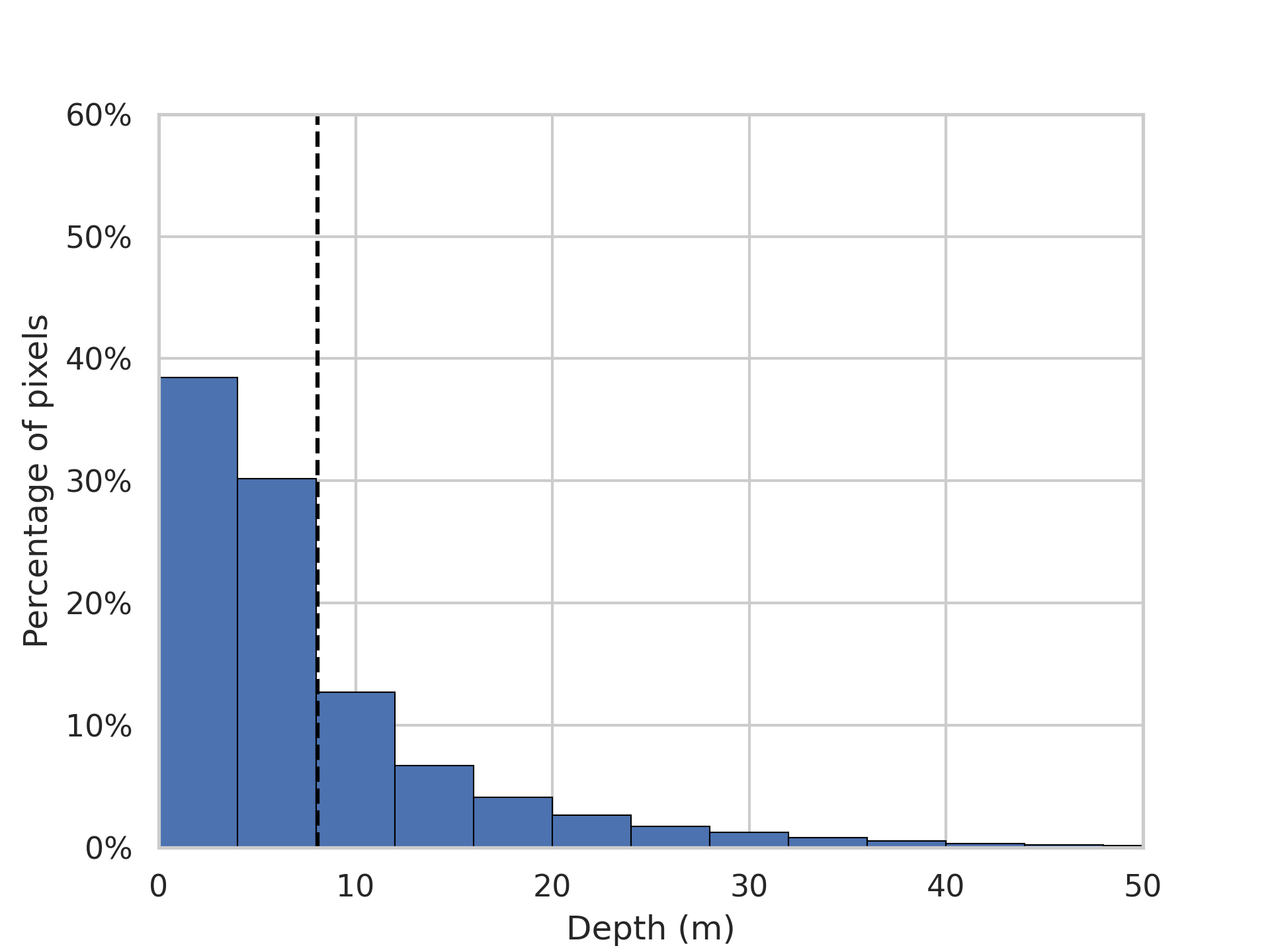}
        \caption{All sequences}
        \label{fig:depth-all}
    \end{subfigure}%
    %\hspace{2mm}
    \begin{subfigure}[t]{0.30\linewidth}
        \includegraphics[width=\linewidth]{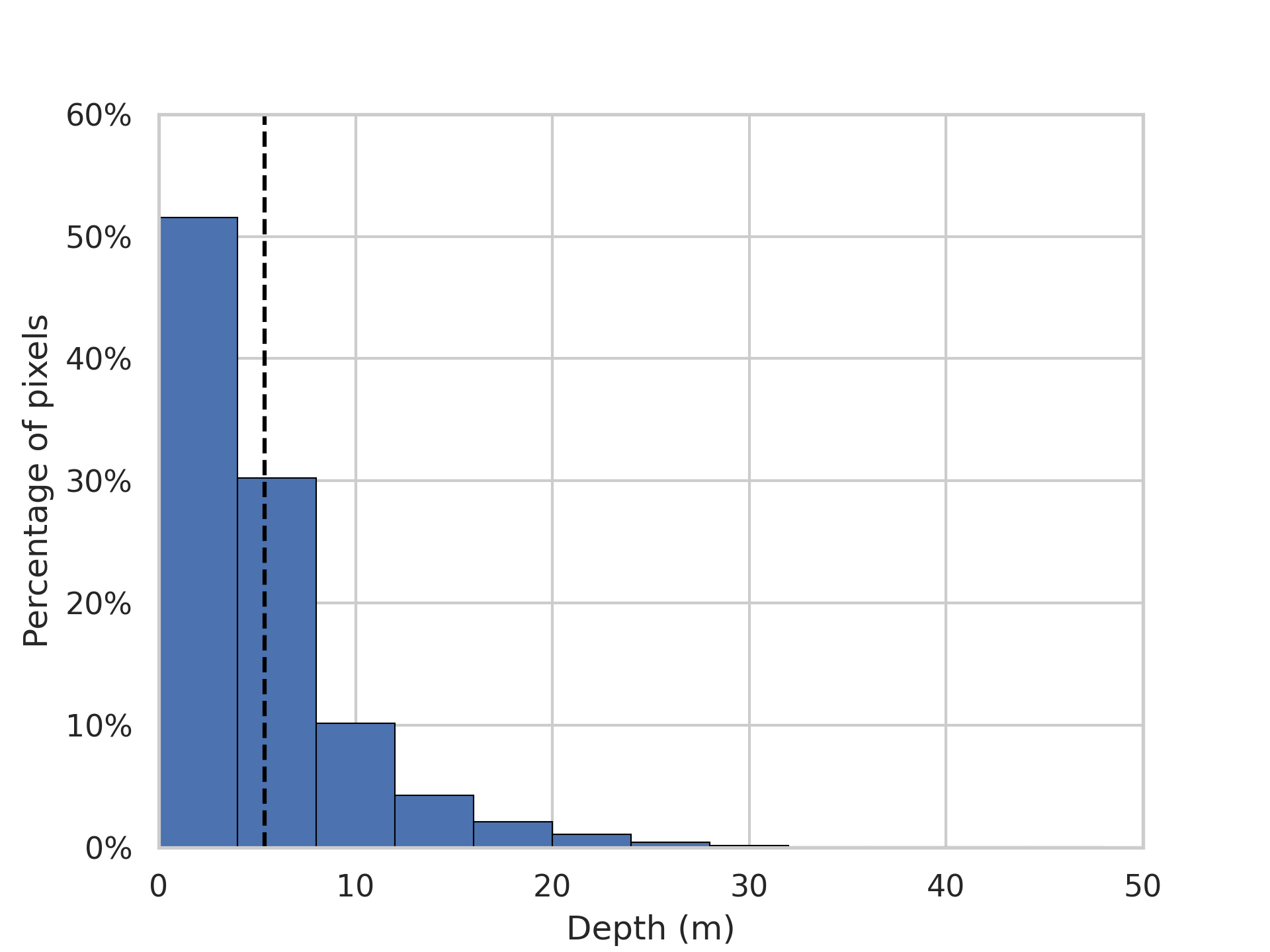}
        \caption{Indoor sequences}
        \label{fig:depth-indoor}
    \end{subfigure}%
    %\hspace{2mm}
    \begin{subfigure}[t]{0.30\linewidth}
        \includegraphics[width=\linewidth]{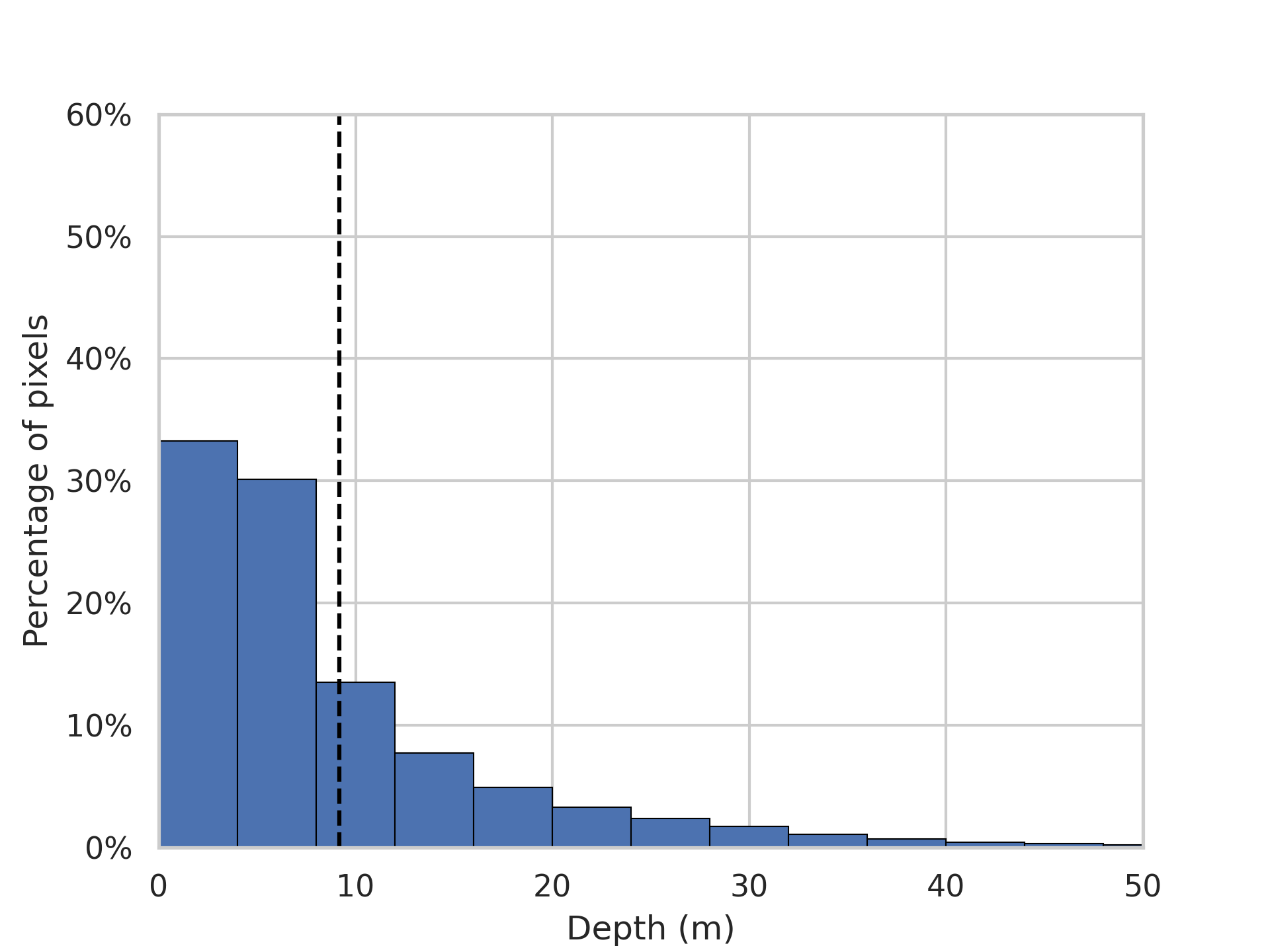}
        \caption{Outdoor sequences}
        \label{fig:depth-outdoor}
    \end{subfigure}
    \caption{\textbf{Histograms of depth values} for the entire dataset without depth completion, indoor and outdoor (day and night) sequences. Vertical dotted lines indicate the average depth for each setting. Depth values range from 0.5m to 225m, with averages of 8.1m overall, 5.4m for indoor scenes, and 9.2m for combined day and night outdoor scenes.}
    \label{fig:stats}
\end{figure*}

%=--------------------------------------------------------------------------------%
\subsection{Depth Completion from LiDAR Point Clouds}
\label{subsec:depth_completion}
%=--------------------------------------------------------------------------------%

Given the sparse nature of LiDAR point clouds compared to high-resolution camera images, ground-truth depth labels cover only a fraction of the total pixels. To address this issue, we develop a fully automated depth completion pipeline tailored for 360° images. Existing depth completion methods share the idea of aggregating point clouds but either need manual annotations~\citep{Geiger2012CVPR,Menze2015CVPR} or are limited to rectilinear images~\citep{drivingstereo}. Our approach interpolates depth on spherical grid points by using temporally aggregated point clouds and applies a filtering process to exclude points with high uncertainty or lacking nearby measurements, ensuring minimal error before mapping the remaining points. 

\vspace{0.2cm}
\noindent
\textbf{Temporal aggregation.}
To increase the number of valid points, we aggregate the the current frame's point cloud with those from the four preceding and four succeeding frames. Given this limited temporal window, the error introduced by the movement of the robot, or within the scene, is negligible, given the LiDAR's high frame rate and the rig's sub-pedestrian moving speed (see \cref{appx_subsec:comp_temp_agg}). Note that this approach is limited for fast-moving objects like cars.

\vspace{0.2cm}
\noindent
\textbf{Interpolation.}
A point on a spherical grid is defined by its polar angle $\theta \in [0, \pi]$, azimuthal angle $\varphi \in [-\pi, \pi]$, and radial distance $r$. For a query point $(\theta_q, \varphi_q)$ within the LiDAR's field of view, we estimate its depth through a weighted average of its $k$-nearest neighbors, $r_q = \sum_{i=1}^{k} w_i \times r_i$, where $w_i$ represents the weight of each neighboring depth value $r_i$, calculated as: \begin{equation}
    w_j = \frac{d_{q\leftrightarrow j}(\theta_q, \theta_j, \varphi_q, \varphi_j)^{-1}}{\sum_{i=1}^{k} d_{q\leftrightarrow i}(\theta_q, \theta_i, \varphi_q, \varphi_i)^{-1}}.
\end{equation}
Here, $d_{q\leftrightarrow i}$ is the Euclidean distance between the query point and its neighbors in spherical coordinates.

\vspace{0.2cm}
\noindent
\textbf{Filtering.}
To ensure high-quality depth labels, we filter query points exceeding either an uncertainty criterion or an out-of-distribution threshold before mapping the remaining points to the image. For uncertainty estimation, given that we use the weighted mean of the $k$-nearest neighbors to interpolate depth values, we calculate the relative weighted variance of interpolated depth values:
\begin{equation}
    \sigma_{r_q}^2 = \sum_{i=1}^{k} w_i \times \left( \frac{r_q - r_i}{r_q} \right)^2.
\end{equation}
The relative term $(r_q - r_i)/r_q$ ensures that points across the entire depth range of the scene are treated equally.
The choice of the uncertainty threshold is based on a predefined Ratio of Interpolated Points (\textit{RIP}) after filtering.

Additionally, to avoid labeling regions with insufficient nearby points (e.g., sky areas), we compute the average distance to the $k$ nearest neighbors for each query point: 
\begin{equation}
    d_q = \frac{\sum_{i=1}^{k} d_{q\leftrightarrow i}(\theta_q, \theta_i, \varphi_q, \varphi_i)}{k}.
\end{equation}
If this distance exceeds a threshold, the depth value is excluded. Details on the heuristic used to determine the threshold are provided in \cref{appx_subsec:hyperparameters}.

\vspace{0.2cm}
\noindent
\textbf{Quantitative evaluation.}
To validate our depth completion pipeline, we split each LiDAR point cloud into training and test sets, and compute depth estimation metrics on the test set. Applied on the \textsc{Helvipad} dataset, our pipeline significantly increases the ratio of labeled pixels from 12\% to 61\% while maintaining very low error rates. Additional details on the validation method, hyper-parameters and error metrics can be found in  \cref{appx_sec:depth_completion}.

%\textcolor{red}{Hyperparameters $n_\text{grid}$ (number of points in spherical grid), $t_\theta$ (boundaries of field of view of LiDAR sensor) are missing but appendix is also okay if you think that we don't have the space}

%=--------------------------------------------------------------------------------%
\subsection{Benchmark and Data Statistics}
\label{subsec:statistics}
%=--------------------------------------------------------------------------------%

After cropping unnecessary regions (sky, ground), the resulting dataset comprises 39,553 paired images with their corresponding depth and disparity maps, at a resolution of 1920 $\times$ 512 pixels. To create the benchmark, we split the captured sequences into a train-val set and a test set, ensuring the same proportion of \textit{outdoor}, \textit{indoor} and \textit{night outdoor} sequences. Each test sequence has been manually reviewed to ensure no overlap of scenes or areas with the train set. The train-val set includes 20 sequences -- 13 \textit{outdoor}, 5 \textit{indoor} and 2 \textit{night outdoor} sequences -- with a total of 29,407 frames. The test split consists of 6 sequences -- 3 \textit{outdoor}, 2 \textit{indoor} and 1 \textit{night outdoor} sequences -- with a total of 10,146 frames. \cref{fig:stats} presents depth distributions of the entire dataset and by scene types. Disparity distributions and distributions for augmented training data are also provided in \cref{appx_subsec:histograms}.

\begin{figure*}[t]
\centering
 \includegraphics[width=\linewidth]{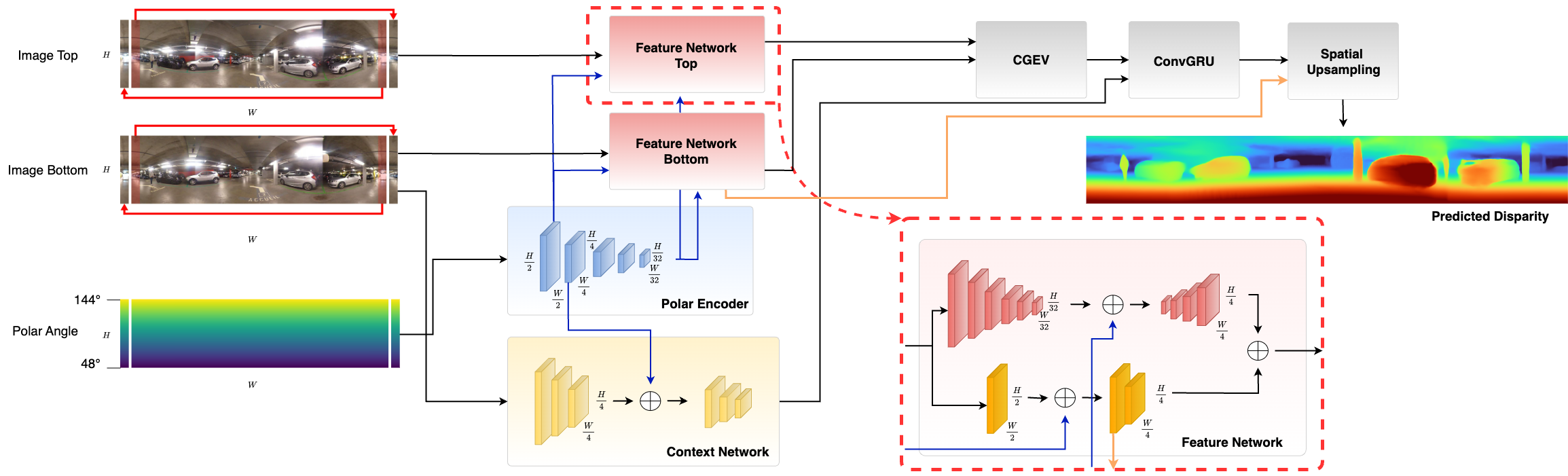}
 \caption{\textbf{Overview of 360-IGEV-Stereo architecture}.
 The model takes the circular padded top and bottom image as well as a polar angle map with equal size as an input.
 At the bottleneck of the feature network the feature map is concatenated with the encoded polar angle at 1/32 of the original image size.
The encoded polar angle is also concatenated with the context feature maps at 1/4 resolution.
Subsequently, the Combined Geometry Encoding Volume (CGEV) is constructed by vertical warping.
The iterative refinement of the disparity with the ConvGRU and the spatial upsampling are equivalent to IGEV-Stereo~\citep{xu2023iterative}.}
 \label{fig:architecture_360_igev_stereo}
\end{figure*}

%=================================================================================%
\section{Adapting Stereo Matching for 360° Imaging}
\label{sec:omni_stereo}
%=================================================================================%

Omnidirectional images with equirectangular projections differ from traditional rectilinear images due to spherical distortions and the continuity at the left and right edges.
As most existing stereo depth estimation models are designed for two rectilinear cameras in a left-right setup, we propose two key adaptations -- polar angle incorporation and circular padding -- to enable these models to handle omnidirectional stereo datasets effectively. In this section, we illustrate how these adaptations were applied to adapt the recent state-of-the-art model, IGEV-Stereo~\citep{xu2023iterative}.

\vspace{0.2cm}
\noindent
\textbf{Incorporating polar angle.} In the top-bottom setup, images are warped vertically to form cost volumes. Distortions in omnidirectional images vary with the polar angle $\theta$, which affects the calculations of the disparity in the vertical field of view. To address this, we follow the approach by \citet{wang20icra} and add a polar angle map as input to the model. To minimize parameter overhead, we use a shared polar map encoder for both the feature network and the context network. The polar map encoder comprises strided convolutional layers, and each lowest possible resolution is selected for concatenation with the encoded images.

\vspace{0.2cm}
\noindent
\textbf{Circular padding.} The continuity of omnidirectional images can be leveraged by applying circular padding along the horizontal dimension. The image's left side is padded with the 64 rightmost pixel columns, while the right side is extended with the leftmost part of the original image. This enables the network to take advantage of contextual information across image boundaries. \\

These adaptations and the general network architecture of the method, named \textit{360-IGEV-Stereo}, are illustrated in \cref{fig:architecture_360_igev_stereo}. Additionally, to handle the variability in lighting conditions in the dataset, we apply photometric data augmentation during training.

\begin{table*}[t]
    \centering
    \setlength{\tabcolsep}{5pt}
    \resizebox{\textwidth}{!}{%
        \begin{tabular}{lc ccc cccc c}
            \toprule
            \multirow{3}{*}{\vspace{0.5em}\textbf{Method}} & \multirow{3}{*}{\vspace{0.5em}\textbf{Stereo Setting}} & \multicolumn{3}{c}{\textbf{Disparity} (°)} & \multicolumn{4}{c}{\textbf{Depth} (m)} & \multicolumn{1}{c}{\textbf{Runtime} (s)} \\
            \cmidrule(lr){3-5} \cmidrule(lr){6-9}
            & & MAE $\downarrow$ & RMSE $\downarrow$ & MARE $\downarrow$ & MAE $\downarrow$ & RMSE $\downarrow$ & MARE $\downarrow$ & LRCE $\downarrow$ & \\
            \midrule
            PSMNet~\citep{chang2018pyramid} & conventional & 0.286 & 0.496 & 0.248 & 2.509 & 5.673 & 0.176 & 1.809 & 0.66 \\
            360SD-Net~\citep{wang20icra} & omnidirectional & 0.224 & 0.419 & 0.191 & 2.122 & 5.077 & 0.152 & 0.904 & 0.63 \\
            IGEV-Stereo~\citep{xu2023iterative} & conventional & 0.225 & 0.423 & 0.172 & 1.860 & 4.474 & 0.146 & 1.203 & \bfseries 0.24 \\
            \midrule
            \rowcolor{gray!25} 360-IGEV-Stereo & omnidirectional & \bfseries 0.188 & \bfseries 0.404 & \bfseries 0.146 & \bfseries 1.720 & \bfseries 4.297 & \bfseries 0.130 & \bfseries0.388  & 0.25 \\
            \bottomrule
        \end{tabular}
    }
    \caption{\textbf{Comparative results of depth estimation baselines on the \textsc{Helvipad} test set}. All methods are trained on the augmented train set. 360-IGEV-Stereo achieves the strongest performance across all disparity and depth metrics. The best results are highlighted in \textbf{bold}.}
    \label{tab:comparative_results}
\end{table*}

%=================================================================================%
\section{Experiments}
\label{sec:experiments}
%=================================================================================%

In this section, we evaluate the performance of multiple state-of-the-art and popular stereo matching methods, both for standard and 360° images. We also analyze their cross-scene generalization within the dataset and include ablation studies to evaluate the proposed adaptations and the depth completion pipeline as training data augmentation.

%=--------------------------------------------------------------------------------%
\subsection{Setup and Baselines}
\label{subsec:setup}
%=--------------------------------------------------------------------------------%

All models are trained on a single NVIDIA A100 GPU with the largest possible batch size to ensure comparable use of computational resources. Model selection employed early stopping based on validation performance. Implementation details are provided in \cref{appx_subsec:hyperparameters}.

\vspace{0.2cm}
\noindent
\textbf{Baselines.} We implemented two well established learning based stereo depth estimation models, the popular \textbf{PSMNet}~\citep{chang2018pyramid} and \textbf{IGEV-Stereo}~\citep{xu2023iterative} presented above, the latter also enabling to assess the effectiveness of our proposed adaptations in \cref{sec:omni_stereo}. For omnidirectional imaging, we include \textbf{360SD-Net}~\citep{wang20icra}, the only 360° stereo method designed for top-bottom cameras setups.

\vspace{0.2cm}
\noindent
\textbf{Evaluation metrics.} We measure each method's performance by comparing their disparity and depth estimates to sparse ground-truth labels using Mean Absolute Error (MAE), Root Mean Squared Error (RMSE), and Mean Absolute Relative Error (MARE). MAE indicates overall error magnitude, while RMSE emphasizes the impact of outliers. Depth errors can increase significantly due to the non-linear relationship between depth and disparity, making MARE particularly useful as a scale-invariant metric for assessing performance across varying depth ranges. Dealing with 360° imaging, we also evaluate boundary consistency using the Left-Right Consistency Error (LRCE)~\citep{shen2022panoformer}.

\subsection{Comparative Results}
\label{subsec:comparative_results}

\cref{tab:comparative_results} presents the performance of various stereo depth estimation methods evaluated on the \textsc{Helvipad} test set.  While \textbf{360SD-Net} improve every metrics over its rectilinear stereo counterpart \textbf{PSMNet}, it falls short of matching the depth accuracy of more recent models like \textbf{IGEV-Stereo} in metrics such as MAE and RMSE. This indicates that \textbf{IGEV-Stereo} has a notable advantage in minimizing larger depth errors compared to \textbf{360SD-Net}. However, in terms of depth MARE and disparity metrics, \textbf{360SD-Net} performs comparably to \textbf{IGEV-Stereo}, which suggests that \textbf{360SD-Net} design for top-bottom 360° configurations helps to accurately estimate disparity.

With its omnidirectional adaptations, \textbf{360-IGEV-Stereo} achieves the best overall performance across all disparity and depth metrics on the \textsc{Helvipad} test set. Compared to \textbf{IGEV-Stereo}, our adapted model reduces depth MAE from 1.81m to 1.77m and achieves a consistent improvement in disparity metrics, including a reduction of disparity MAE from 0.22° to 0.18°. This highlights the benefits of adapting stereo depth estimation models for omnidirectional data with minimal computational overhead. As expected, \textbf{360-IGEV-Stereo} also achieves the lowest LRCE, thanks to the use of circular padding during inference, and \textbf{360SD-Net}, the other omnidirectional approach, ranks second.

%=--------------------------------------------------------------------------------%
\subsection{Ablations}
\label{subsec:ablations}
%=--------------------------------------------------------------------------------%

\begin{figure}[ht]
  \centering
    \includegraphics[width=0.97\linewidth]{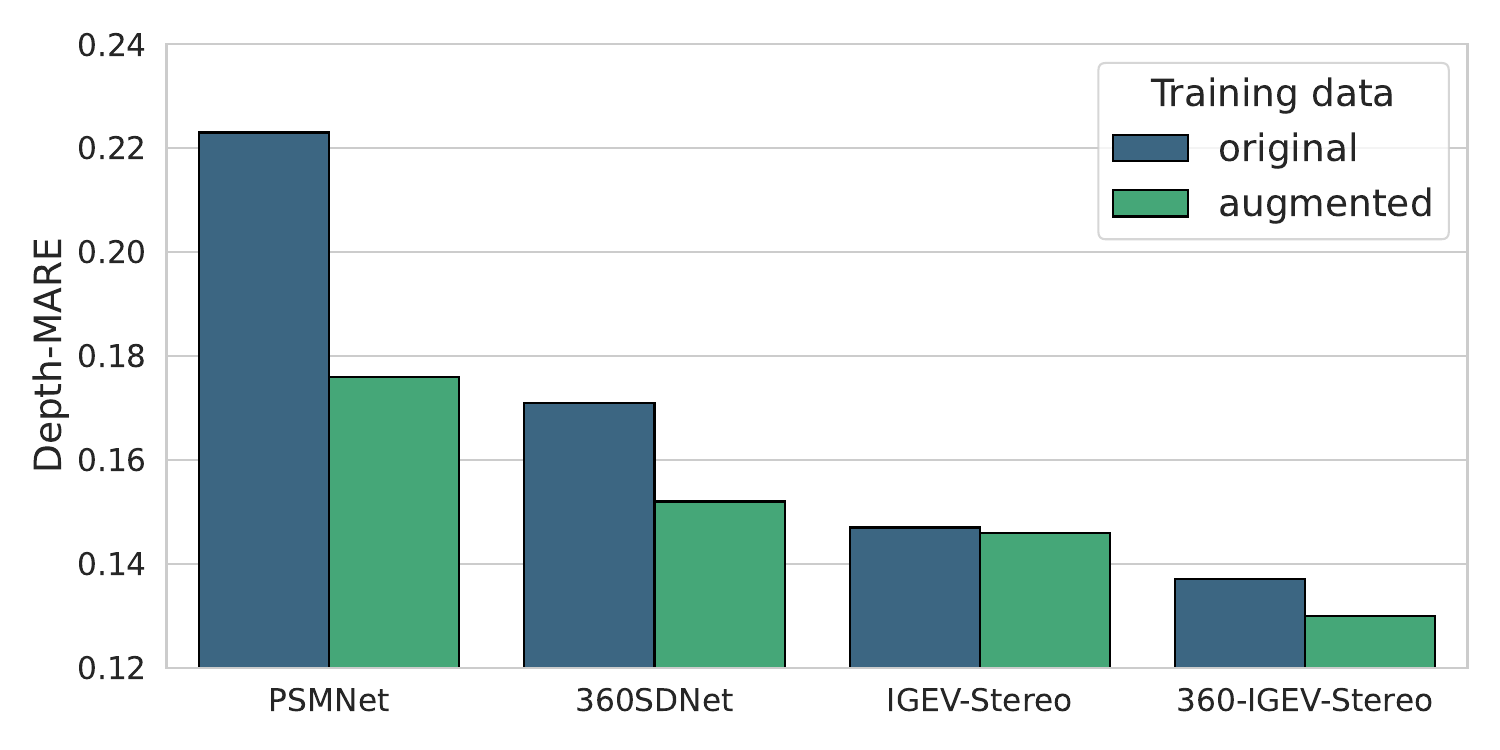}
    \caption{\textbf{Depth MARE comparison} across different depth estimation methods trained on original vs. augmented depth labels.}
    \label{fig:results_depth_completion}
\end{figure}

\vspace{0.2cm}
\noindent
\textbf{Depth completion as data augmentation.}
As shown in \cref{fig:results_depth_completion}, training with the depth-completed augmented set improves performances. For instance, \textbf{360SD-Net}'s depth MARE drops from 0.17 to 0.15 with augmentation, with \textbf{PSMNet} showing a similar reduction. The performance gains are less pronounced for more advanced models like \textbf{IGEV-Stereo} and \textbf{360-IGEV-Stereo}.

\begin{figure*}[t]
\centering
 \includegraphics[width=\textwidth]{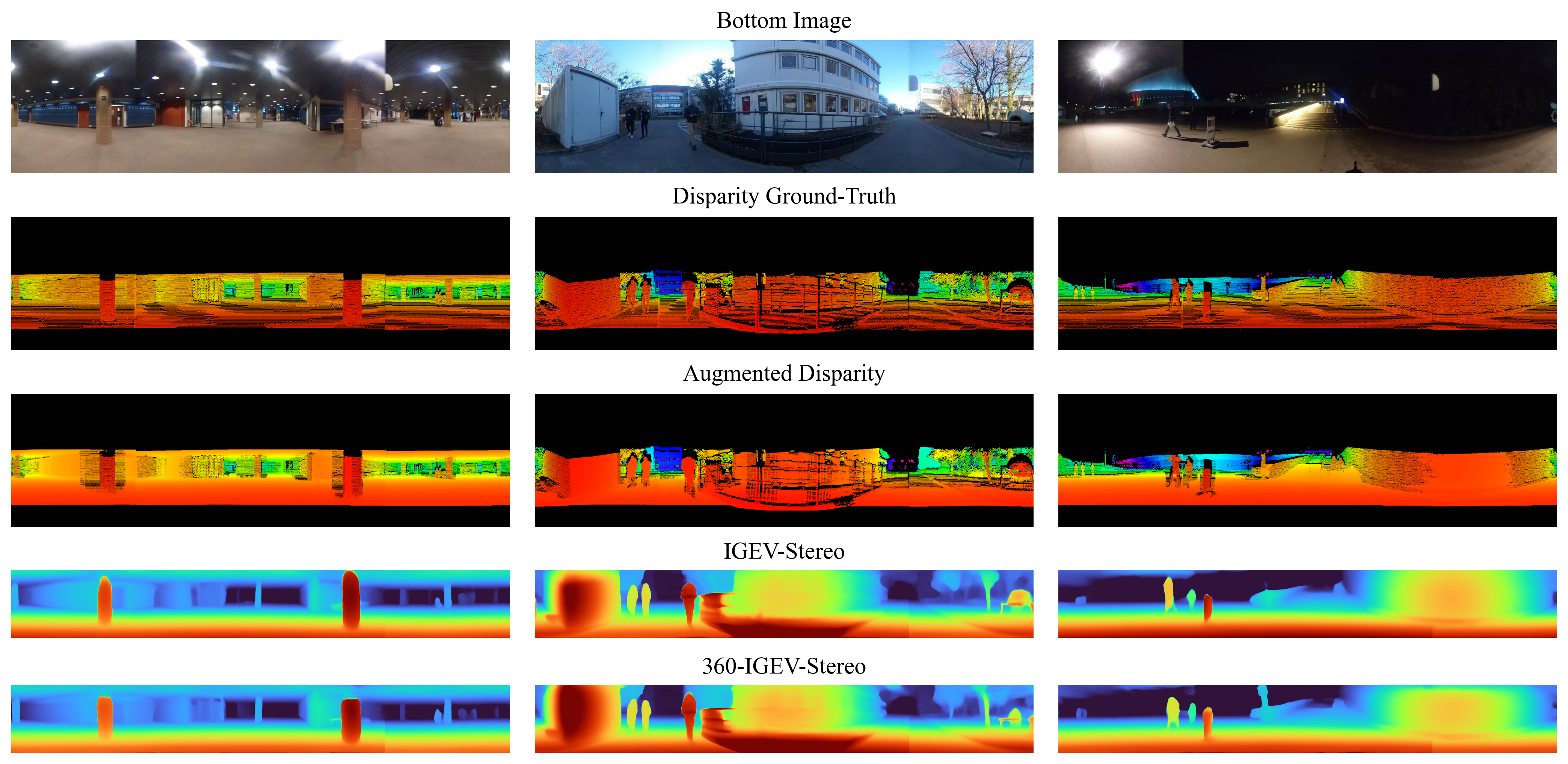}
 \caption{\textbf{Visualization of disparity map predictions using IGEV-Stereo and 360-IGEV-Stereo compared to ground-truth sparse and augmented labels}. The top row depicts images from the bottom camera, followed by the original and the augmented disparity maps from depth completion. The final two rows presents the predicted disparity maps from IGEV-Stereo and 360-IGEV-Stereo.}
 \label{fig:qualitative_results}
\end{figure*}

\vspace{0.2cm}
\noindent
\textbf{Omnidirectional adaptations.}
During inference, circular padding is applied to account for the continuous 360° view, which helps to reduce the LRCE from 1.18m to 0.36m, and consequently also the depth RMSE, as shown in \cref{table:ablation}. This enhancement is visually noticeable in \cref{fig:qualitative_results}, where padding contributes to smoother depth transitions across the left and right image boundaries. Further, photometric data augmentation helps models to generalize across the diverse lighting conditions in the \textsc{Helvipad} dataset.

\begin{table}[ht]
\centering
\setlength\tabcolsep{5pt}
\resizebox{\columnwidth}{!}{%
\begin{tabular}{l c c c c}
\toprule
\textbf{Variant} & \textbf{MAE~$\downarrow$} & \textbf{RMSE~$\downarrow$} & \textbf{MARE~$\downarrow$} & \textbf{LRCE~$\downarrow$} \\
\midrule
360-IGEV-Stereo & \bfseries 1.720 & \bfseries 4.297 & \bfseries 0.130 & 0.388 \\
\hspace{2mm} - w/o circular padding & 1.726 & 4.314  & \bfseries 0.130 & $1.153$ \\
\hspace{2mm} - w/o photometric augmentation & $1.845$ & $4.466$ & 0.135 & \bfseries 0.347 \\
\bottomrule
\end{tabular}%
}
\caption{\textbf{Ablation study on adaptations for 360-IGEV-Stereo method.} Results are reported using depth (m) metrics.}
\label{table:ablation}
\end{table}

\begin{figure*}[t]
\centering
 \includegraphics[width=\textwidth]{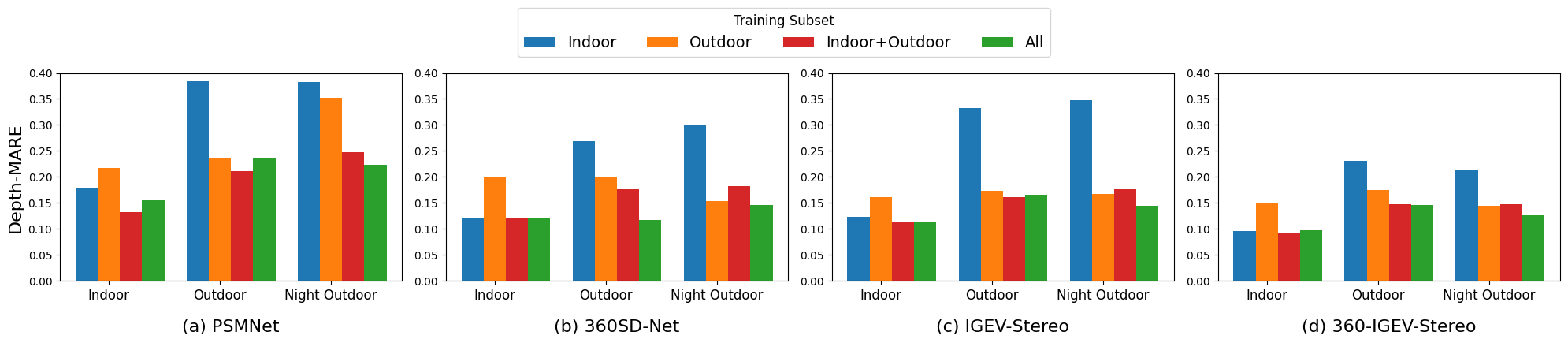}
 \caption{\textbf{Cross-scene generalization analysis} of each model when trained on different subsets (Indoor, Outdoor, Indoor+Outdoor, All) and evaluated under various testing conditions (indoor, outdoor, night outdoor).
 We use the depth MARE for comparison.
 }
 \label{fig:analys_per_test_condition_test}
\end{figure*}

%=--------------------------------------------------------------------------------%
\subsection{Cross-Scene Generalization}
\label{subsec:generalization}
%=--------------------------------------------------------------------------------%

To safely deploy learning systems in real-world settings, it is essential to assess their ability to generalize to unseen or underrepresented conditions, like outdoor environments at night. The bar chart in \cref{fig:analys_per_test_condition_test} shows the performance of our proposed omnidirectional adaptation, \textbf{360-IGEV-Stereo}, alongside other evaluated methods, using depth MARE as the metric. We trained model variants on different subsets (Indoor, Outdoor, Indoor+Outdoor) and tested them under various conditions (Indoor, Outdoor, Night Outdoor). While models trained on specific environments perform well within their respective domains, indoor-trained models show a clear limitation in generalizing to outdoor scenes, particularly at night. Interestingly, this performance drop is less pronounced in omnidirectional models, such as \textbf{360-IGEV-Stereo} and \textbf{360SD-Net}, than in conventional models, like \textbf{IGEV-Stereo} and \textbf{PSMNet}. This result suggests that omnidirectional approaches benefit from improved cross-scene generalization capacities. 

Models trained on Indoor+Outdoor\footnote{This subset excludes night outdoor images from training.} subset outperform those trained solely on singular environments. Finally, best results are obtained with models trained on the whole training set, except for indoor test setting with \textbf{PSMNet} where the model trained on only indoor and daytime outdoor scenes performs slightly better. This underscores the importance of diverse training data for robust learning.

%=--------------------------------------------------------------------------------%
\subsection{Qualitative Results}
\label{subsec:qualitative}
%=--------------------------------------------------------------------------------%

We present visualizations of the depth completion pipeline and qualitative results of \textbf{IGEV-Stereo} and \textbf{360-IGEV-Stereo} on test images in \cref{fig:qualitative_results}. Depth completion produces dense, coherent disparity maps aligned with the ground truth and does not incorporate incorrect labels at depth boundaries. For stereo matching, \textbf{360-IGEV-Stereo} displays closer alignment to the ground truth. For example, in the right image, \textbf{IGEV-Stereo} misses two small people on the left, whereas \textbf{360-IGEV-Stereo} successfully captures them, indicating a better capacity to detect small objects. In the left image, the use of circular padding by \textbf{360-IGEV-Stereo} results in smoother transitions at the image boundaries. The incorporation of the polar angle also contributes to straighter lines and sharper visualizations, improving the representation of elements affected by distortions, such as pillars in the left image.

%=================================================================================%
\section{Conclusion}
\label{sec:conclusion}
%=================================================================================%

We build \textsc{Helvipad}, a real-world stereo dataset for omnidirectional depth estimation, and show that incorporating polar angle as input and circular padding help to adapt deep stereo matching models to equirectangular images. We also introduce a depth completion pipeline tailored to equirectangular projections which increases label density from LiDAR point clouds. We hope this work encourages further research into omnidirectional stereo matching, especially for real-time applications. This dataset is an ideal testbed for assessing the robustness of depth estimation methods to diverse lighting conditions and depth ranges.

%=================================================================================%
\section*{Acknowledgments}
\label{sec:acknowledgments}
%=================================================================================%

This work was supported by the EPFL Center for Imaging through a Collaborative Imaging Grant. We thank the VITA lab members for their valuable feedback, which helped to enhance the quality of this manuscript. We also express our gratitude to Dr. Simone Schaub-Meyer and Oliver Hahn for their insightful advice during the project's final stages. Additionally, we sincerely thank the anonymous reviewers for their constructive feedback and thoughtful suggestions, which significantly contributed to improving the clarity and robustness of this work.

{
    \small
    \bibliographystyle{ieeenat_fullname}
    \bibliography{main}

\begin{thebibliography}{45}
\providecommand{\natexlab}[1]{#1}
\providecommand{\url}[1]{\texttt{#1}}
\expandafter\ifx\csname urlstyle\endcsname\relax
  \providecommand{\doi}[1]{doi: #1}\else
  \providecommand{\doi}{doi: \begingroup \urlstyle{rm}\Url}\fi

\bibitem[Ai et~al.(2022)Ai, Cao, Zhu, Bai, Chen, and Wang]{Ai2022DeepLF}
Hao Ai, Zidong Cao, Jin Zhu, Haotian Bai, Yucheng Chen, and Ling Wang.
\newblock Deep learning for omnidirectional vision: A survey and new perspectives.
\newblock \emph{arXiv:2205.10468 [cs.CV]}, 2022.

\bibitem[Albanis et~al.(2021)Albanis, Zioulis, Drakoulis, Gkitsas, Sterzentsenko, Alvarez, Zarpalas, and Daras]{Albanis_2021_CVPR}
Georgios Albanis, Nikolaos Zioulis, Petros Drakoulis, Vasileios Gkitsas, Vladimiros Sterzentsenko, Federico Alvarez, Dimitrios Zarpalas, and Petros Daras.
\newblock {Pano3D}: A holistic benchmark and a solid baseline for 360{\textdegree} depth estimation.
\newblock In \emph{Proceedings of the IEEE/CVF Conference on Computer Vision and Pattern Recognition (CVPR) Workshops}, 2021.

\bibitem[{Armeni} et~al.(2017){Armeni}, {Sax}, {Zamir}, and {Savarese}]{SF3D}
I. {Armeni}, A. {Sax}, A.~R. {Zamir}, and S. {Savarese}.
\newblock Joint 2d-3d-semantic data for indoor scene understanding.
\newblock \emph{arxiv:1702.01105 [cs.CV]}, 2017.

\bibitem[Caesar et~al.(2020)Caesar, Bankiti, Lang, Vora, Liong, Xu, Krishnan, Pan, Baldan, and Beijbom]{nuscenes}
Holger Caesar, Varun Bankiti, Alex~H Lang, Sourabh Vora, Venice~Erin Liong, Qiang Xu, Anush Krishnan, Yu Pan, Giancarlo Baldan, and Oscar Beijbom.
\newblock nuscenes: A multimodal dataset for autonomous driving.
\newblock In \emph{Proceedings of the IEEE/CVF Conference on Computer Vision and Pattern Recognition (CVPR)}, 2020.

\bibitem[Chang et~al.(2017)Chang, Dai, Funkhouser, Halber, Niessner, Savva, Song, Zeng, and Zhang]{Matterport3D}
Angel Chang, Angela Dai, Thomas Funkhouser, Maciej Halber, Matthias Niessner, Manolis Savva, Shuran Song, Andy Zeng, and Yinda Zhang.
\newblock Matterport3d: Learning from rgb-d data in indoor environments.
\newblock In \emph{Proceedings of the International Conference on 3D Vision (3DV)}, 2017.

\bibitem[Chang and Chen(2018)]{chang2018pyramid}
Jia-Ren Chang and Yong-Sheng Chen.
\newblock Pyramid stereo matching network.
\newblock In \emph{Proceedings of the IEEE/CVF Conference on Computer Vision and Pattern Recognition (CVPR)}, 2018.

\bibitem[Chen et~al.(2023)Chen, Lin, Lang, Liao, and Zhao]{Chen2023UnsupervisedOE}
Zisong Chen, Chunyu Lin, Nie Lang, Kang Liao, and Yao Zhao.
\newblock Unsupervised omnimvs: Efficient omnidirectional depth inference via establishing pseudo-stereo supervision.
\newblock In \emph{IEEE/RSJ International Conference on Intelligent Robots and Systems (IROS)}, 2023.

\bibitem[Cho et~al.(2014)Cho, van Merrienboer, Bahdanau, and Bengio]{cho2014properties}
Kyunghyun Cho, Bart van Merrienboer, Dzmitry Bahdanau, and Yoshua Bengio.
\newblock On the properties of neural machine translation: Encoder-decoder approaches.
\newblock In \emph{Proceedings of {SSST}-8, Eighth Workshop on Syntax, Semantics and Structure in Statistical Translation}, pages 103--111, 2014.

\bibitem[Couprie et~al.(2013)Couprie, Farabet, Najman, and LeCun]{NYUv2}
Camille Couprie, Cl{\'e}ment Farabet, Laurent Najman, and Yann LeCun.
\newblock Indoor semantic segmentation using depth information.
\newblock In \emph{International Conference on Representation Learning (ICLR)}, 2013.

\bibitem[Cui et~al.(2017)Cui, Ji, Shan, Gong, and Liu]{line_registration_2017}
Tingting Cui, Shunping Ji, Jie Shan, Jianya Gong, and Kejian Liu.
\newblock Line-based registration of panoramic images and lidar point clouds for mobile mapping.
\newblock \emph{Sensors}, 17, 2017.

\bibitem[De~Alvis et~al.(2019)De~Alvis, Shan, Worrall, and Nebot]{uncertainty_projection_2019}
Charika De~Alvis, Mao Shan, Stewart Worrall, and Eduardo Nebot.
\newblock Uncertainty estimation for projecting lidar points onto camera images for moving platforms.
\newblock In \emph{International Conference on Robotics and Automation (ICRA)}, 2019.

\bibitem[Droukas et~al.(2023)Droukas, Doulgeri, Tsakiridis, Triantafyllou, Kleitsiotis, Mariolis, Giakoumis, Tzovaras, Kateris, and Bochtis]{Droukas_2023}
Leonidas Droukas, Zoe Doulgeri, Nikolaos~L. Tsakiridis, Dimitra Triantafyllou, Ioannis Kleitsiotis, Ioannis Mariolis, Dimitrios Giakoumis, Dimitrios Tzovaras, Dimitrios Kateris, and Dionysis Bochtis.
\newblock A survey of robotic harvesting systems and enabling technologies.
\newblock \emph{Journal of Intelligent and Robotic Systems}, 107\penalty0 (2), 2023.

\bibitem[Fletcher(1987)]{Flet87}
Roger Fletcher.
\newblock \emph{Practical Methods of Optimization}.
\newblock John Wiley \& Sons, New York, NY, USA, second edition, 1987.

\bibitem[Geiger et~al.(2012)Geiger, Lenz, and Urtasun]{Geiger2012CVPR}
Andreas Geiger, Philip Lenz, and Raquel Urtasun.
\newblock Are we ready for autonomous driving? the kitti vision benchmark suite.
\newblock In \emph{Proceedings of the IEEE Conference on Computer Vision and Pattern Recognition (CVPR)}, 2012.

\bibitem[Guan et~al.(2024)Guan, Wang, and Liu]{guan2024neural}
Tongfan Guan, Chen Wang, and Yun-Hui Liu.
\newblock Neural markov random field for stereo matching.
\newblock In \emph{Proceedings of the IEEE/CVF Conference on Computer Vision and Pattern Recognition (CVPR)}, 2024.

\bibitem[Guo et~al.(2019)Guo, Yang, Yang, Wang, and Li]{guo2019group}
Xiaoyang Guo, Kai Yang, Wukui Yang, Xiaogang Wang, and Hongsheng Li.
\newblock Group-wise correlation stereo network.
\newblock In \emph{Proceedings of the IEEE/CVF Conference on Computer Vision and Pattern Recognition (CVPR)}, 2019.

\bibitem[Huang et~al.(2019)Huang, Wang, Cheng, Zhou, Geng, and Yang]{Apolloscape}
Xinyu Huang, Peng Wang, Xinjing Cheng, Dingfu Zhou, Qichuan Geng, and Ruigang Yang.
\newblock The apolloscape open dataset for autonomous driving and its application.
\newblock \emph{IEEE Transactions on Pattern Analysis and Machine Intelligence (PAMI)}, 42\penalty0 (10):\penalty0 2702--2719, 2019.

\bibitem[Janai et~al.(2020)Janai, G\"{u}ney, Behl, and Geiger]{janai2017computer}
Joel Janai, Fatma G\"{u}ney, Aseem Behl, and Andreas Geiger.
\newblock Computer vision for autonomous vehicles: Problems, datasets and state of the art.
\newblock \emph{Found. Trends. Comput. Graph. Vis.}, 2020.

\bibitem[Kendall et~al.(2017)Kendall, Martirosyan, Dasgupta, Henry, Kennedy, Bachrach, and Bry]{kendall2017end}
Alex Kendall, Hayk Martirosyan, Saumitro Dasgupta, Peter Henry, Ryan Kennedy, Abraham Bachrach, and Adam Bry.
\newblock End-to-end learning of geometry and context for deep stereo regression.
\newblock In \emph{Proceedings of the International Conference on Computer Vision ({ICCV})}, 2017.

\bibitem[Kyrarini et~al.(2021)Kyrarini, Lygerakis, Rajavenkatanarayanan, Sevastopoulos, Nambiappan, Chaitanya, Babu, Mathew, and Makedon]{technologies9010008}
Maria Kyrarini, Fotios Lygerakis, Akilesh Rajavenkatanarayanan, Christos Sevastopoulos, Harish~Ram Nambiappan, Kodur~Krishna Chaitanya, Ashwin~Ramesh Babu, Joanne Mathew, and Fillia Makedon.
\newblock A survey of robots in healthcare.
\newblock \emph{Technologies}, 9, 2021.

\bibitem[Liao et~al.(2023)Liao, Xu, Lin, Ren, Wei, and Zhao]{cylinpainting2023}
Kang Liao, Xiangyu Xu, Chunyu Lin, Wenqi Ren, Yunchao Wei, and Yao Zhao.
\newblock Cylin-painting: Seamless 360° panoramic image outpainting and beyond.
\newblock \emph{IEEE Transactions on Image Processing}, 2023.

\bibitem[Lipson et~al.(2021)Lipson, Teed, and Deng]{raftstereo}
Lahav Lipson, Zachary Teed, and Jia Deng.
\newblock {RAFT-Stereo}: Multilevel recurrent field transforms for stereo matching.
\newblock In \emph{Proceedings of the International Conference on 3D Vision (3DV)}, 2021.

\bibitem[Liu et~al.(2024)Liu, Chen, and Fan]{li2024roadformer}
Chuang-Wei Liu, Qijun Chen, and Rui Fan.
\newblock Playing to vision foundation model's strengths in stereo matching.
\newblock \emph{IEEE Transactions on Intelligent Vehicles}, 2024.

\bibitem[Martin-Martin et~al.(2021)Martin-Martin, Patel, Rezatofighi, Shenoi, Gwak, Frankel, Sadeghian, and Savarese]{JRDB}
Roberto Martin-Martin, Mihir Patel, Hamid Rezatofighi, Abhijeet Shenoi, JunYoung Gwak, Eric Frankel, Amir Sadeghian, and Silvio Savarese.
\newblock Jrdb: A dataset and benchmark of egocentric robot visual perception of humans in built environments.
\newblock \emph{IEEE Transactions on Pattern Analysis and Machine Intelligence (PAMI)}, 2021.

\bibitem[Mayer et~al.(2016)Mayer, Ilg, H{\"a}usser, Fischer, Cremers, Dosovitskiy, and Brox]{Sceneflow}
N. Mayer, E. Ilg, P. H{\"a}usser, P. Fischer, D. Cremers, A. Dosovitskiy, and T. Brox.
\newblock A large dataset to train convolutional networks for disparity, optical flow, and scene flow estimation.
\newblock In \emph{Proceedings of the IEEE Conference on Computer Vision and Pattern Recognition (CVPR)}, 2016.

\bibitem[Menze and Geiger(2015)]{Menze2015CVPR}
Moritz Menze and Andreas Geiger.
\newblock Object scene flow for autonomous vehicles.
\newblock In \emph{Proceedings of the IEEE Conference on Computer Vision and Pattern Recognition (CVPR)}, 2015.

\bibitem[Meuleman et~al.(2021)Meuleman, Jang, Jeon, and Kim]{Meuleman_2021_CVPR}
Andreas Meuleman, Hyeonjoong Jang, Daniel~S. Jeon, and Min~H. Kim.
\newblock Real-time sphere sweeping stereo from multiview fisheye images.
\newblock In \emph{Proceedings of the IEEE/CVF Conference on Computer Vision and Pattern Recognition (CVPR)}, 2021.

\bibitem[Rey-Area et~al.(2022)Rey-Area, Yuan, and Richardt]{reyarea2021360monodepth}
Manuel Rey-Area, Mingze Yuan, and Christian Richardt.
\newblock {360MonoDepth}: High-resolution 360° monocular depth estimation.
\newblock In \emph{Proceedings of the IEEE/CVF Conference on Computer Vision and Pattern Recognition (CVPR)}, 2022.

\bibitem[Shen et~al.(2023)Shen, Zhang, Tian, Chen, and Sherony]{Shen2023AnEP}
Dan Shen, Zhengming Zhang, Renran Tian, Yaobin Chen, and Rini Sherony.
\newblock An efficient probabilistic solution to mapping errors in lidar-camera fusion for autonomous vehicles, 2023.

\bibitem[Shen et~al.(2022)Shen, Lin, Liao, Nie, Zheng, and Zhao]{shen2022panoformer}
Zhijie Shen, Chunyu Lin, Kang Liao, Lang Nie, Zishuo Zheng, and Yao Zhao.
\newblock Panoformer: Panorama transformer for indoor 360° depth estimation.
\newblock In \emph{European Conference on Computer Vision (ECCV)}, 2022.

\bibitem[Song et~al.(2017)Song, Yu, Zeng, Chang, Savva, and Funkhouser]{song2016ssc}
Shuran Song, Fisher Yu, Andy Zeng, Angel~X Chang, Manolis Savva, and Thomas Funkhouser.
\newblock Semantic scene completion from a single depth image.
\newblock In \emph{Proceedings of the IEEE/CVF Conference on Computer Vision and Pattern Recognition (CVPR)}, 2017.

\bibitem[Sun et~al.(2020)Sun, Kretzschmar, Dotiwalla, Chouard, Patnaik, Tsui, Guo, Zhou, Chai, Caine, et~al.]{waymo}
Pei Sun, Henrik Kretzschmar, Xerxes Dotiwalla, Aurelien Chouard, Vijaysai Patnaik, Paul Tsui, James Guo, Yin Zhou, Yuning Chai, Benjamin Caine, et~al.
\newblock Scalability in perception for autonomous driving: Waymo open dataset.
\newblock In \emph{Proceedings of the IEEE/CVF Conference on Computer Vision and Pattern Recognition (CVPR)}, 2020.

\bibitem[Vizzo et~al.(2023)Vizzo, Guadagnino, Mersch, Wiesmann, Behley, and Stachniss]{vizzo2023kiss}
Ignacio Vizzo, Tiziano Guadagnino, Benedikt Mersch, Louis Wiesmann, Jens Behley, and Cyrill Stachniss.
\newblock Kiss-icp: In defense of point-to-point icp--simple, accurate, and robust registration if done the right way.
\newblock \emph{IEEE Robotics and Automation Letters}, 8\penalty0 (2):\penalty0 1029--1036, 2023.

\bibitem[Wang et~al.(2020)Wang, andYi Hsuan~Tsai, Chiu, and Sun]{wang20icra}
Ning-Hsu Wang, Bolivar~Solarte andYi Hsuan~Tsai, Wei-Chen Chiu, and Min Sun.
\newblock {360SD-Net}: 360° stereo depth estimation with learnable cost volume.
\newblock In \emph{International Conference on Robotics and Automation (ICRA)}, 2020.

\bibitem[Weinzaepfel et~al.(2023)Weinzaepfel, Lucas, Leroy, Cabon, Arora, Br{\'e}gier, Csurka, Antsfeld, Chidlovskii, and Revaud]{weinzaepfel2023croco}
Philippe Weinzaepfel, Thomas Lucas, Vincent Leroy, Yohann Cabon, Vaibhav Arora, Romain Br{\'e}gier, Gabriela Csurka, Leonid Antsfeld, Boris Chidlovskii, and J{\'e}r{\^o}me Revaud.
\newblock Croco v2: Improved cross-view completion pre-training for stereo matching and optical flow.
\newblock In \emph{Proceedings of the IEEE/CVF International Conference on Computer Vision (ICCV)}, 2023.

\bibitem[Won et~al.(2019)Won, Ryu, and Lim]{Won2019SweepNetWO}
Changhee Won, Jongbin Ryu, and Jongwoo Lim.
\newblock Sweepnet: Wide-baseline omnidirectional depth estimation.
\newblock In \emph{International Conference on Robotics and Automation (ICRA)}, 2019.

\bibitem[Won et~al.(2020)Won, Ryu, and Lim]{won2020end}
Changhee Won, Jongbin Ryu, and Jongwoo Lim.
\newblock End-to-end learning for omnidirectional stereo matching with uncertainty prior.
\newblock \emph{IEEE Transactions on Pattern Analysis and Machine Intelligence (PAMI)}, 2020.

\bibitem[Xu et~al.(2023)Xu, Wang, Ding, and Yang]{xu2023iterative}
Gangwei Xu, Xianqi Wang, Xiaohuan Ding, and Xin Yang.
\newblock Iterative geometry encoding volume for stereo matching.
\newblock In \emph{Proceedings of the IEEE/CVF Conference on Computer Vision and Pattern Recognition (CVPR)}, 2023.

\bibitem[Yang et~al.(2019)Yang, Song, Huang, Deng, Shi, and Zhou]{drivingstereo}
Guorun Yang, Xiao Song, Chaoqin Huang, Zhidong Deng, Jianping Shi, and Bolei Zhou.
\newblock Drivingstereo: A large-scale dataset for stereo matching in autonomous driving scenarios.
\newblock In \emph{Proceedings of the IEEE/CVF Conference on Computer Vision and Pattern Recognition (CVPR)}, 2019.

\bibitem[Yang et~al.(2020)Yang, Hu, Chen, Xiang, Wang, and Stiefelhagen]{kailun2020}
Kailun Yang, Xinxin Hu, Hao Chen, Kaite Xiang, Kaiwei Wang, and Rainer Stiefelhagen.
\newblock Ds-pass: Detail-sensitive panoramic annular semantic segmentation through swaftnet for surrounding sensing.
\newblock In \emph{IEEE Intelligent Vehicles Symposium (IV)}, 2020.

\bibitem[Yang et~al.(2024)Yang, Kang, Huang, Xu, Feng, and Zhao]{depthanything2024}
Lihe Yang, Bingyi Kang, Zilong Huang, Xiaogang Xu, Jiashi Feng, and Hengshuang Zhao.
\newblock Depth anything: Unleashing the power of large-scale unlabeled data.
\newblock In \emph{Proceedings of the IEEE/CVF Conference on Computer Vision and Pattern Recognition (CVPR)}, 2024.

\bibitem[Zbontar and LeCun(2016)]{zbontar2016stereo}
Jure Zbontar and Yann LeCun.
\newblock Stereo matching by training a convolutional neural network to compare image patches.
\newblock \emph{Journal of Machine Learning Research (JMLR)}, 2016.

\bibitem[Zeng et~al.(2024)Zeng, Yao, Wu, and Jia]{zeng2024temporally}
Jiaxi Zeng, Chengtang Yao, Yuwei Wu, and Yunde Jia.
\newblock Temporally consistent stereo matching.
\newblock In \emph{European Conference on Computer Vision (ECCV)}, 2024.

\bibitem[Zhang et~al.(2019)Zhang, Liwicki, Smith, and Cipolla]{zhang2019orientation}
Chao Zhang, Stephan Liwicki, William Smith, and Roberto Cipolla.
\newblock Orientation-aware semantic segmentation on icosahedron spheres.
\newblock In \emph{Proceedings of the IEEE International Conference on Computer Vision (ICCV)}, pages 3533--3541, 2019.

\bibitem[Zheng et~al.(2020)Zheng, Zhang, Li, Tang, Gao, and Zhou]{zheng2020structured3d}
Jia Zheng, Junfei Zhang, Jing Li, Rui Tang, Shenghua Gao, and Zihan Zhou.
\newblock Structured3d: A large photo-realistic dataset for structured 3d modeling.
\newblock In \emph{European Conference on Computer Vision (ECCV)}, 2020.

\end{thebibliography}
}

\clearpage
\setcounter{page}{1}
\maketitlesupplementary

\appendix

%=================================================================================%
\section{Dataset Specifications}
\label{appx_sec:details_dataset}
%=================================================================================%

This section complements the overview of the dataset by detailing the video sequences and providing additional histograms for depth and disparity.

%=--------------------------------------------------------------------------------%
\subsection{Sequences}
\label{subsec:sequences}
%=--------------------------------------------------------------------------------%

The \textsc{Helvipad} dataset includes 26 video sequences captured between December 2023 and February 2024., recorded at a frame rate of 10 Hz. %\textcolor{blue}{The frame rate at which we collected the dataset is of 10Hz}

Each sequence is synchronized with its corresponding LiDAR point clouds, which are projected on frames to obtain depth maps, and subsequently disparity maps. Top images, bottom images, depth maps, and disparity maps are then cropped to remove unnecessary borders and downsized from a width of 1024 pixels to 512 to enable more efficient training.

The sequences span a diverse array of settings and conditions, as detailed in \cref{appx_tab:seq}. The dataset includes recordings taken at various times of the day, from early morning to night, under a variety of weather conditions, including cloudy and sunny skies. These recordings were made in multiple indoor and outdoor locations, from pedestrian squares and footpaths to corridors and parking areas, offering a wide spectrum of environmental contexts. The table also provides information on the duration of each sequence, with an average of 2 minutes and 41 seconds. Furthermore, the dynamic nature of the recorded scenes is emphasized by the presence of pedestrians, with an average of 13.33 pedestrians in indoor sequences and 17.65 pedestrians in outdoor sequences.

%=--------------------------------------------------------------------------------%
\subsection{Additional Histograms of Depth and Disparity Labels}
\label{appx_subsec:histograms}
%=--------------------------------------------------------------------------------%

\begin{figure*}[t]
\centering % Centers the entire figure
\resizebox{\textwidth}{!}{%
    % Row 1 - Train plots
    \begin{subfigure}[t]{0.45\textwidth}
        \includegraphics[width=\linewidth]{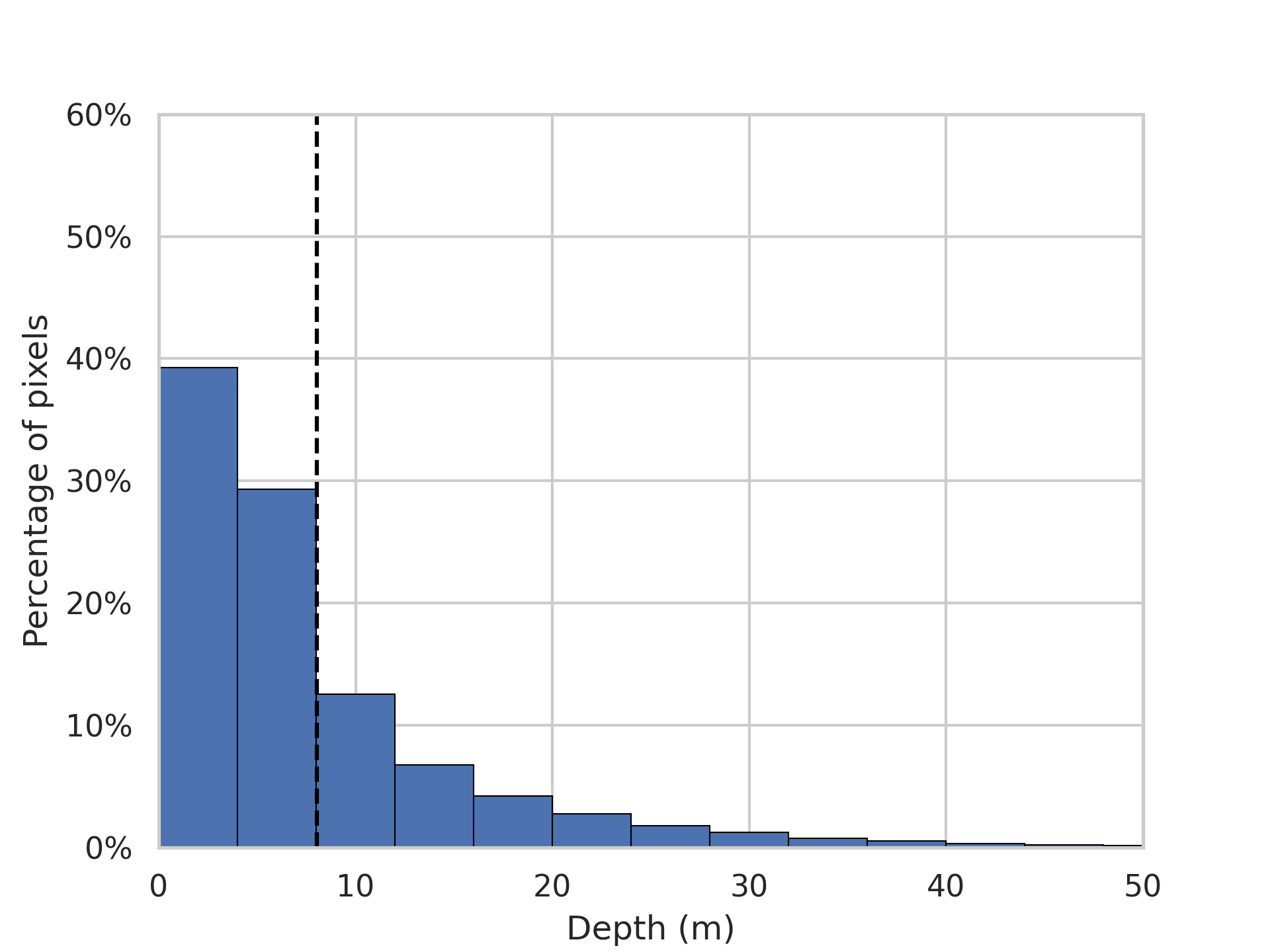}
        \caption{All train sequences - Depth}
        \label{appx_fig:depth_all_train}
    \end{subfigure}%
    \hfill
    \begin{subfigure}[t]{0.45\textwidth}
        \includegraphics[width=\linewidth]{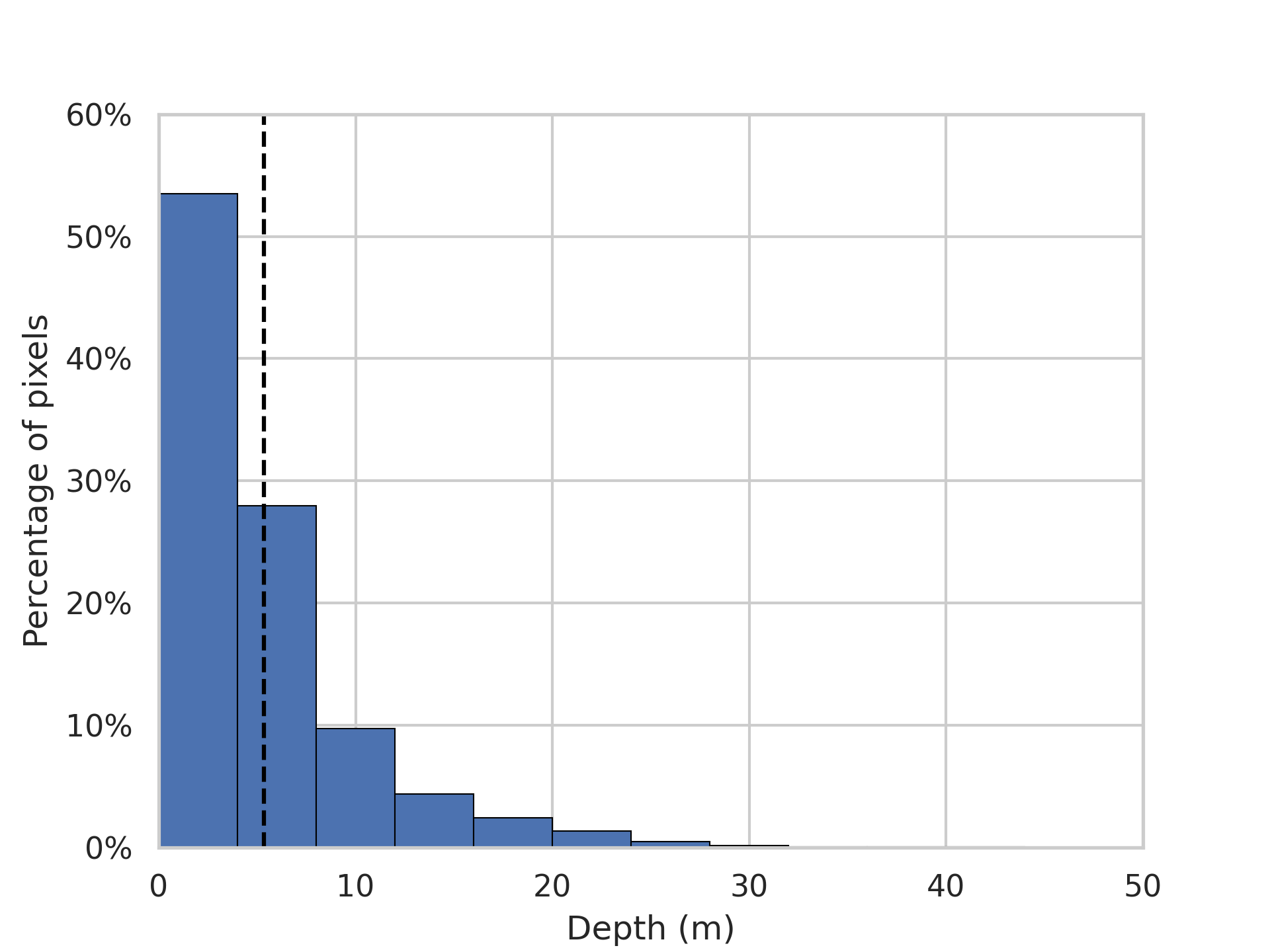}
        \caption{Train indoor sequences - Depth}
        \label{appx_fig:depth_indoor_train}
    \end{subfigure}%
    \hfill
    \begin{subfigure}[t]{0.45\textwidth}
        \includegraphics[width=\linewidth]{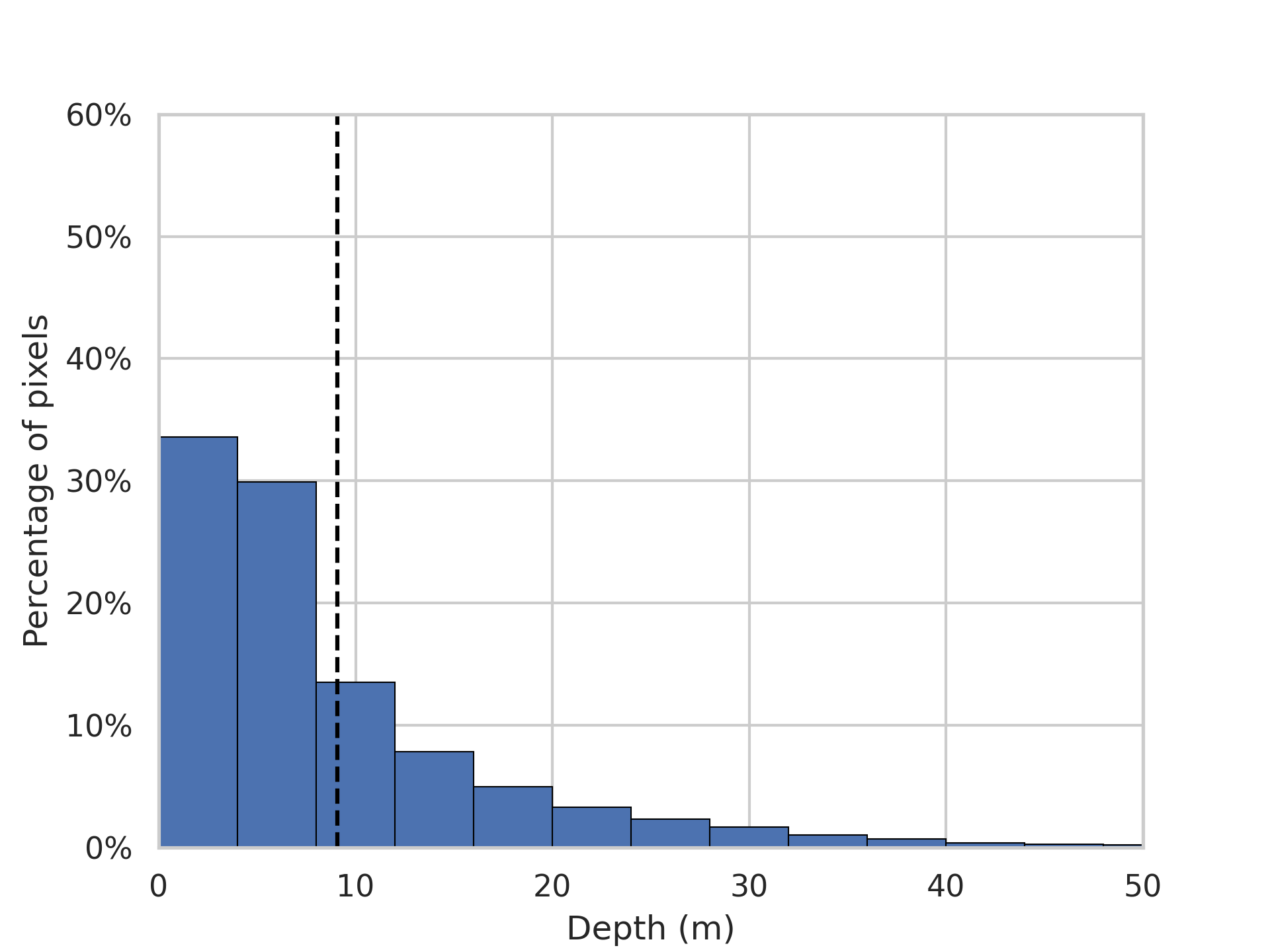}
        \caption{Train outdoor sequences - Depth}
        \label{appx_fig:depth_outdoor_train}
    \end{subfigure}%
    \hfill
    \begin{subfigure}[t]{0.45\textwidth}
        \includegraphics[width=\linewidth]{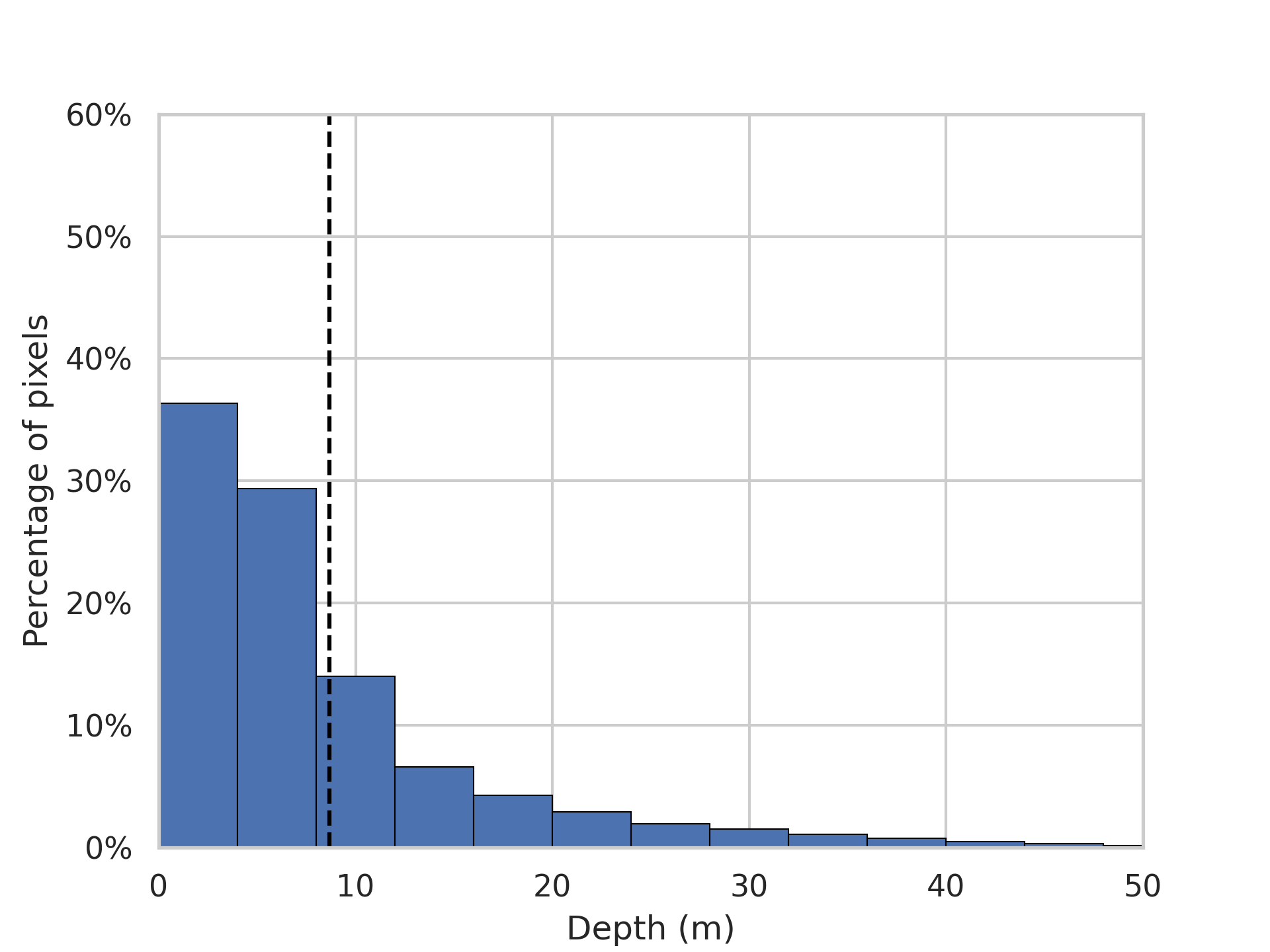}
        \caption{Train night outdoor sequences - Depth}
        \label{appx_fig:depth_night_outdoor_train}
    \end{subfigure}
}

\vspace{0.3cm} % Adds some spacing between rows

\resizebox{\textwidth}{!}{%
    % Row 2 - Test plots
    \begin{subfigure}[t]{0.45\textwidth}
        \includegraphics[width=\linewidth]{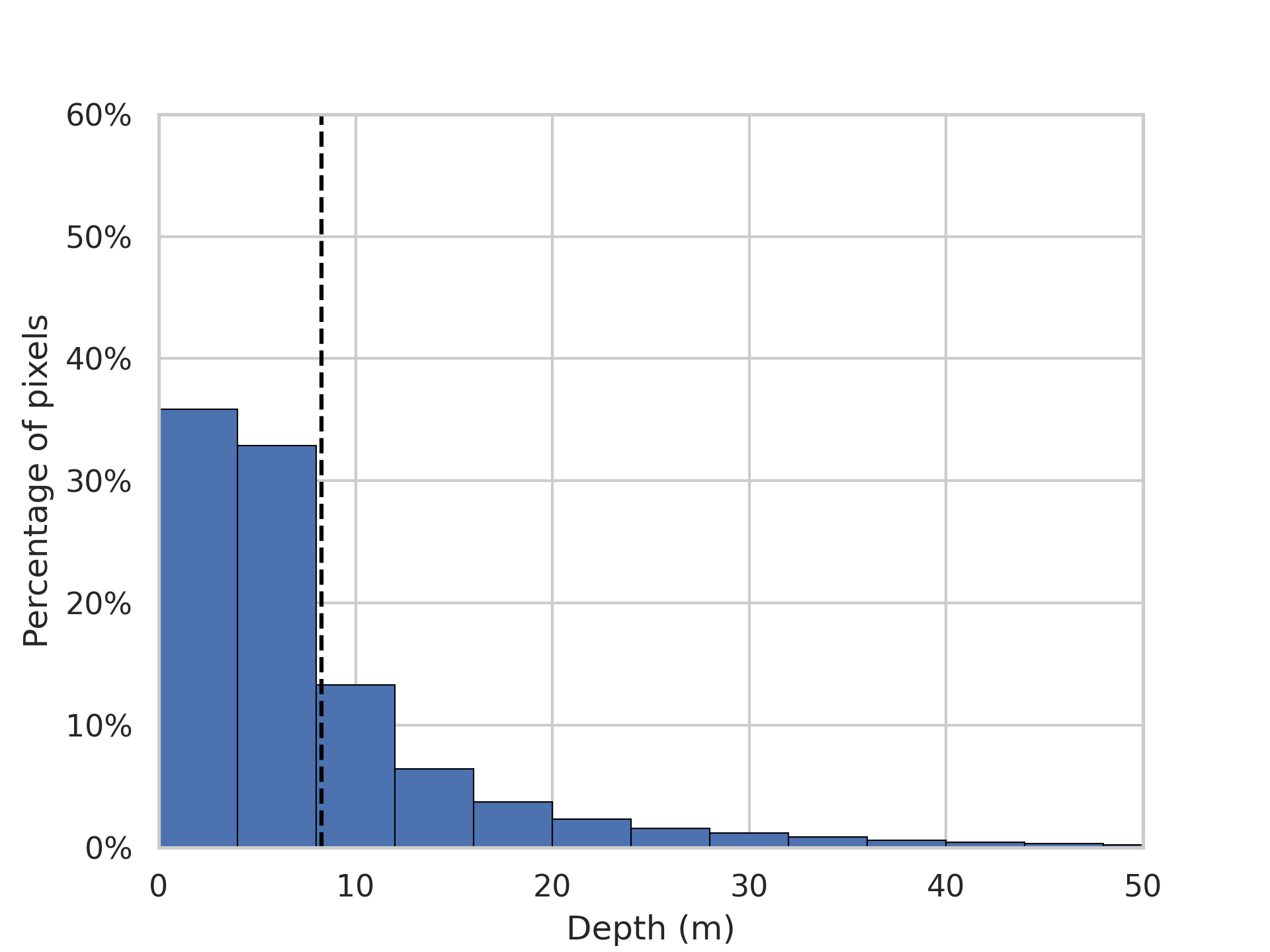}
        \caption{All test sequences - Depth}
        \label{appx_fig:depth_all_test}
    \end{subfigure}%
    \hfill
    \begin{subfigure}[t]{0.45\textwidth}
        \includegraphics[width=\linewidth]{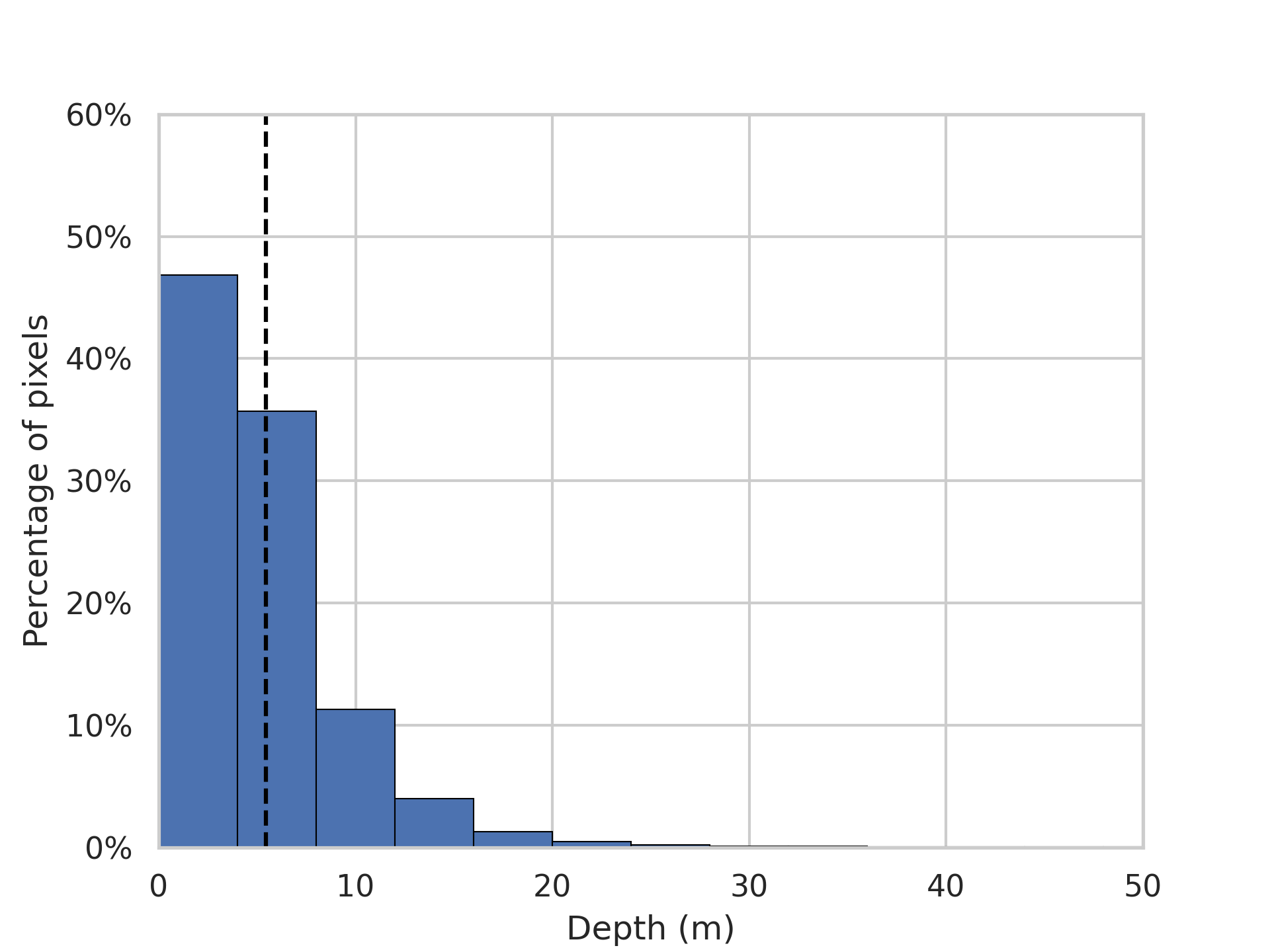}
        \caption{Test indoor sequences - Depth}
        \label{appx_fig:depth_indoor_test}
    \end{subfigure}%
    \hfill
    \begin{subfigure}[t]{0.45\textwidth}
        \includegraphics[width=\linewidth]{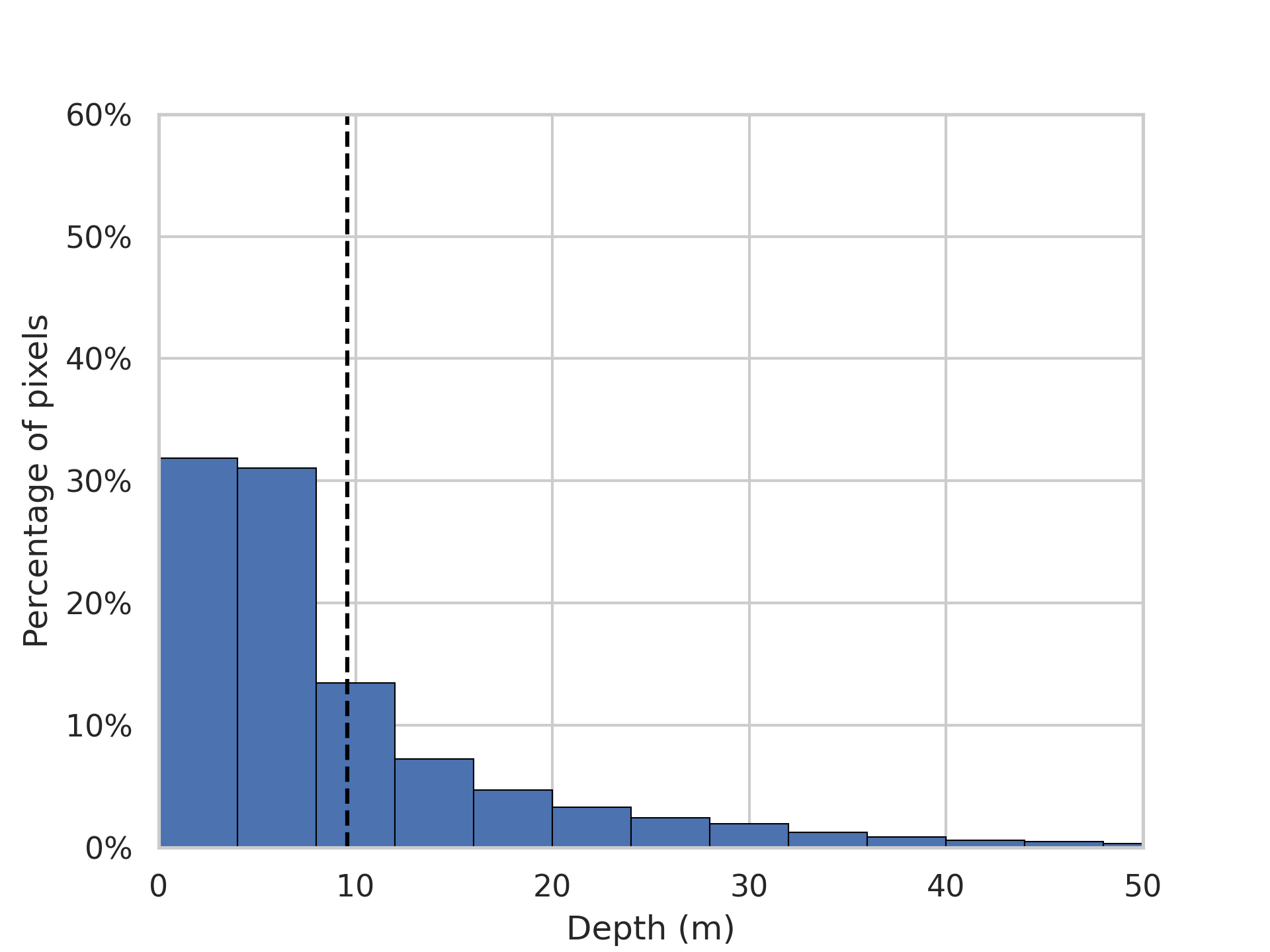}
        \caption{Test outdoor sequences - Depth}
        \label{appx_fig:depth_outdoor_test}
    \end{subfigure}%
    \hfill
    \begin{subfigure}[t]{0.45\textwidth}
        \includegraphics[width=\linewidth]{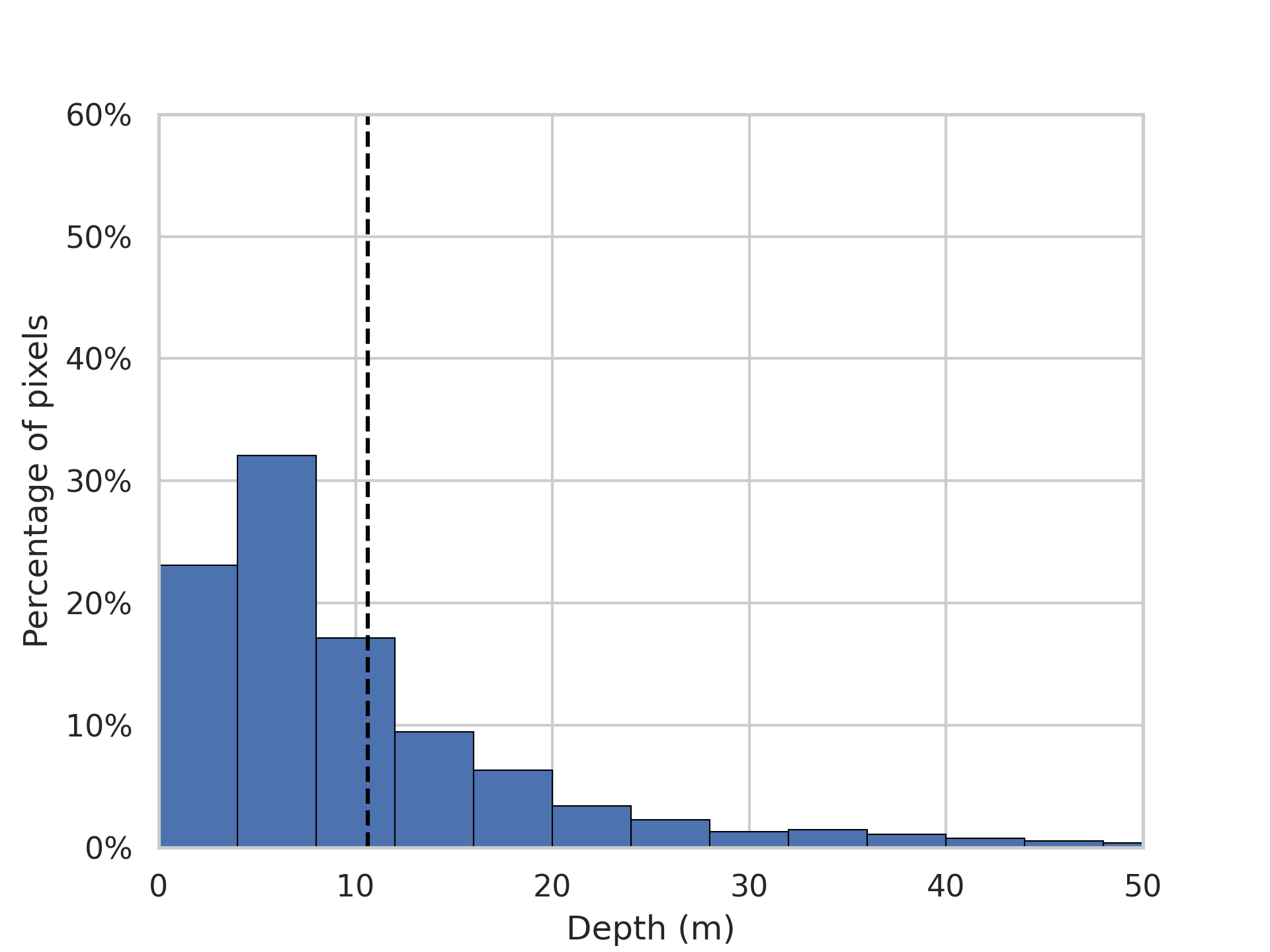}
        \caption{Test night outdoor sequences - Depth}
        \label{appx_fig:depth_night_outdoor_test}
    \end{subfigure}
}

\caption{\textbf{Histograms of depth labels across train} (first row) \textbf{and test splits} (second row). Each plot's vertical dotted line denotes the average depth for the respective setting.}
\label{appx_fig:depth_histograms}
\end{figure*}

\begin{figure*}[ht]
\resizebox{\textwidth}{!}{%
    \centering % Centers the entire figure
    
    % Row 1
    \begin{subfigure}[t]{0.45\textwidth}
        \includegraphics[width=\linewidth]{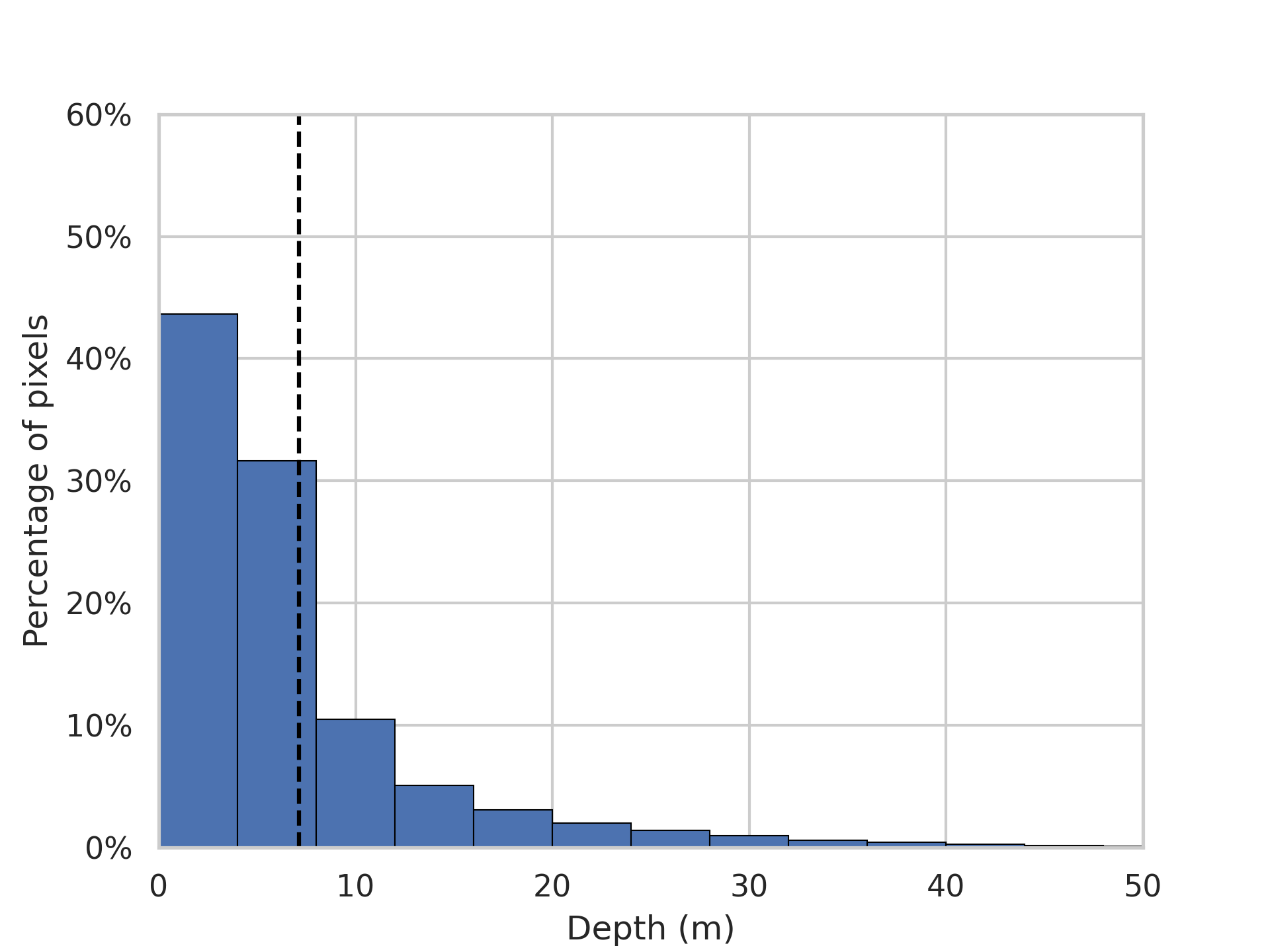}
        \caption{All train sequences - Depth}
        \label{appx_fig:augmented_depth_all_train}
    \end{subfigure}%
    \hfill
    \begin{subfigure}[t]{0.45\textwidth}
        \includegraphics[width=\linewidth]{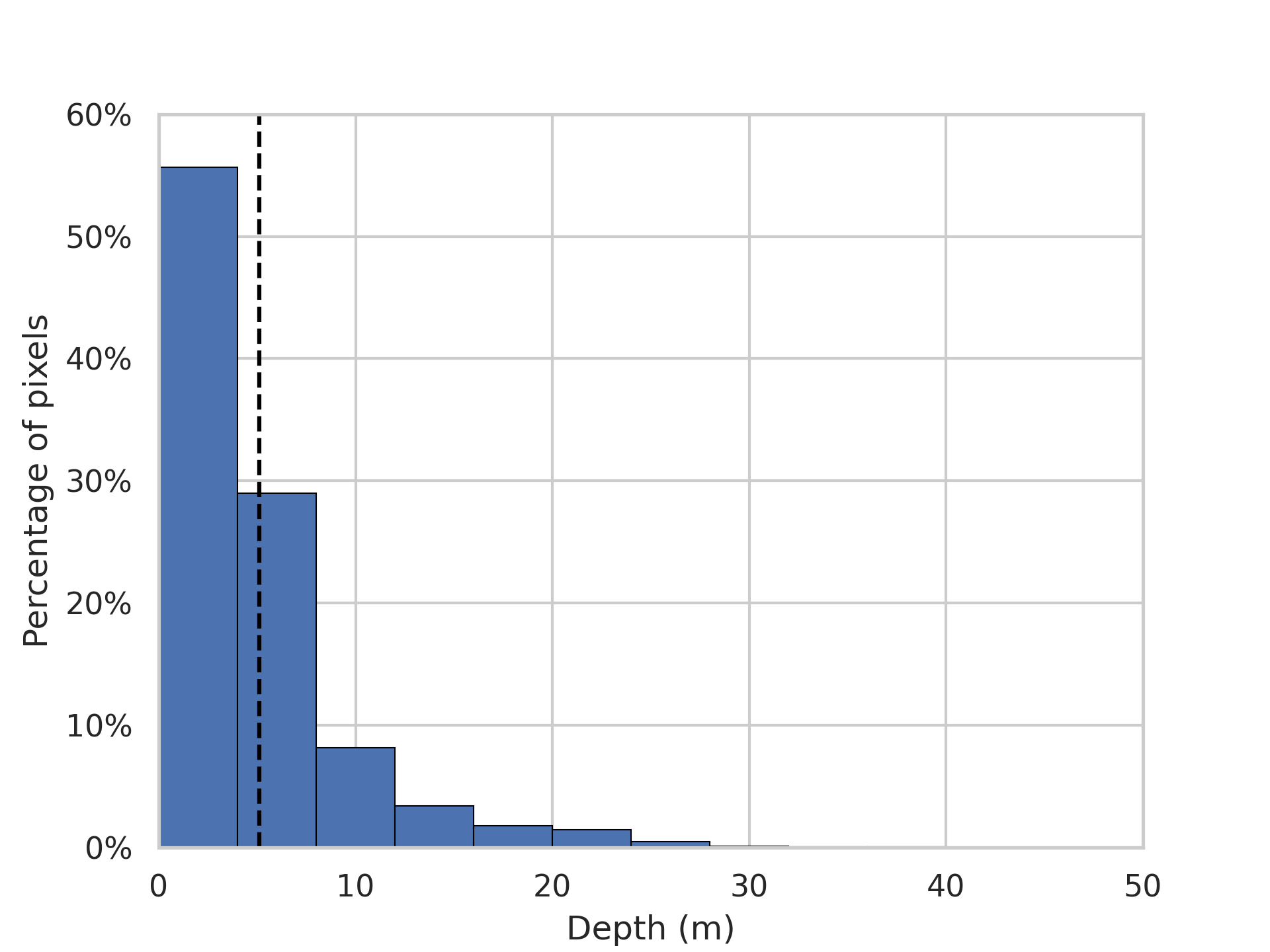}
        \caption{Train indoor sequences - Depth}
        \label{appx_fig:augmented_depth_indoor_train}
    \end{subfigure}%
    \hfill
    \begin{subfigure}[t]{0.45\textwidth}
        \includegraphics[width=\linewidth]{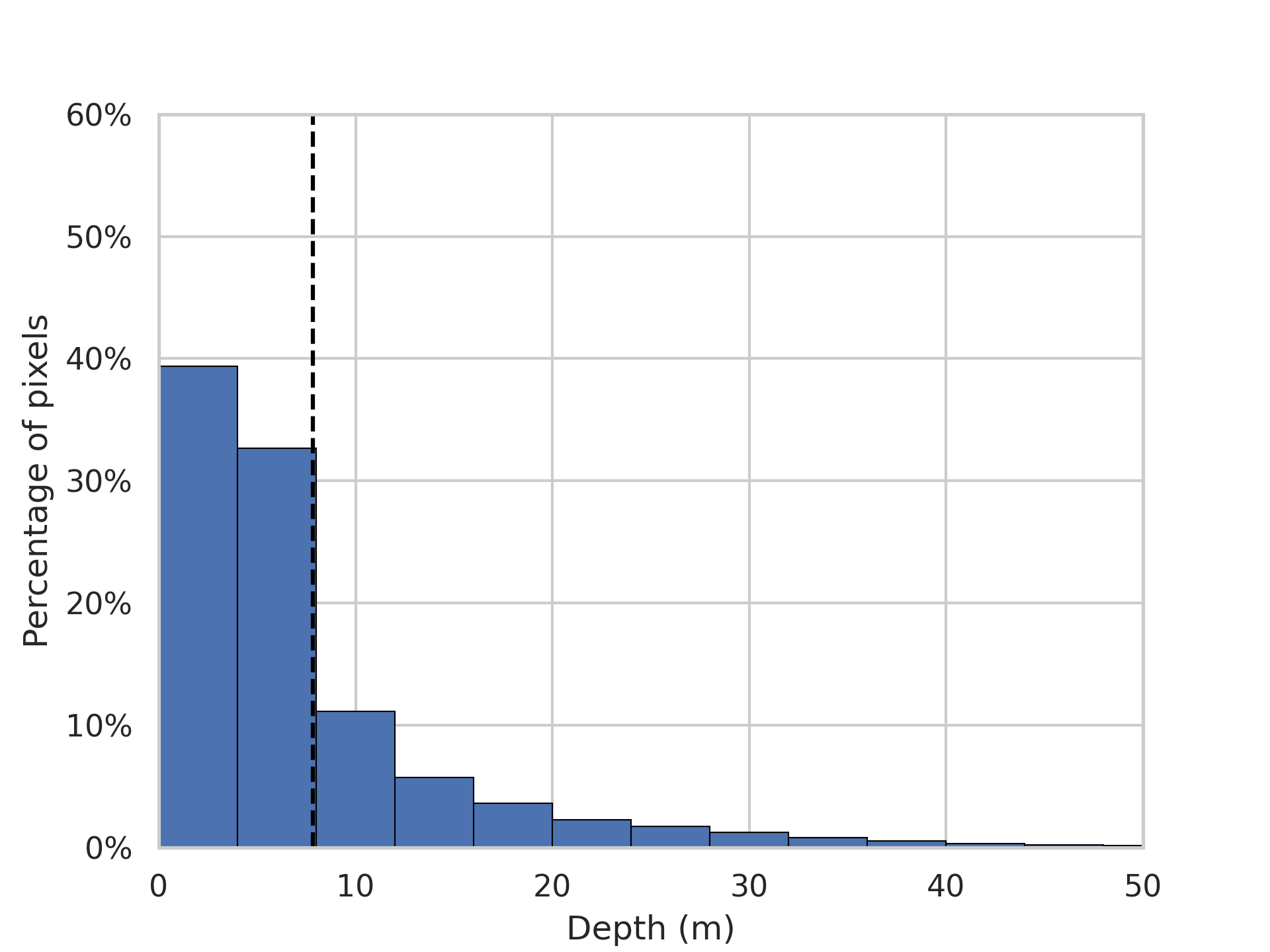}
        \caption{Train outdoor sequences - Depth}
        \label{appx_fig:augmented_depth_outdoor_train}
    \end{subfigure}%
    \hfill
    \begin{subfigure}[t]{0.45\textwidth}
        \includegraphics[width=\linewidth]{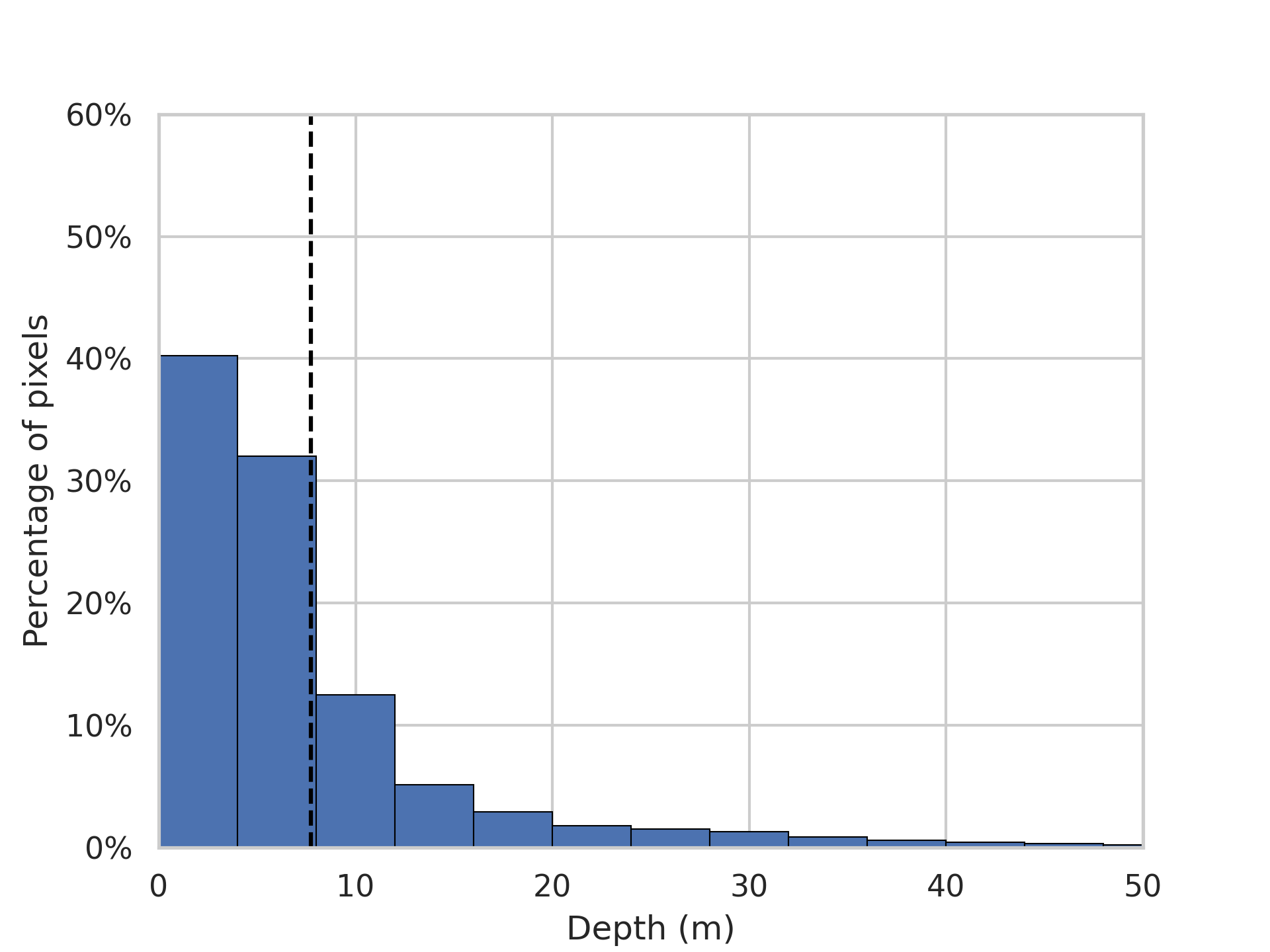}
        \caption{Train night outdoor sequences - Depth}
        \label{appx_fig:augmented_depth_night_outdoor_train}
    \end{subfigure}%
}  
    \caption{\textbf{Histograms of depth labels across across train splits \underline{after depth completion}}. Each plot's vertical dotted line denotes the considered setting average.}
    \label{appx_fig:augmented_depth_histograms}
\end{figure*}

\begin{figure*}[ht]
\centering % Centers the entire figure
\resizebox{\textwidth}{!}{%
    % Row 1 - Train plots
    \begin{subfigure}[t]{0.45\textwidth}
        \includegraphics[width=\linewidth]{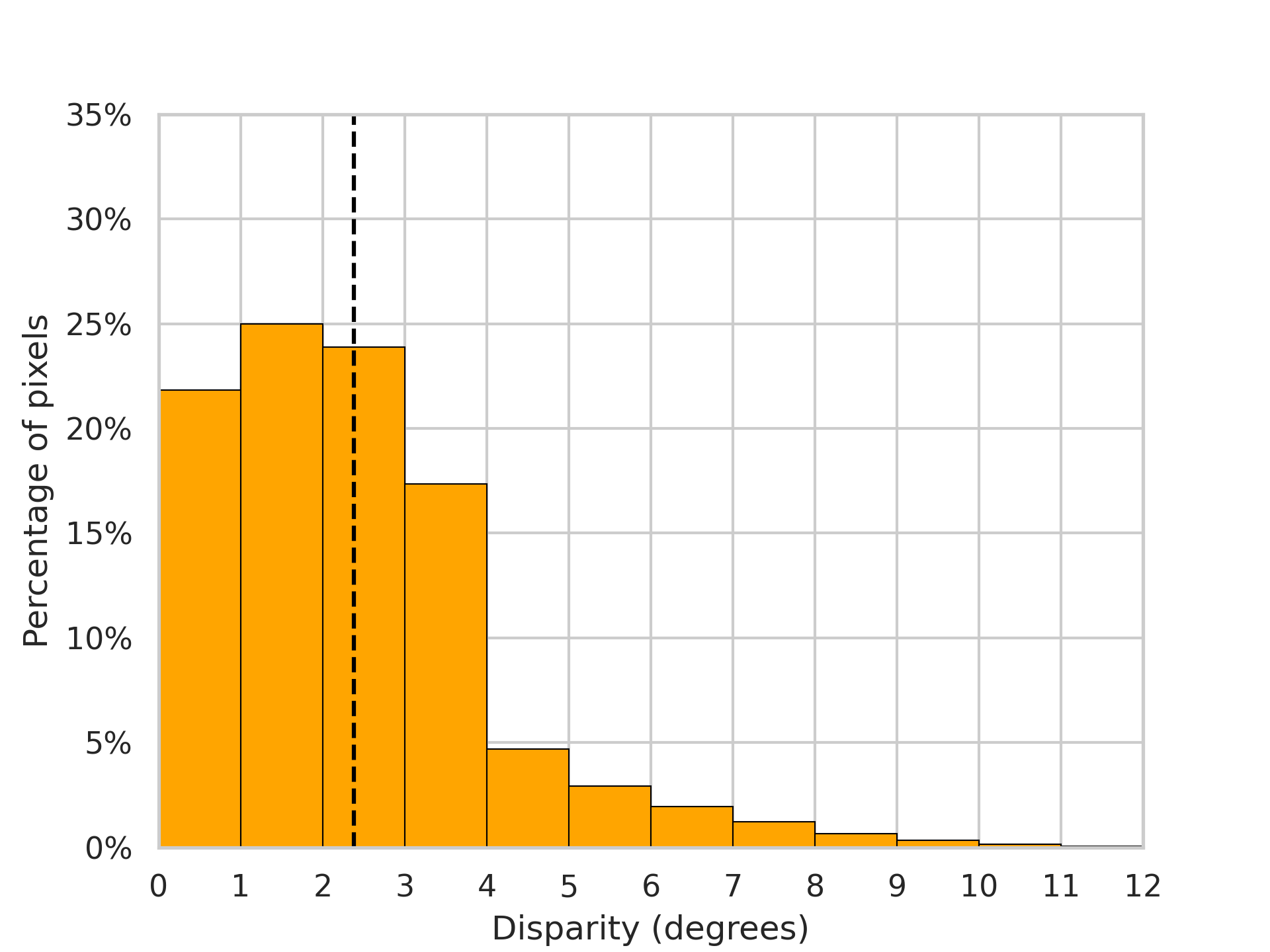}
        \caption{All train sequences - Disparity}
        \label{appx_fig:disp_all_train}
    \end{subfigure}%
    \hfill
    \begin{subfigure}[t]{0.45\textwidth}
        \includegraphics[width=\linewidth]{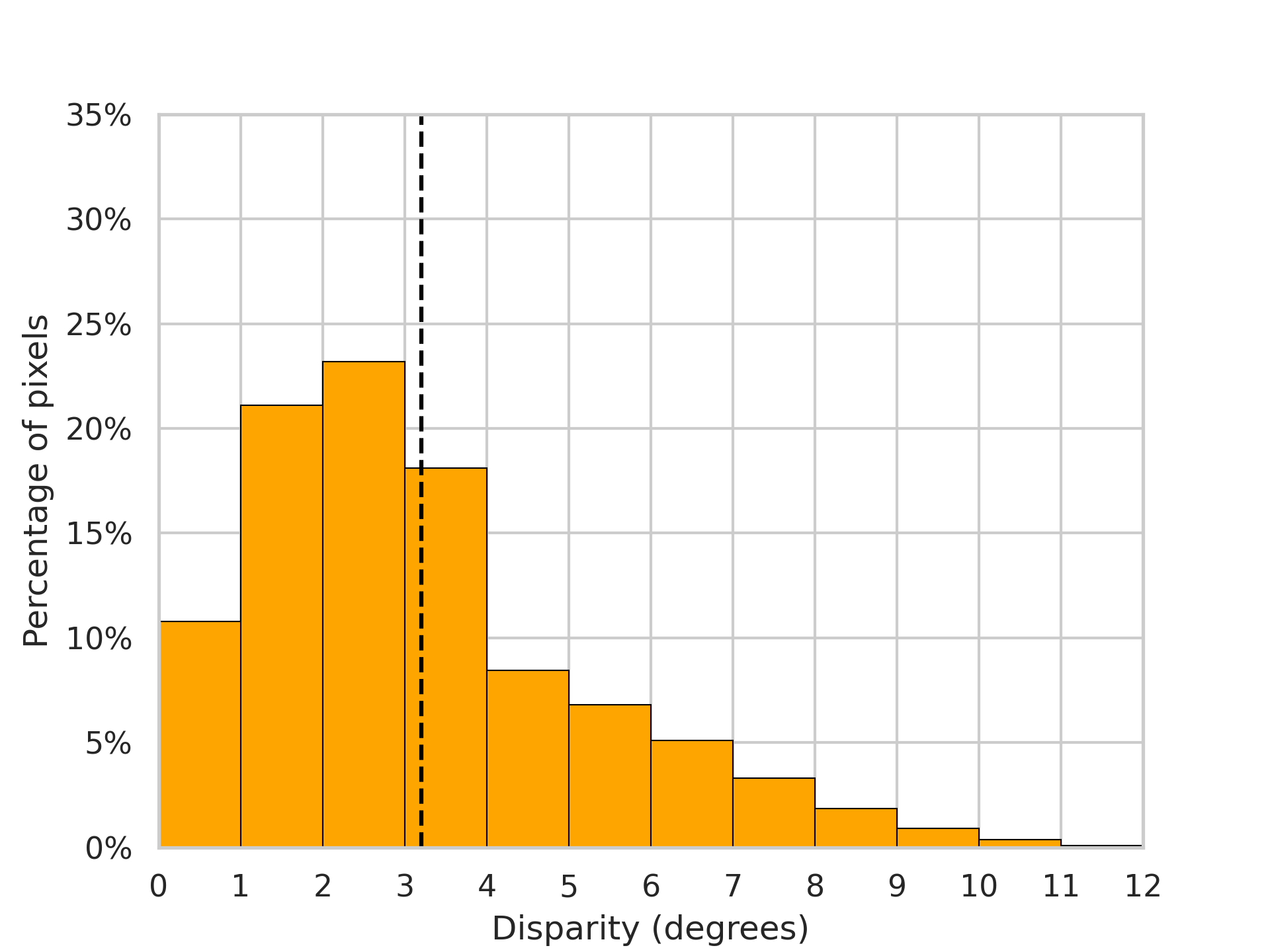}
        \caption{Train indoor sequences - Disparity}
        \label{appx_fig:disp_indoor_train}
    \end{subfigure}%
    \hfill
    \begin{subfigure}[t]{0.45\textwidth}
        \includegraphics[width=\linewidth]{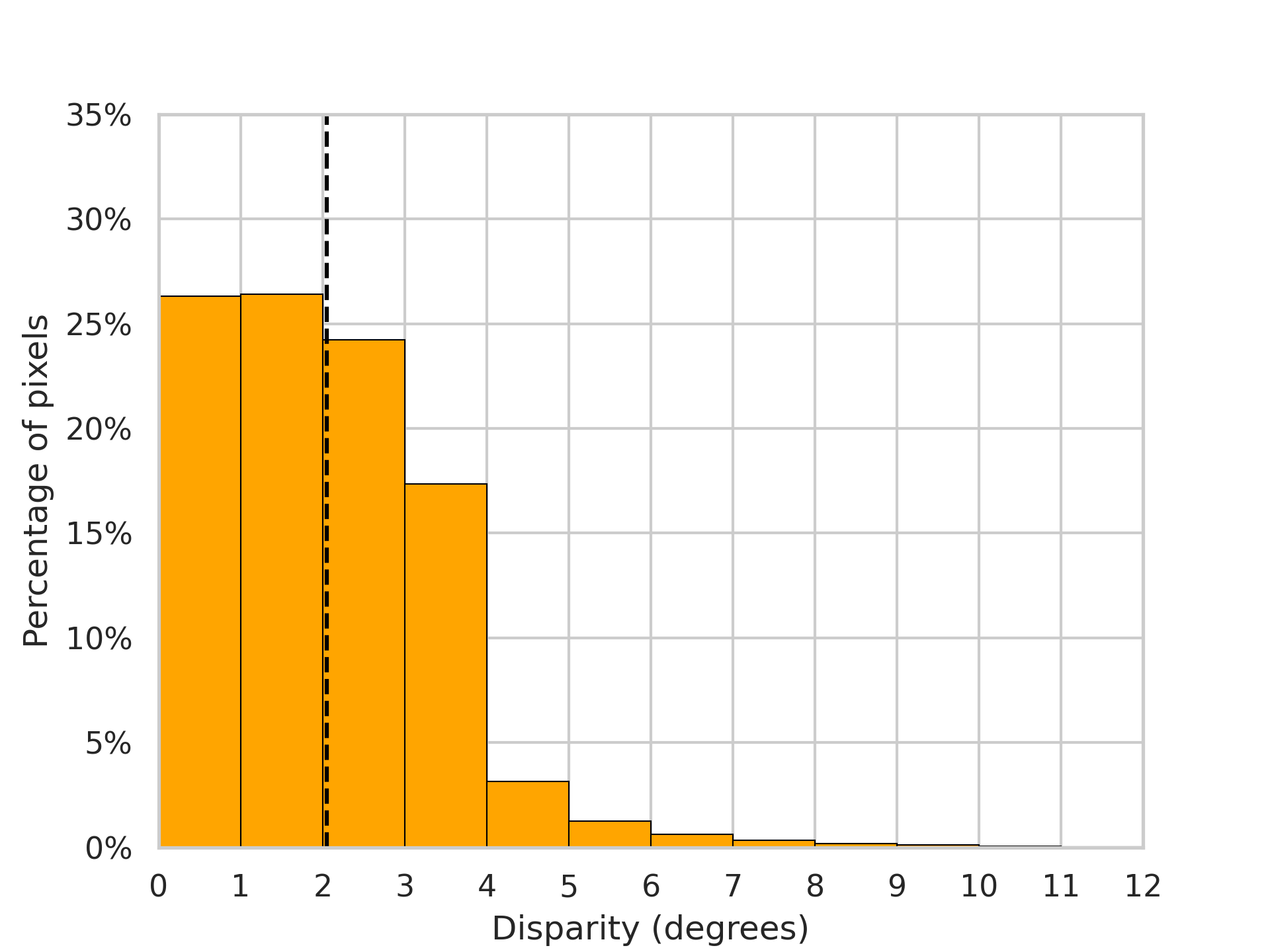}
        \caption{Train outdoor sequences - Disparity}
        \label{appx_fig:disp_outdoor_train}
    \end{subfigure}%
    \hfill
    \begin{subfigure}[t]{0.45\textwidth}
        \includegraphics[width=\linewidth]{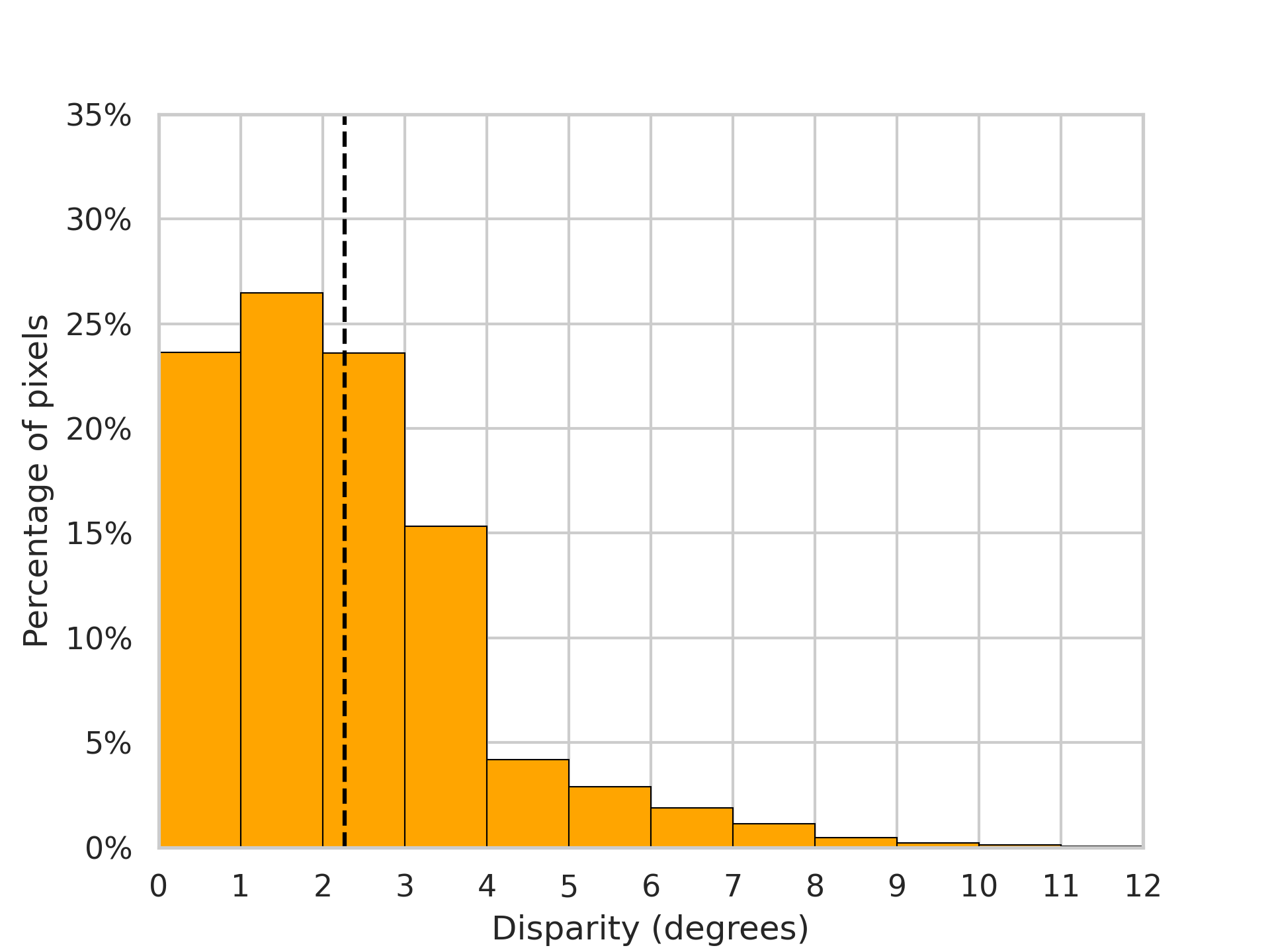}
        \caption{Train night outdoor sequences - Disparity}
        \label{appx_fig:disp_night_outdoor_train}
    \end{subfigure}
}

\vspace{0.3cm} % Adds some spacing between rows

\resizebox{\textwidth}{!}{%
    % Row 2 - Test plots
    \begin{subfigure}[t]{0.45\textwidth}
        \includegraphics[width=\linewidth]{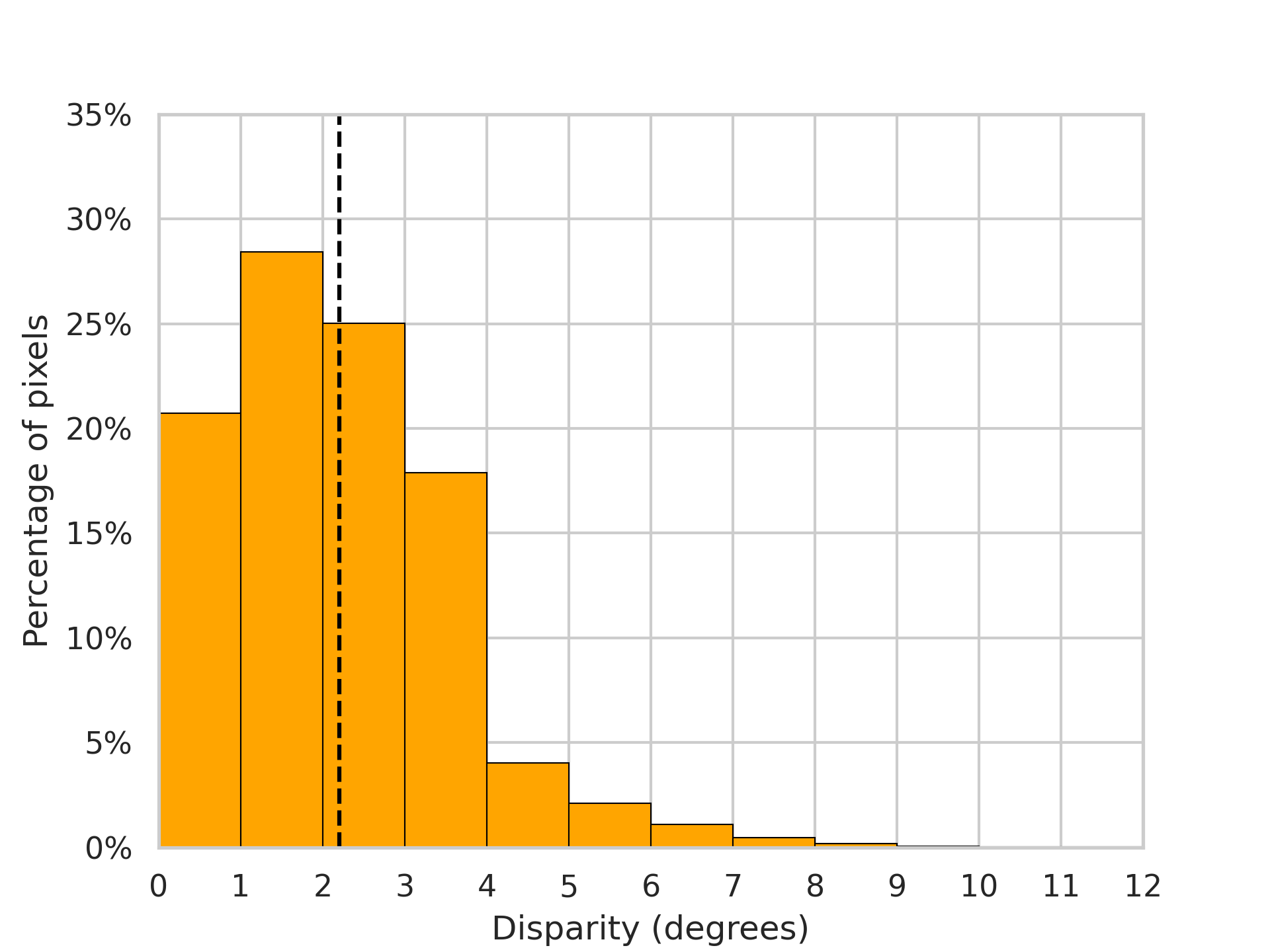}
        \caption{All test sequences - Disparity}
        \label{appx_fig:disp_all_test}
    \end{subfigure}%
    \hfill
    \begin{subfigure}[t]{0.45\textwidth}
        \includegraphics[width=\linewidth]{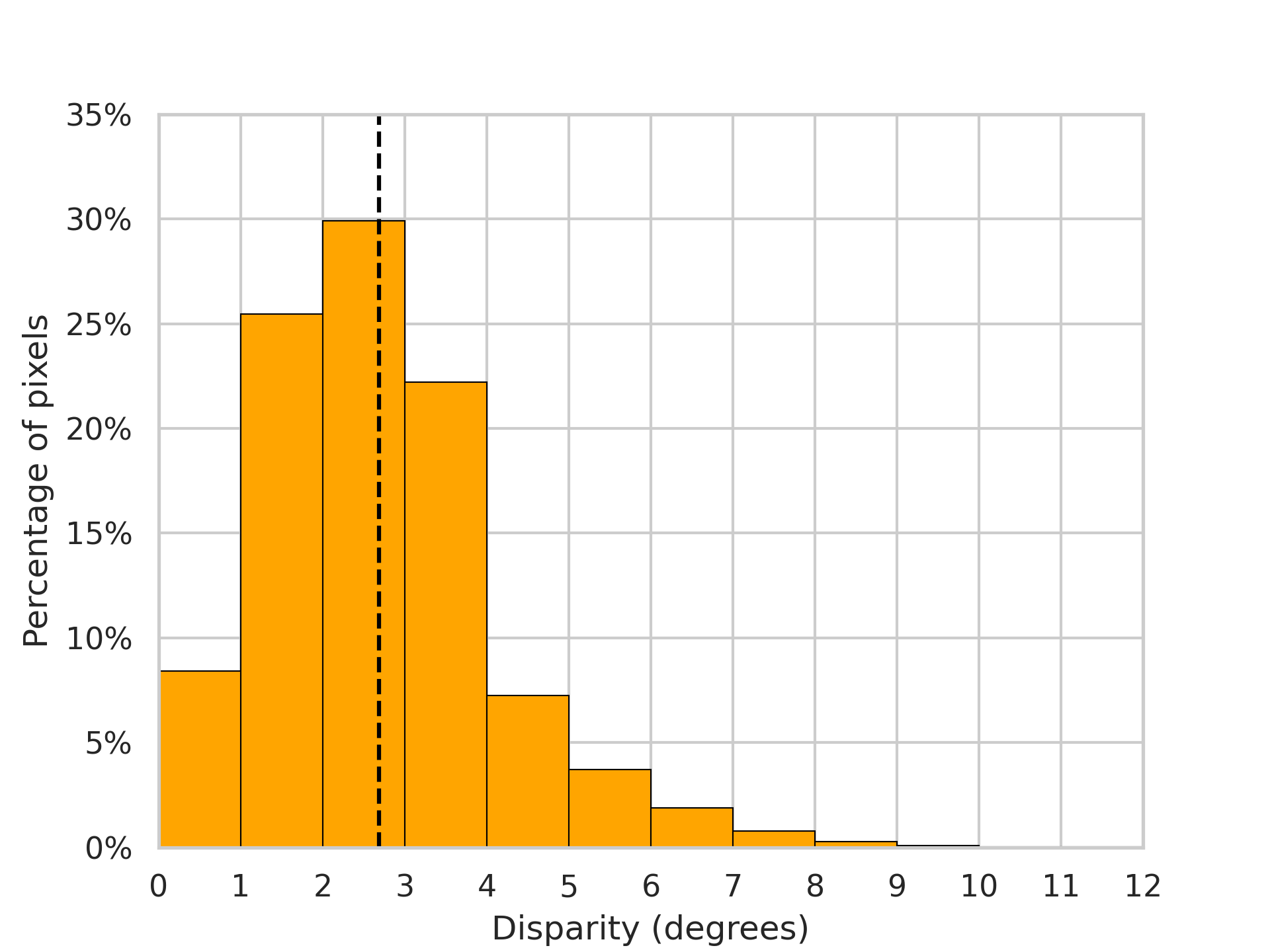}
        \caption{Test indoor sequences - Disparity}
        \label{fig:disp_indoor_test}
    \end{subfigure}%
    \hfill
    \begin{subfigure}[t]{0.45\textwidth}
        \includegraphics[width=\linewidth]{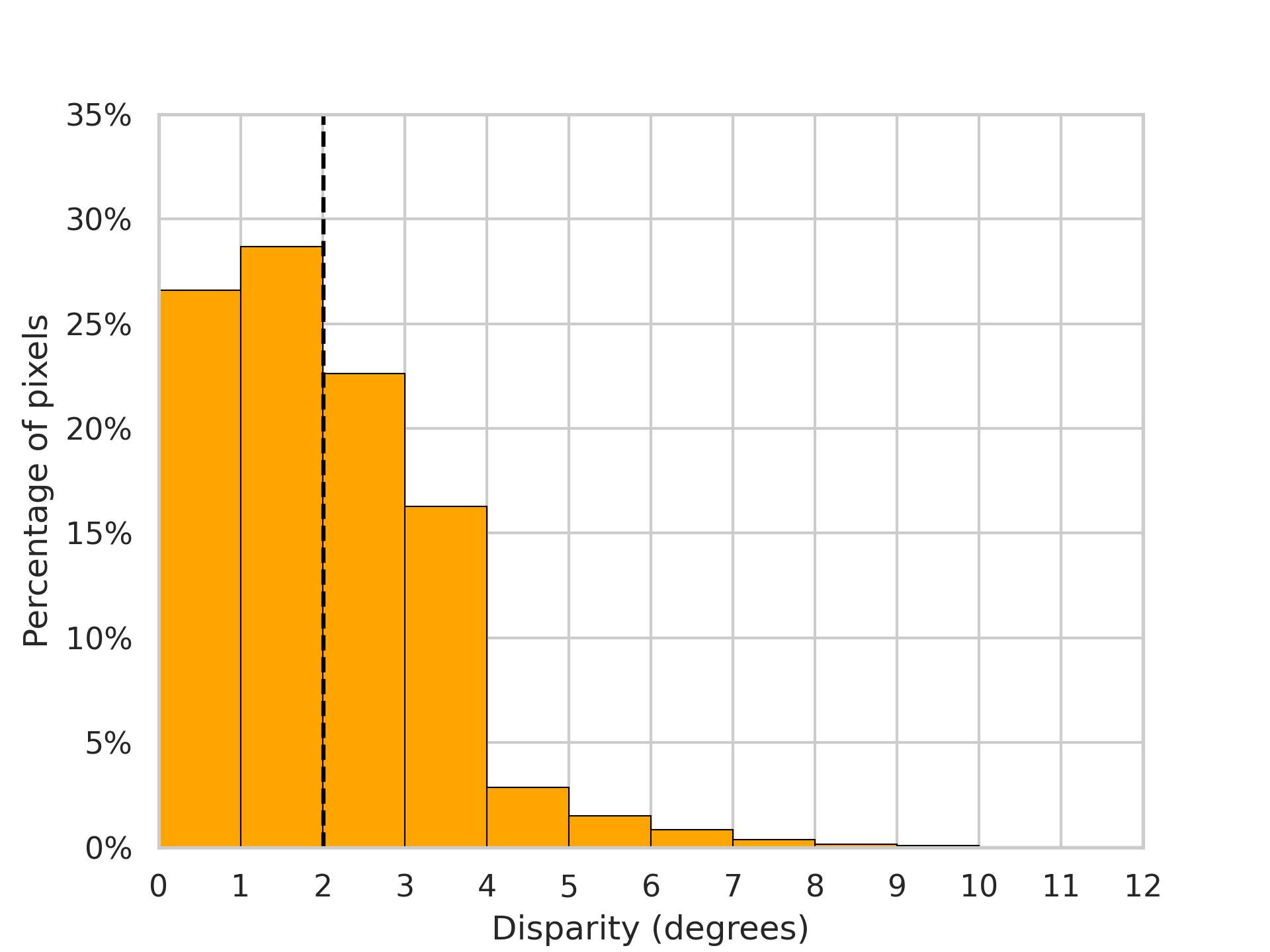}
        \caption{Test outdoor sequences - Disparity}
        \label{appx_fig:disp_outdoor_test}
    \end{subfigure}%
    \hfill
    \begin{subfigure}[t]{0.45\textwidth}
        \includegraphics[width=\linewidth]{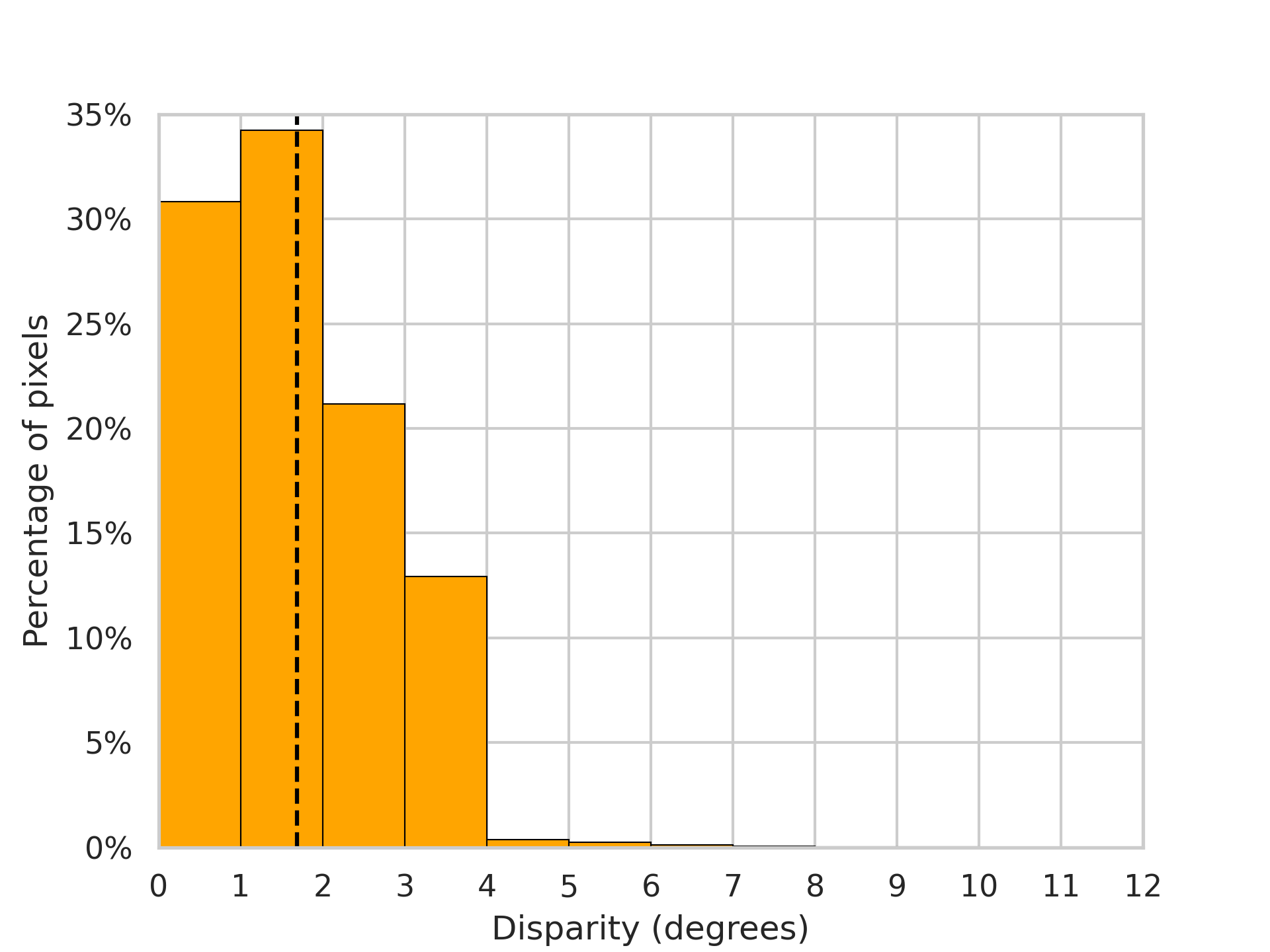}
        \caption{Test night outdoor sequences - Disparity}
        \label{appx_fig:disp_night_outdoor_test}
    \end{subfigure}
}

\caption{\textbf{Histograms of disparity labels  across train} (first row) \textbf{and test splits} (second row). Each plot's vertical dotted line denotes the average disparity for the respective setting.}
\label{appx_fig:disparity_histograms}
\end{figure*}

\begin{figure*}[ht]
\resizebox{\textwidth}{!}{%
    \centering % Centers the entire figure
    
    % Row 1
    \begin{subfigure}[t]{0.45\textwidth}
        \includegraphics[width=\linewidth]{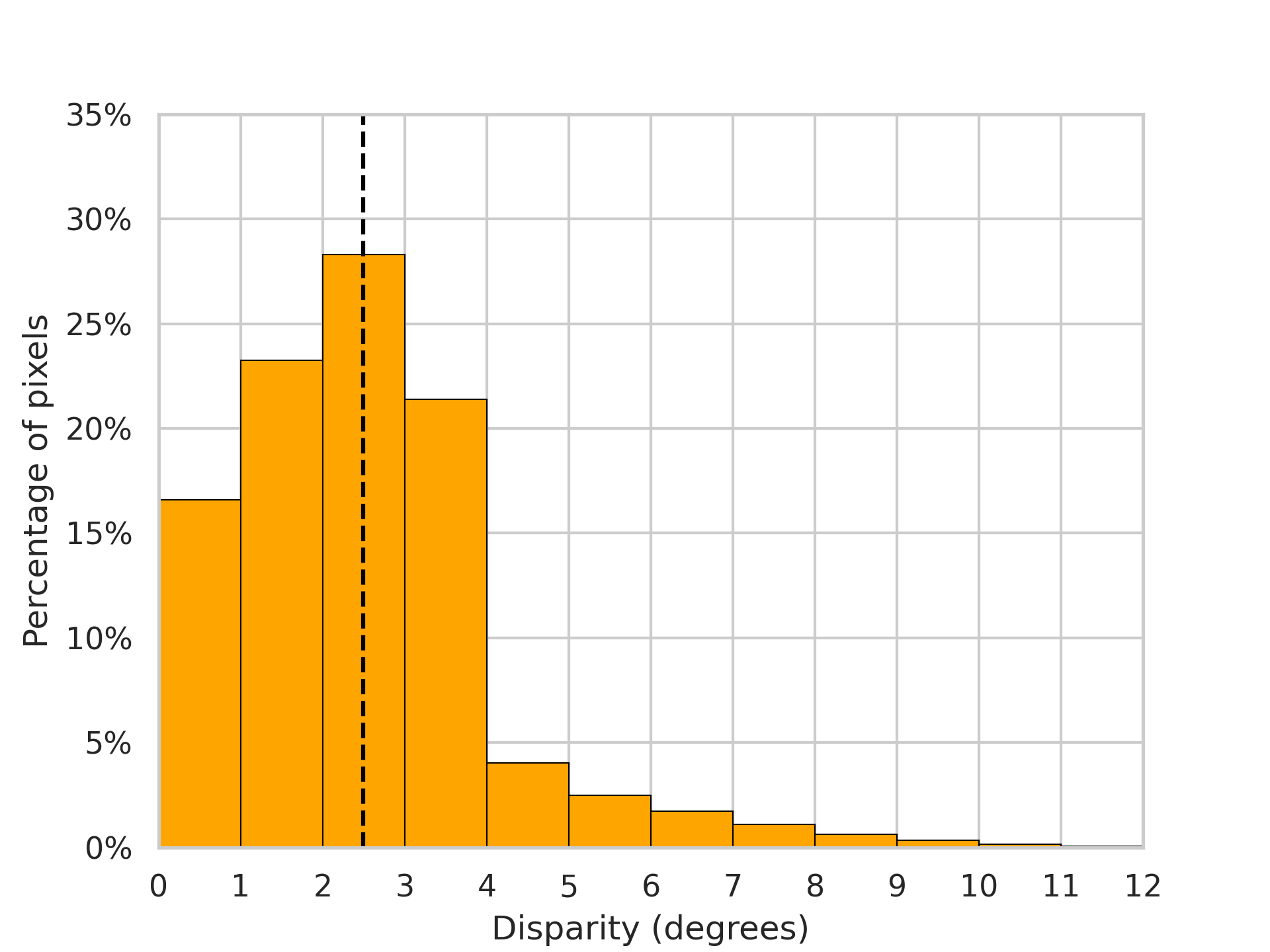}
        \caption{All train sequences - Disparity}
        \label{appx_fig:augmented_disp_all_train}
    \end{subfigure}%
    \hfill
    \begin{subfigure}[t]{0.45\textwidth}
        \includegraphics[width=\linewidth]{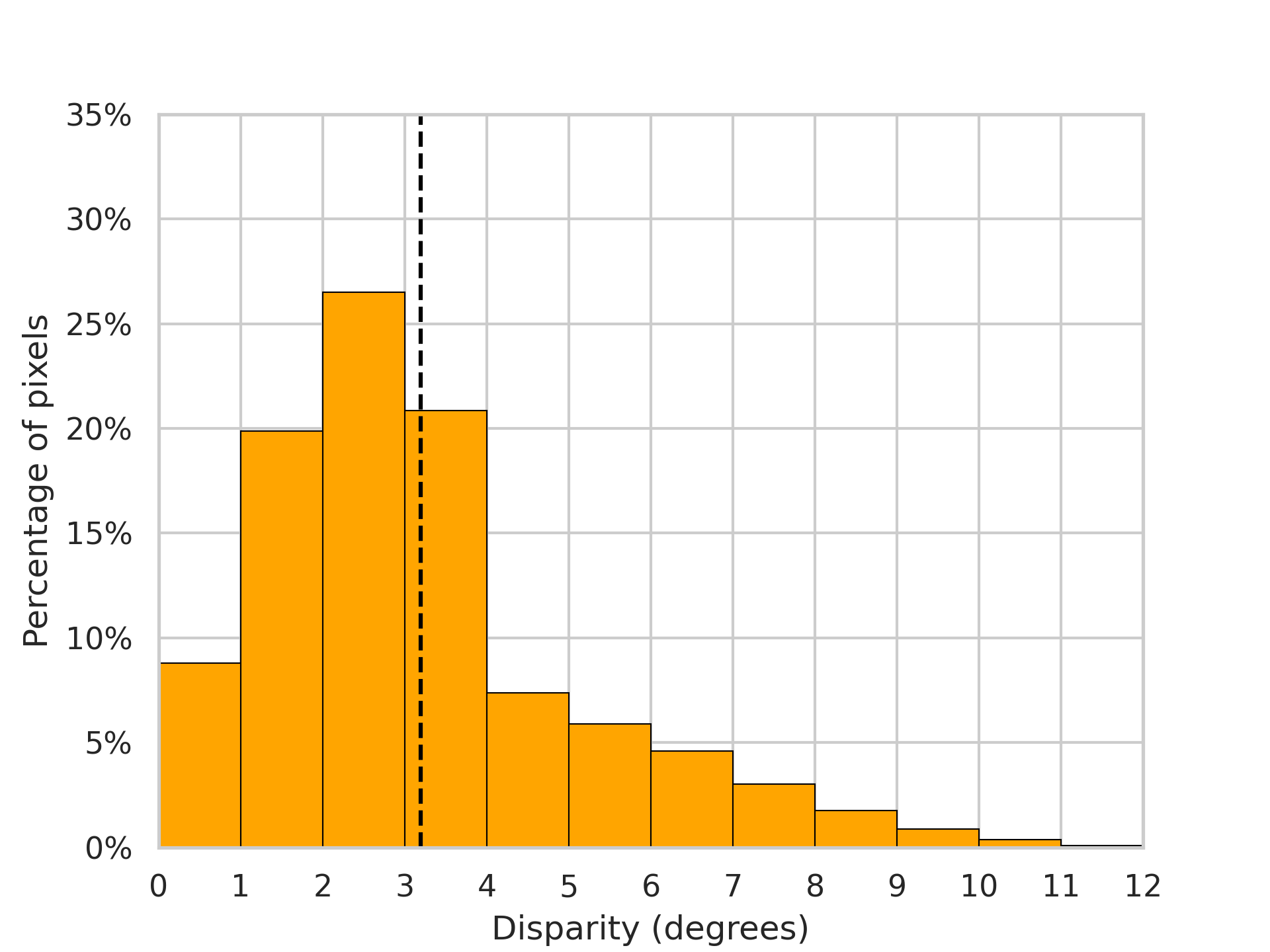}
        \caption{Train indoor sequences - Disparity}
        \label{appx_fig:augmented_disp_indoor_train}
    \end{subfigure}%
    \hfill
    \begin{subfigure}[t]{0.45\textwidth}
        \includegraphics[width=\linewidth]{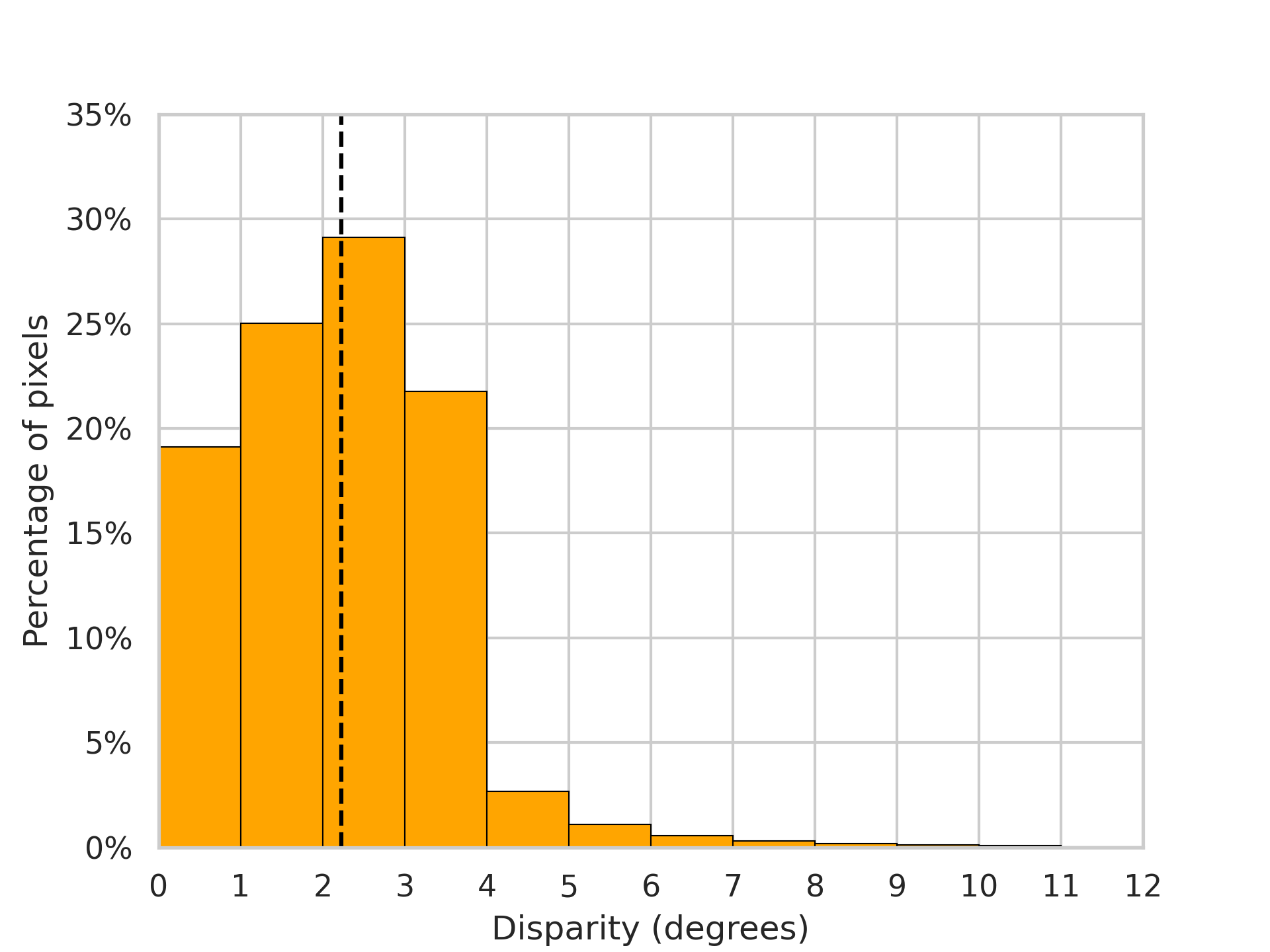}
        \caption{Train outdoor sequences - Disparity}
        \label{appx_fig:augmented_disp_outdoor_train}
    \end{subfigure}%
    \hfill
    \begin{subfigure}[t]{0.45\textwidth}
        \includegraphics[width=\linewidth]{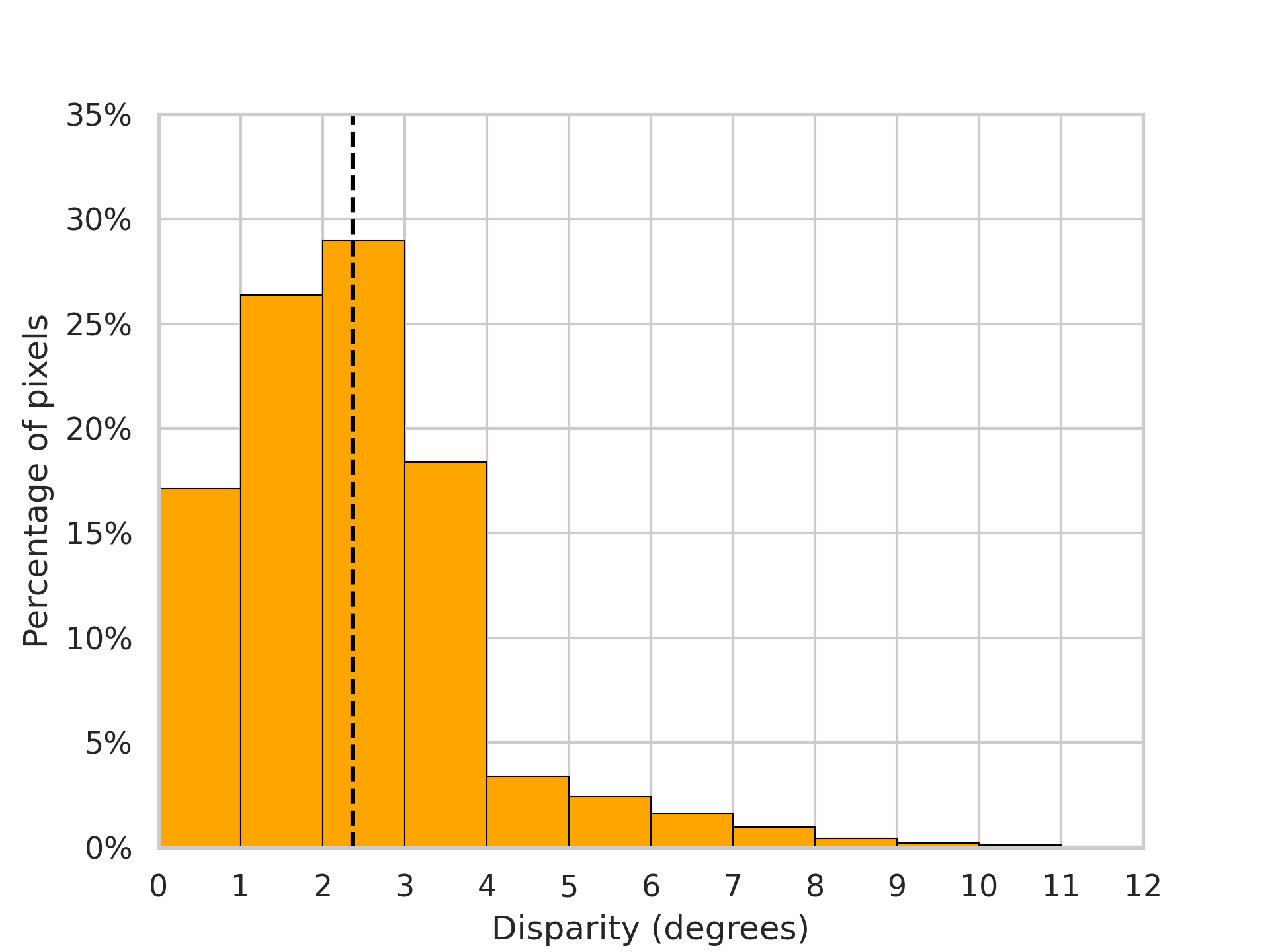}
        \caption{Train night outdoor sequences - Disparity}
        \label{appx_fig:augmented_disp_night_outdoor_train}
    \end{subfigure}%
}    
    \caption{\textbf{Histograms of disparity labels across train splits \underline{after depth completion}}. Each plot's vertical dotted line denotes the considered setting average.}
    \label{appx_fig:augmented_disparity_histograms}
\end{figure*}

In addition to the histograms provided in the main paper, we include more detailed histograms of depth labels (\cref{appx_fig:depth_histograms}) and disparity labels (\cref{appx_fig:disparity_histograms}) across the train and test splits. Additionally, we provide histograms for the augmented train set, \textit{i.e.,} after depth completion, including depth (\cref{appx_fig:augmented_depth_histograms}) and disparity \cref{appx_fig:augmented_disparity_histograms}. 

The analysis shows that train and test distributions maintain consistent patterns across each environment — indoor, outdoor, and night outdoor settings —, suggesting a well-aligned partition. Notably, a distinct concentration of short-range distances within the 0–10 meter range in all settings is attributed to the LiDAR laser capturing the ground surface in each environment. Furthermore, the data exhibits an exponential decay in pixel percentage with increasing depth, correlating with the decrease in LiDAR precision over longer distances. A comparison between indoor and outdoor scenes reveals that indoor sequences, both in training and testing, exhibit a more rapid decline in pixel percentage beyond 10 meters. A similar pattern is observed for disparity values, where train and test distributions align closely. As anticipated, we identify higher disparity values for indoor sequences compared to outdoor scenes due to the greater angular difference between the two cameras for points projected closer to the recording system and lower disparity for further away objects.

When comparing the depth distributions in the training set before and after depth completion, we observe that depth completion increases the density of low-range depth values, particularly within the 0–10 meter range. This shift is likely a result of the algorithm interpolating missing values and removing out-of-distribution samples. This process tends to enrich the representation of closer objects in the dataset. Similar conclusions can be drawn when comparing disparity distributions before and after depth completion.

%=================================================================================%
\section{Data Collection}
\label{appx_sec:details_collection}
%=================================================================================%

This section provides details about the acquisition device, the synchronization between sensors and dataset quality.

%=--------------------------------------------------------------------------------%
\subsection{Data Capture Setup}
\label{appx_subsec:acquisition}
%=--------------------------------------------------------------------------------%

\begin{figure}[ht]
  \centering
  \includegraphics[width=0.35\textwidth]{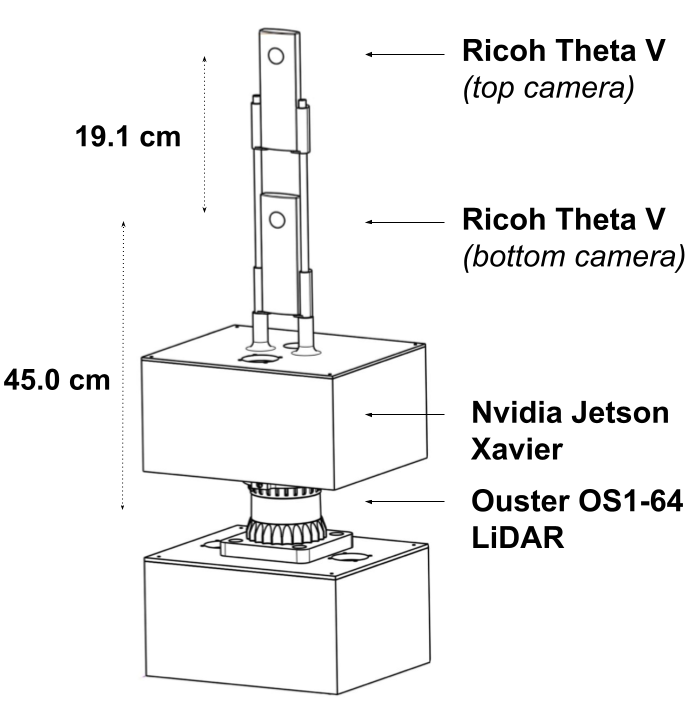}
  \caption{\textbf{\textsc{Helvipad} data acquisition setup}: dual Ricoh Theta V cameras in a top-bottom configuration above an Ouster OS1-64 LiDAR Sensor, and integrated with a NVIDIA Jetson Xavier.}
  \label{fig:acquisition_device}
\end{figure}

The hardware setup consists of a custom-designed support system that aligns all devices horizontally and stacks them vertically. The system integrates two Ricoh Theta V cameras, which capture images in 4K/UHD equirectangular format at 30 fps with an initial resolution of 3840 × 1920 pixels. It also includes an Ouster OS1-64 LiDAR sensor, operating at 10 fp with 64 beams and a vertical field of view of 42.4°. The LiDAR is mounted at the bottom, with the first camera at the top (``\textit{top camera}") and the second camera in the middle (``\textit{bottom camera}"). The vertical distances between the devices are precisely configured, with 19.1 cm separating the two camera lenses and 45 cm between the LiDAR and the bottom camera, as shown in \cref{fig:acquisition_device}. 

\begin{table*}[ht]
\renewcommand{\arraystretch}{1}
\setlength\tabcolsep{5pt}
\rowcolors{2}{gray!10}{white}
\centering
\resizebox{\textwidth}{!}{%
\begin{tabular}{l l l l l l l l r r}
\toprule
\textbf{Sequence} & \textbf{Split} & \textbf{Date} & \textbf{Setting} & \textbf{Dur.} & \textbf{Time of day} & \textbf{Area type} & \textbf{Weather} & \textbf{\# Fr.} & \textbf{\# Peds} \\
\midrule
20231206\_REC\_01\_OUT & train & 2023-12-06 & outdoor & 01:46 & afternoon & ped. sq. & cloudy & 1001 & 37 \\
20231206\_REC\_02\_OUT & test & 2023-12-06 & outdoor & 03:45 & afternoon & ped. sq. & cloudy & 1911 & 45 \\
20240120\_REC\_02\_OUT & train & 2024-01-20 & outdoor & 02:37 & afternoon & road & sunny & 1445 & 7 \\
20240120\_REC\_03\_OUT & train & 2024-01-20 & outdoor & 02:34 & afternoon & road & sunny & 1379 & 9 \\
20240120\_REC\_04\_OUT & train & 2024-01-20 & outdoor & 02:54 & afternoon & footpath & sunny & 1569 & 11 \\
20240120\_REC\_05\_IN & train & 2024-01-20 & indoor & 02:43 & end of day & corridor & n.a. & 1530 & 4 \\
20240120\_REC\_06\_IN & test & 2024-01-20 & indoor & 02:11 & end of day & corridor & n.a. & 998 & 29 \\
20240120\_REC\_07\_IN & train & 2024-01-20 & indoor & 01:54 & end of day & corridor & n.a. & 998 & 5 \\
20240121\_REC\_01\_OUT & train & 2024-01-21 & outdoor & 02:47 & afternoon & footpath & sunny & 1650 & 6 \\
20240121\_REC\_02\_OUT & train & 2024-01-21 & outdoor & 02:26 & afternoon & footpath & sunny & 1425 & 2 \\
20240121\_REC\_03\_OUT & train & 2024-01-21 & outdoor & 02:21 & afternoon & road & sunny & 1375 & 4 \\
20240121\_REC\_04\_OUT & train & 2024-01-21 & outdoor & 02:54 & afternoon & road & sunny & 1780 & 7 \\
20240121\_REC\_05\_OUT & train & 2024-01-21 & outdoor & 02:03 & end of day & road & sunny & 1237 & 6 \\
20240124\_REC\_01\_OUT & train & 2024-01-24 & outdoor & 02:48 & morning & road & cloudy & 1549 & 10 \\
20240124\_REC\_02\_OUT & val & 2024-01-24 & outdoor & 03:21 & morning & footpath & cloudy & 1675 & 10 \\
20240124\_REC\_03\_OUT & test & 2024-01-24 & outdoor & 02:54 & morning & ped. sq. & cloudy & 1681 & 9 \\
20240124\_REC\_04\_OUT & train & 2024-01-24 & outdoor & 03:51 & morning & road & cloudy & 2180 & 6 \\
20240124\_REC\_05\_OUT & train & 2024-01-24 & outdoor & 02:46 & morning & road & cloudy & 1500 & 4 \\
20240124\_REC\_06\_IN & train & 2024-01-24 & indoor & 02:32 & afternoon & corridor & n.a. & 1429 & 44 \\
20240124\_REC\_07\_NOUT & train & 2024-01-24 & outdoor & 02:51 & night & footpath & night & 1700 & 22 \\
20240124\_REC\_08\_NOUT & test & 2024-01-24 & outdoor & 03:51 & night & ped. sq. & night & 2925 & 54 \\
20240124\_REC\_09\_NOUT & train & 2024-01-24 & outdoor & 02:50 & night & footpath & night & 1800 & 58 \\
20240124\_REC\_11\_IN & train & 2024-01-24 & indoor & 01:39 & end of day & hall & n.a. & 1000 & 11 \\
20240124\_REC\_12\_IN & val & 2024-01-24 & indoor & 02:13 & end of day & hall & n.a. & 1320 & 13 \\
20240127\_REC\_01\_IN & test & 2024-01-27 & indoor & 02:01 & morning & corridor & n.a. & 1201 & 2 \\
20240127\_REC\_02\_OUT & test & 2024-01-27 & indoor & 02:20 & morning & parking & n.a. & 1430 & 2 \\
\bottomrule
\end{tabular}%
}
\caption{\textbf{Overview of the collected sequences.} Abbreviations: Dur. = Duration, displayed in minutes:seconds; ped. sq. = pedestrian square; \# Fr. = number of frames; \# Peds = number of pedestrians.}
\label{appx_tab:seq}
\end{table*}

%%%%%%%%%%%%%%%%%%%%%%%%%%%AUGMENTED DEPTH%%%%%%%%%%%%%%%%%%%%%%%%%

The cameras function as external modules, while the LiDAR operates via Robot Operating System (ROS) on an embedded NVIDIA Jetson Xavier. This central processor manages data capture and ensures synchronization across all devices. The entire setup is mounted on a custom-built, remotely controlled robot chassis, offering mobility and a fully integrated, portable acquisition solution.

%=--------------------------------------------------------------------------------%
\subsection{Synchronization between sensors device}
\label{appx_subsec:sync}
%=--------------------------------------------------------------------------------%

\begin{figure}[ht]
  \centering
  \includegraphics[width=\linewidth]{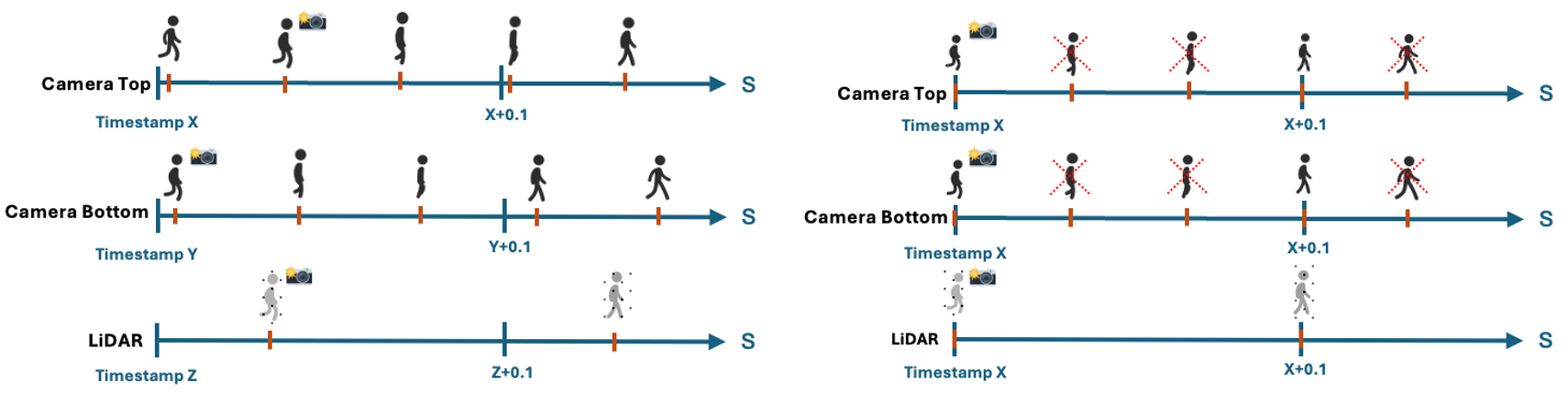}
   \caption{\textbf{Synchronization of LiDAR and cameras using a flash trigger.} The illustration shows data alignment before (\textbf{left}) and after (\textbf{right}) synchronization.}
   \label{appx_fig:onecol}
\end{figure}

To ensure accurate synchronization between the sensors, we use a hardware-triggered synchronization method. At the start of each recording, an external flash lasting 33 ms is activated in front of the cameras, creating a visible synchronization marker in the video streams. Simultaneously, the precise ROS timestamp of the flash event is recorded in the LiDAR data, which provides a precise timestamp. During post-processing, we identify the blinded frames and corresponding ROS timestamps, re-aligning all data streams to start from this synchronized reference, as shown in \cref{appx_fig:onecol}

To match the LiDAR’s frame rate (10 fps), we retain one out of every three camera frames. The 33ms flash duration ensures it is captured in at least one camera frame, with a maximum potential de-synchronization of half a frame interval (16.67 ms) if the flash occurs just after a frame is captured. This reasoning extends to the LiDAR-camera synchronization, resulting in a maximum de-synchronization of 16.67 ms across all sensors.

%=--------------------------------------------------------------------------------%
\subsection{LiDAR-Image Projection Quality Assessment}
\label{appx_subsec:calib_quality}
%=--------------------------------------------------------------------------------%

In the absence of a reliable, standardized method to evaluate the accuracy of LiDAR point projections onto equirectangular image planes, manual validation serves as a practical and precise alternative. Visual inspection and manual selection of corresponding points have been highlighted in studies such as ~\citep{line_registration_2017}, where the authors emphasized the role of manual evaluation in aligning data when automated methods are insufficient. Similarly, in ~\citep{uncertainty_projection_2019}, the challenges of achieving accurate projections without ground truth were addressed, underscoring the importance of visual assessment for high-precision tasks. 

Therefore, in this work, we adopt a manual point selection approach to evaluate the projection of LiDAR points onto 2D equirectangular image planes. This process involves manually selecting corresponding points, such as object edges, on both the LiDAR-projected data and the images. These selected points are then used to compute the pixel-wise error, which quantifies the projection's accuracy.

For each selected point pair, we calculate the Euclidean distance between the projection point $(x_{\text{proj}}, y_{\text{proj}})$ and the corresponding real point $(x_{\text{real}}, y_{\text{real}})$:
\begin{equation}
\text{Error} = \sqrt{(x_{\text{proj}} - x_{\text{real}})^2 + (y_{\text{proj}} - y_{\text{real}})^2}.
\end{equation}
This error is averaged across multiple images to provide a comprehensive assessment of the projection's accuracy. Additionally, a relative error metric is computed by normalizing the pixel error by the image diagonal, enabling a consistent evaluation across different resolutions.

\begin{figure}[t]
    \centering
    \includegraphics[width=\columnwidth]{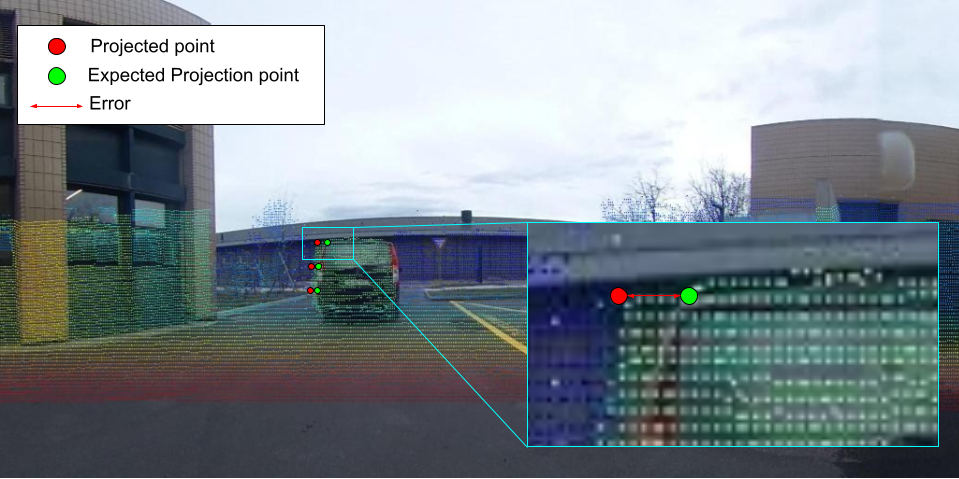}
    \caption{\textbf{Illustration of LiDAR point projection onto an equirectangular image}. The red dots represent the projected points from LiDAR data, while the green dots indicate the expected projection points on the image. The red arrows show the pixel-wise error between the projected points and the expected points, which is used to quantify the projection accuracy. This error metric aids in evaluating the fidelity of LiDAR-to-image projection in the \textsc{Helvipad} dataset.}
    \label{fig:projection_error}
\end{figure}

\cref{tab:errors} provides a summary of the pixel-wise errors measured accross sequences in the dataset. Each sequence was evaluated by manually selecting corresponding points between the projected LiDAR points and the equirectangular images, followed by the calculation of Euclidean distances as described earlier, and average errors for each sequence are reported along with the overall dataset average.

\begin{table}[ht]
\centering
\setlength{\tabcolsep}{8pt} % Adjust column spacing
\renewcommand{\arraystretch}{1.2} % Adjust row spacing
\resizebox{\columnwidth}{!}{%
\begin{tabular}{lrr}
\toprule
\textbf{Sequence Name} & \textbf{Avg. Pixel Error } (px) & \textbf{Relative Error } (\%) \\
\midrule
20231206\_REC\_01\_OUT & 6.4 & 0.32 \\
20240120\_REC\_02\_OUT & 9.5 & 0.48 \\
20240120\_REC\_03\_OUT & 8.7 & 0.44 \\
20240120\_REC\_04\_OUT & 7.1 & 0.36 \\
20240120\_REC\_05\_IN  & 10.2 & 0.51 \\
20240120\_REC\_07\_IN  & 6.8 & 0.34 \\
20240121\_REC\_01\_OUT & 9.1 & 0.46 \\
20240121\_REC\_02\_OUT & 7.5 & 0.38 \\
20240121\_REC\_03\_OUT & 8.3 & 0.42 \\
20240121\_REC\_04\_OUT & 7.9 & 0.40 \\
20240121\_REC\_05\_OUT & 8.0 & 0.40 \\
20240124\_REC\_01\_OUT & 10.4 & 0.52 \\
20240124\_REC\_02\_OUT & 6.9 & 0.35 \\
20240124\_REC\_04\_OUT & 8.1 & 0.41 \\
20240124\_REC\_05\_OUT & 7.3 & 0.37 \\
20240124\_REC\_06\_IN  & 9.6 & 0.48 \\
20240124\_REC\_07\_NOUT & 7.4 & 0.37 \\
20240124\_REC\_09\_NOUT & 8.9 & 0.45 \\
20240124\_REC\_11\_IN  & 6.7 & 0.33 \\
20240124\_REC\_12\_IN  & 8.8 & 0.44 \\
\midrule
\textbf{Overall} & \textbf{8.0} & \textbf{0.40} \\
\bottomrule
\end{tabular}%
}
\caption{\textbf{Pixel-wise projection errors of LiDAR points onto equirectangular images}, per sequence. Relative error is expressed as a percentage of the image diagonal for context clarity.}
\label{tab:errors}
\end{table}

The overall average pixel error across the dataset is 8.0 pixels, corresponding to a relative error of 0.40\% of the image diagonal. This level of precision validates the high-quality LiDAR-to-image projection tasks in the \textsc{Helvipad} dataset.

%%%%%%%%%%%%%%%%%%%%%%%%%%%DISP%%%%%%%%%%%%%%%%%%%%%%%%%

%%%%%%%%%%%%%%%%%%%%%%%%%%%AUGMENTED DISP%%%%%%%%%%%%%%%%%%%%%%%%%

%=================================================================================%
\section{Depth Completion}
\label{appx_sec:depth_completion}
%=================================================================================%

This section provides an in-depth evaluation of our depth completion pipeline, detailing the evaluation methods, hyperparameter selection, and comparison of temporal aggregation techniques. Quantitative and qualitative results are also presented to demonstrate the effectiveness of our approach.

%=--------------------------------------------------------------------------------%
\subsection{Evaluation Method}
\label{appx_subsec:evaluation}
%=--------------------------------------------------------------------------------%

The evaluation of our depth completion method follows a structure akin to standard machine learning. The dataset is split into training and test sets, then performance metrics are computed on the test set within the 3D space.

\paragraph{Creation of training and test set.}
To identify the optimal hyperparameters for depth completion, the points of each measured point cloud are divided into a training and test set using a classical 80-20-split. Points are sampled from a uniform distribution over all input points without replacement.

This approach primarily evaluates metrics over points with low distances to its neighbors, which are easier to estimate. When the pipeline generates points on a uniform grid (e.g., an image), the metrics reflect lower bounds on actual errors due to this distribution shift. While this bias limits direct comparison to image-based errors, it is acceptable for hyperparameter optimization and data augmentation purposes.

\paragraph{Evaluation metrics.}
To evaluate the merits of different options for the depth completion pipeline, we calculate the following metrics: Mean Absolute Error (\textit{MAE}), Root Mean Squared Error (\textit{RMSE}), mean absolute relative error (\textit{MARE}), Inlier Ratio (\textit{IR}) and Actual Ratio of Interpolated Points (\textit{ARIP}).
The calculation of the first three mentioned metrics is the same as for depth estimation methods and can be found in \cref{appx_subsec:metrics}.
The \textit{IR} corresponds to the ratio of estimated depth labels that have an absolute error of less than $t_\text{{inlier}} = 1 \%$.
Given the number $N$ of estimated depths among points of all point clouds and sequences, the \textit{IR} can be calculated in the following way:
\begin{equation}
    \text{IR} =  \frac{1}{N} \sum_{i=1}^{N} \mathbb{I}_{\left|r_\text{est, i} - r_\text{true, i}\right| < t_\text{inlier}}.
\end{equation}
In \cref{subsec:depth_completion}, the \textit{RIP} is defined as the ratio of interpolated points after filtering.
As the uncertainty estimates of all query points within one sequence do not fit into memory, thresholds for the uncertainty-based filtering are calculated for each point cloud and then averaged per sequence. This leads to an \textit{ARIP} which differs slightly from the desired \textit{RIP}.

\paragraph{Ratio of labeled pixels.}
To calculate the ratio of labeled pixels the region within an image that contains potential labeled pixels must be identified.
While the whole image width $W$ can be labeled, the potential labeled region along the height is restricted.
Specifically, the minimum ($H_\text{min}$) and maximum ($H_\text{max}$) height with potential labels depend on the distance $r$ of the points at the minimum ($\theta_\text{min}$) and maximum ($\theta_\text{max}$) of the LiDAR's vertical field of view.
For each image, we individually determine $H_\text{min}$ and $H_\text{max}$ based on the smallest and largest row index containing at least one label in the original depth map.
Any labels in the completed depth map that fall outside this height range are filtered out.
With this, the Ratio of Labeled Pixels (\textit{RLP}) can be calculated as a function of the total number of labeled pixels $n_{label}$ in the entire image:
\begin{equation}
    \text{RLP}(n_\text{label}) = \frac{n_\text{label}}{W \times (H_\text{max} - H_{\text{min}})}
    \label{appx_eq:rlp}
\end{equation}

%=--------------------------------------------------------------------------------%
\subsection{Choice of Hyperparameters}
\label{appx_subsec:hyperparameters}
%=--------------------------------------------------------------------------------%

This section describes how the hyperparameters for the depth completion pipeline, introduced in \cref{subsec:depth_completion}, have been chosen.

\begin{figure*}
    \centering
    \begin{subfigure}[t]{0.238\textwidth}
        \centering
        \includegraphics[width=\columnwidth]{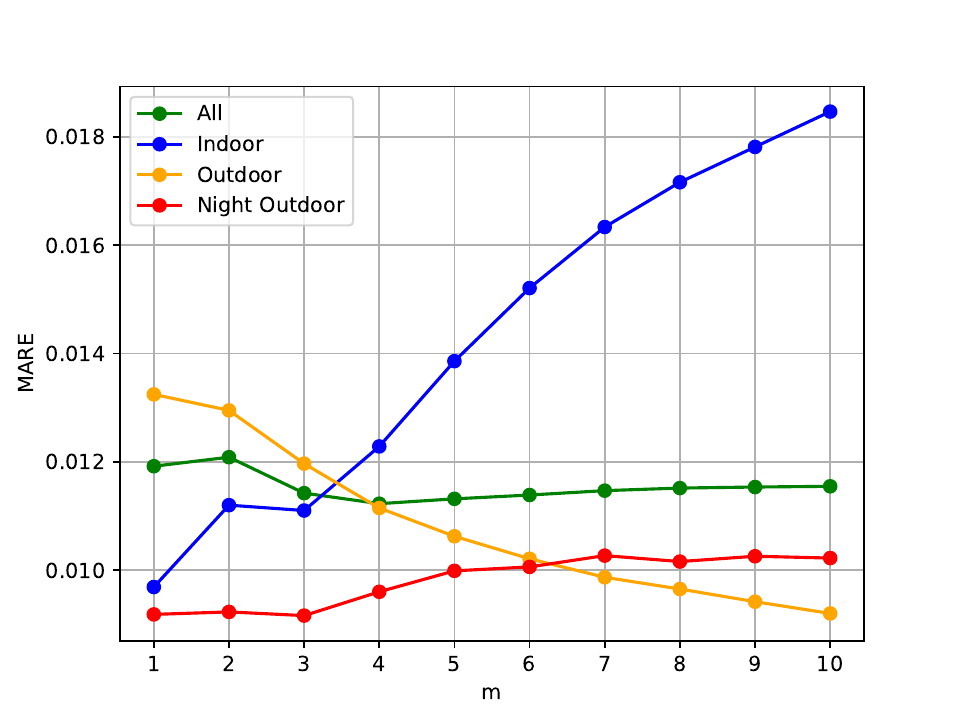}
        \caption{Mean Absolute Relative Error (\textit{MARE}).}
        \label{subfig:number_agg_mare}
    \end{subfigure}
    \hspace{0.005\textwidth} % Add horizontal space
    \begin{subfigure}[t]{0.238\textwidth}
        \centering
        \includegraphics[width=\columnwidth]{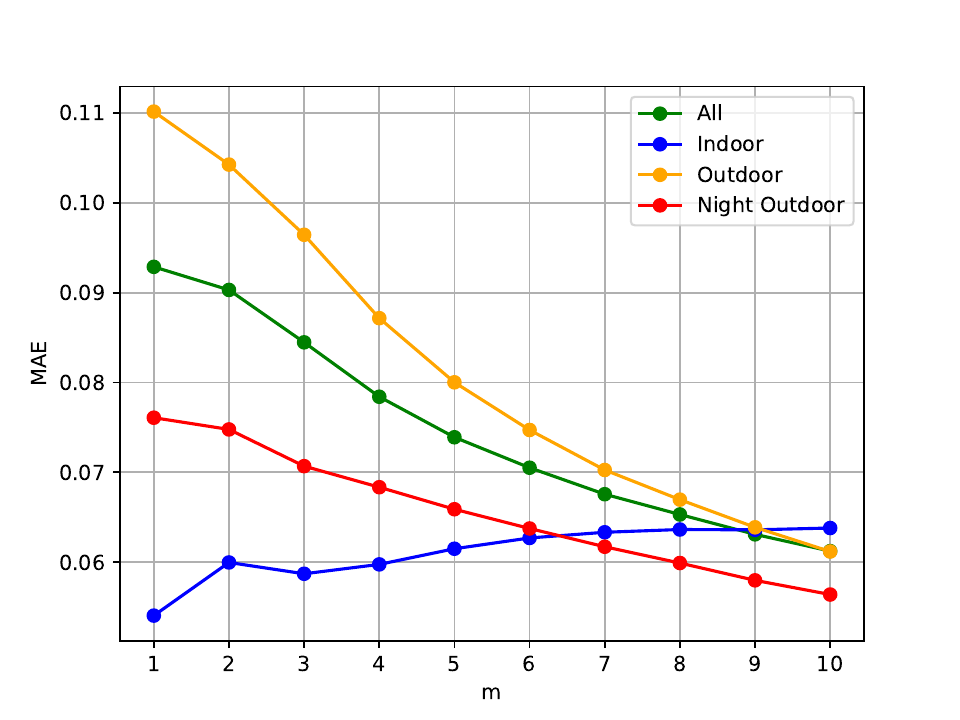}
        \caption{Mean Absolute Error (\textit{MAE}).}
        \label{subfig:number_agg_mae}
    \end{subfigure}
    \hspace{0.005\textwidth} % Add horizontal space
    \begin{subfigure}[t]{0.238\textwidth}
        \centering
        \includegraphics[width=\columnwidth]{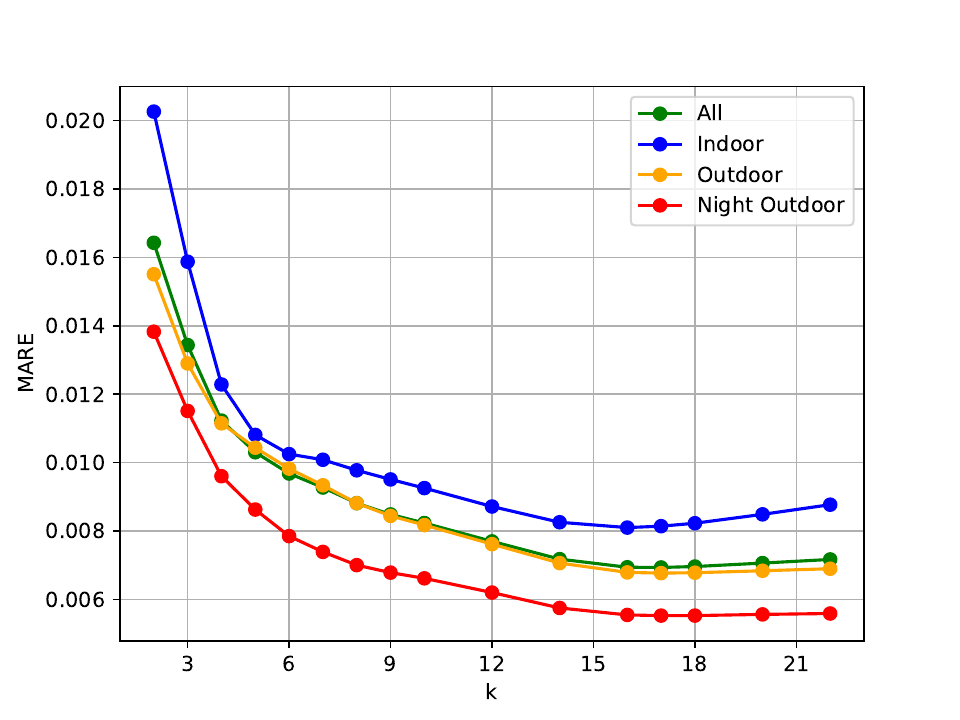}
        \caption{Mean Absolute Relative Error (\textit{MARE}).}
        \label{subfig:number_neighbors_mare}
    \end{subfigure}
    \hspace{0.005\textwidth} % Add horizontal space
    \begin{subfigure}[t]{0.238\textwidth}
        \centering
        \includegraphics[width=\columnwidth]{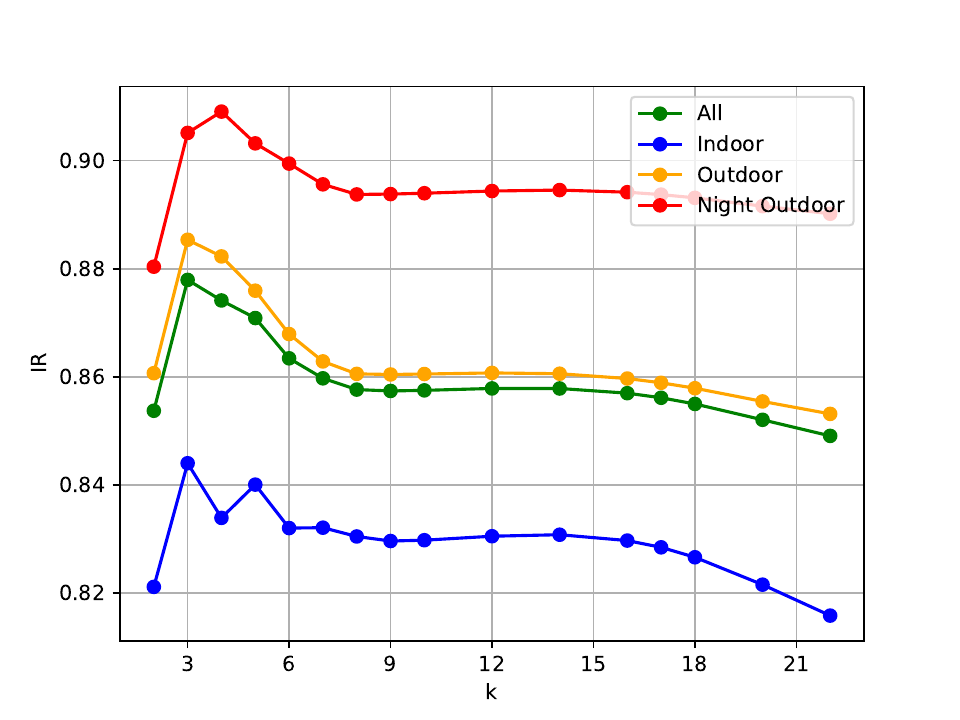}
        \caption{Inlier Ratio (\textit{IR}).}
        \label{subfig:number_neighbors_ir}
    \end{subfigure}
    \caption{Selected metrics with a variable number of aggregated point clouds (\cref{subfig:number_agg_mare} and \cref{subfig:number_agg_mae}) or $m = 4$ previous / next point clouds (\cref{subfig:number_neighbors_mare} and \cref{subfig:number_neighbors_ir}) on \textit{all}, \textit{indoor}, \textit{outdoor} and \textit{night outdoor} train sequences.
    The ratio of interpolated points is set to $\text{RIP} = 80\%$ and $k=4$ is chosen as the numbers of neighbors (\cref{subfig:number_agg_mare} and \cref{subfig:number_agg_mae}) or a variable number of neighbors $k$ (\cref{subfig:number_neighbors_mare} and \cref{subfig:number_neighbors_ir}).
    We report all depth metrics in m}
\end{figure*}%

\vspace{0.2cm}
\noindent
\textbf{Number of aggregated point clouds.}
For temporal aggregation, we fuse the $m$ previous and $m$ following point clouds.
The minimum \textit{MARE} for \textit{all} sequences is observed for $m=4$ (\cref{subfig:number_agg_mare}).
However, the \textit{MAE} continues to decrease for all sequences except \textit{indoor}, where it increases for higher $m$ values (\cref{subfig:number_agg_mae}).
As the total number of points for all \textit{indoor} sequences is lower than for all \textit{outdoor} and \textit{night outdoor} train sequences, the error for \textit{all} sequences is less influenced by \textit{indoor} sequences.
We set $m = 4$ to provide high quality depth labels also for \textit{indoor} sequences and low depth values.

\paragraph{Number of neighbors.}
The number of neighbors $k$ in interpolation can be observed to reduce the \textit{MARE} until $k=17$ is reached (\cref{subfig:number_neighbors_mare}).
Then, the \textit{MARE} then becomes greater again.
However, the \textit{IR} starts to decrease from $k=3$ already (\cref{subfig:number_neighbors_ir}).
We select $k=17$ because the positive influence on the \textit{MARE} is greater than the negative influence on the \textit{IR}.

\paragraph{Ratio of interpolated points.}

\begin{figure}[ht]
    \centering
    \includegraphics[width=0.238\textwidth]{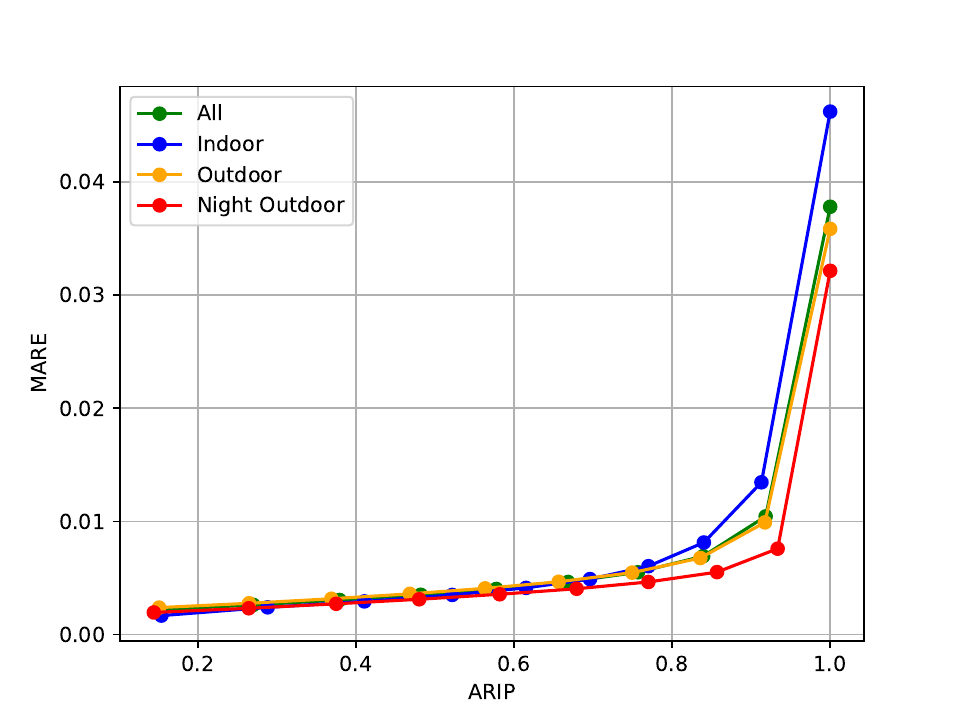}
    \caption{\textit{MARE} for $m = 4$ on \textit{all}, \textit{indoor}, \textit{outdoor} and \textit{night outdoor} sequences.
    A variable \textit{ARIP} is interpolated with $k=17$ neighbors.}
    \label{fig:calc_points_ratio}
\end{figure}

As expected, the evaluation metrics become worse for a higher \textit{ARIP}.
By observing the \textit{MARE} in dependence of the \textit{ARIP} in \cref{fig:calc_points_ratio}, it can be seen that setting this ratio slightly below $1$ reduces the \textit{MARE} significantly.
Setting $\text{RIP}=0.8$ for $\text{ARIP} \approx 0.8$ appears to represent a good balance between the provision of labels for a large number of points and the simultaneous minimization of the induced errors.

\paragraph{Out-of-distribution threshold.}
The out-of-distribution threshold $t_\text{OOD}$ to filter query points with insufficient neighbors cannot be determined based on the train-test split. Instead, a heuristic has to be derived theoretically. Each LiDAR scan provides depth labels on a spherical, regular grid with a vertical, angular resolution of
\begin{equation}
    \Delta\theta = \frac{\text{FOV}_\text{v}}{n_\text{beams}} = \frac{42.4^{\circ}}{64} \approx 0.66^{\circ}
\end{equation}
and a horizontal, angular resolution of
\begin{equation}
    \Delta\varphi = \frac{\text{FOV}_\text{h}}{n_\text{channels, h}} = \frac{360^{\circ}}{1024} \approx 0.35^{\circ}.
\end{equation}
Given the number of accumulated point clouds $2m + 1$ and the number of neighbors for interpolation $k$ and the assumption that all accumulated scans provide measurements in the neighborhood of a query position, the number of neighbors, chosen per LiDAR scan, $n_\text{neighbors, grid}$ can be approximated to:
\begin{equation}
    n_\text{neighbors, grid} = \frac{k}{2m + 1} = \frac{17}{9} \approx 2.
\end{equation}
A position in the regular LiDAR depth grid has the maximum distances to its 2 nearest neighbors, if its position is exactly in the middle of two rows and two columns in the grid.
As a result, the average distance $d_q$ of a query position $(\theta_q, \varphi_q)$ to its neighbors is below the following threshold, if all accumulated point clouds provide labels in the region of the query:
\begin{equation}
    t_\text{OOD} = \sqrt{\left(\frac{\Delta\theta}{2}\right)^2 + \left(\frac{\Delta\varphi}{2}\right)^2} \approx 0.37^{\circ}.
\end{equation}

\paragraph{Number of spherical grid points.}
To map the points from the 3D space to the image we create a spherical grid with approximately uniformly distributed points.
The number of spherical grid points is set to $n_\text{grid}=20,000,000$.
This value represents a compromise between high computational loads for high $n_\text{grid}$ values and missing depth information for the subsequent projection for low $n_\text{grid}$.
Missing depth information leads to less labeled pixels.

\paragraph{Range of spherical grid points.}
As the vertical field of view of the LiDAR sensor is limited, we filter all grid points that are out of its view.
The value for the threshold limiting the polar angle can be determined in the following way:
\begin{equation}
    t_\theta = \frac{180^{\circ} - \text{FOV}_\text{v}}{2} = \frac{180^{\circ} - 42.4^{\circ}}{2} = 68.8^{\circ}.
\end{equation}
All query grid points whose polar angle $\theta_q$ is not in the range $\theta_q \in [t_\theta, 180^{\circ} - t_\theta]$ are not mapped to the image.

To summarize, we set the hyperparameters in the following way: $m=4$, $k=17$, $\text{RIP}=0.8$, $t_\text{OOD}\approx0.37^{\circ}$, $n_\text{grid}=20,000,000$ and $t_\theta = 68.8^{\circ}$.

%=--------------------------------------------------------------------------------%
\subsection{Temporal Aggregation Comparison}
\label{appx_subsec:comp_temp_agg}
%=--------------------------------------------------------------------------------%
To aggregate multiple point clouds, previous and following scans can be aggregated directly (\textit{no movement}) or transformed based on odometry information of the robot.
The \textsc{Helvipad} dataset provides only omnidirectional stereo images and LiDAR point clouds but no odometry information.
KISS-ICP~\cite{vizzo2023kiss} is one of the state-of-the-art approaches for LiDAR odometry.
It is based on the ICP algorithm and creates a local map of the environment.
As it is not possible to evaluate the quality of estimated odometry data with the dataset, using a robust method that does not need hyperparameter optimization, such as KISS-ICP, is a suitable choice to obtain odometry information for the dataset.

We compare no temporal aggregation (\textit{no aggregation}), temporal aggregation without transforming point clouds (\textit{no movement}) and temporal aggregation based on odometry information obtained with the KISS-ICP (\textit{KISS-ICP}) in \cref{table:temp_agg}.
\begin{table}[t]
\setlength\tabcolsep{5pt}
\centering
\resizebox{\columnwidth}{!}{%
\begin{tabular}{l c c c c c}
\toprule
\textbf{Method} & \textbf{ARIP~$\uparrow$} & \textbf{MAE~$\downarrow$} & \textbf{RMSE~$\downarrow$} & \textbf{MARE~$\downarrow$} & \textbf{IR~$\uparrow$} \\
\midrule
\textit{No Aggregation} & 0.811 & 0.096 & 0.864 & \textbf{0.011} & \textbf{0.757} \\
\textit{No Movement} & \textbf{0.841} & \textbf{0.093} & \textbf{0.801} & 0.012 & 0.749 \\
\textit{KISS-ICP} & 0.828 & 0.103 & 0.828 & 0.017 & 0.618 \\
\bottomrule
\end{tabular}%
}
\caption{Evaluation metrics for different temporal aggregation methods with $m=1$, $\text{RIP}=0.8$ and $k=4$ on \textit{all} test sequences.
We report all depth metrics in m.
The best results are highlighted in \textbf{bold}.}
\label{table:temp_agg}
\end{table}
The \textit{KISS-ICP} approach yields the least favorable results in most of the metrics.
This may be attributed to the presence of moving people in the scene, coupled with the employed interpolation method.
\textit{No movement} is clearly better than \textit{no aggregation} in \textit{RMSE} and slightly better in \textit{MAE}.
\textit{No aggregation} is the best approach in terms of \textit{MARE} and \textit{IR}.
However, it must be noted that the \textit{ARIP} is also the lowest for this method.
Consequently, the \textit{MARE} and \textit{IR} for the same ratio may be lower than the ones of the \textit{no movement} method.
Overall, the results of the \textit{no movement} temporal aggregation method are the most favorable.
Thus, we use it for the temporal aggregation.

%=--------------------------------------------------------------------------------%
\subsection{Results}
\label{appx_subsec:results}
%=--------------------------------------------------------------------------------%

\vspace{0.2cm}
\noindent
\textbf{Quantitative results.}
The evaluation metrics, described in \cref{appx_subsec:evaluation}, for the final hyperparameter configuration, specified in \cref{appx_subsec:hyperparameters}, are summarized in \cref{table:depth_completion_errors}.
\begin{table}[ht]
\centering
\setlength\tabcolsep{5pt}
\resizebox{\columnwidth}{!}{%
\begin{tabular}{l c c c c c}
\toprule
\textbf{Sequences} & \textbf{ARIP~$\uparrow$} & \textbf{MAE~$\downarrow$} & \textbf{RMSE~$\downarrow$} & \textbf{MARE~$\downarrow$} & \textbf{IR~$\uparrow$} \\
\midrule
    \textit{All} & 0.839 & 0.054 & 0.398 & 0.007 & 0.856 \\
    \midrule
    \textit{Indoor} & 0.840 & 0.046 & 0.264 & 0.008 & 0.828 \\
    \midrule
    \textit{Outdoor} & 0.836 & 0.058 & 0.440 & 0.007 & 0.859 \\
    \midrule
    \textit{Night outdoor} & 0.857 & 0.048 & 0.371 & 0.006 & 0.894 \\
\bottomrule
\end{tabular}%
}
\caption{Evaluation metrics for the final hyperparameters of the depth completion on \textit{all}, \textit{indoor}, \textit{outdoor}, and \textit{night outdoor} train sequences.
We report all depth metrics in m.}
\label{table:depth_completion_errors}
\end{table}
MAE, RMSE and MARE are significantly lower than the depth results of the depth estimation baselines of \cref{tab:comparative_results}.
Thus, the induced errors by the depth completion are acceptable as a data augmentation technique.

\cref{table:depth_completion_stats} exhibits that the \textit{RLP}, as defined in \cref{appx_subsec:evaluation}, is increased approximately by a factor of 5 following the application of depth completion across all sequence types.
The maximum number of labelable pixels $\text{n}_\text{lab, max}$ corresponds to the denominator in \cref{appx_eq:rlp}.
\begin{table}[ht]
\centering
\setlength\tabcolsep{5pt}
\resizebox{\columnwidth}{!}{%
\begin{tabular}{l c c c c c}
\toprule
\textbf{Sequences} & \textbf{$\text{n}_\text{lab, max}$} & \textbf{$\text{n}_\text{lab, ori}$} & \textbf{$\text{RLP}_\text{ori}$} & \textbf{$\text{n}_\text{lab, aug}$} & \textbf{$\text{RLP}_\text{aug}$} \\
\midrule
    \textit{All} & 14.4B & 1.7B & 11.9\% & 9.6B & 60.7\% \\
    \midrule
    \textit{Indoor} & 3.1B & 0.4B & 12.8\% & 1.9B & 62.1\% \\
    \midrule
    \textit{Outdoor} & 9.6B & 1.1B & 11.5\% & 5.7B & 59.9\% \\
    \midrule
    \textit{Night outdoor} & 1.7B & 0.2B & 11.8\% & 1.1B & 62.0\% \\
\bottomrule
\end{tabular}%
}
\caption{Maximum number of labelable pixels $\text{n}_\text{lab, max}$, labeled pixels $\text{n}_\text{lab}$ and ratio of labeled points $\text{RLP}$ for \textit{all}, \textit{indoor}, \textit{outdoor}, and \textit{night outdoor} train sequences.}
\label{table:depth_completion_stats}
\end{table}

\vspace{0.2cm}
\noindent
\textbf{Qualitative results.}
\begin{figure*}
    \centering
    \includegraphics[width=\textwidth]{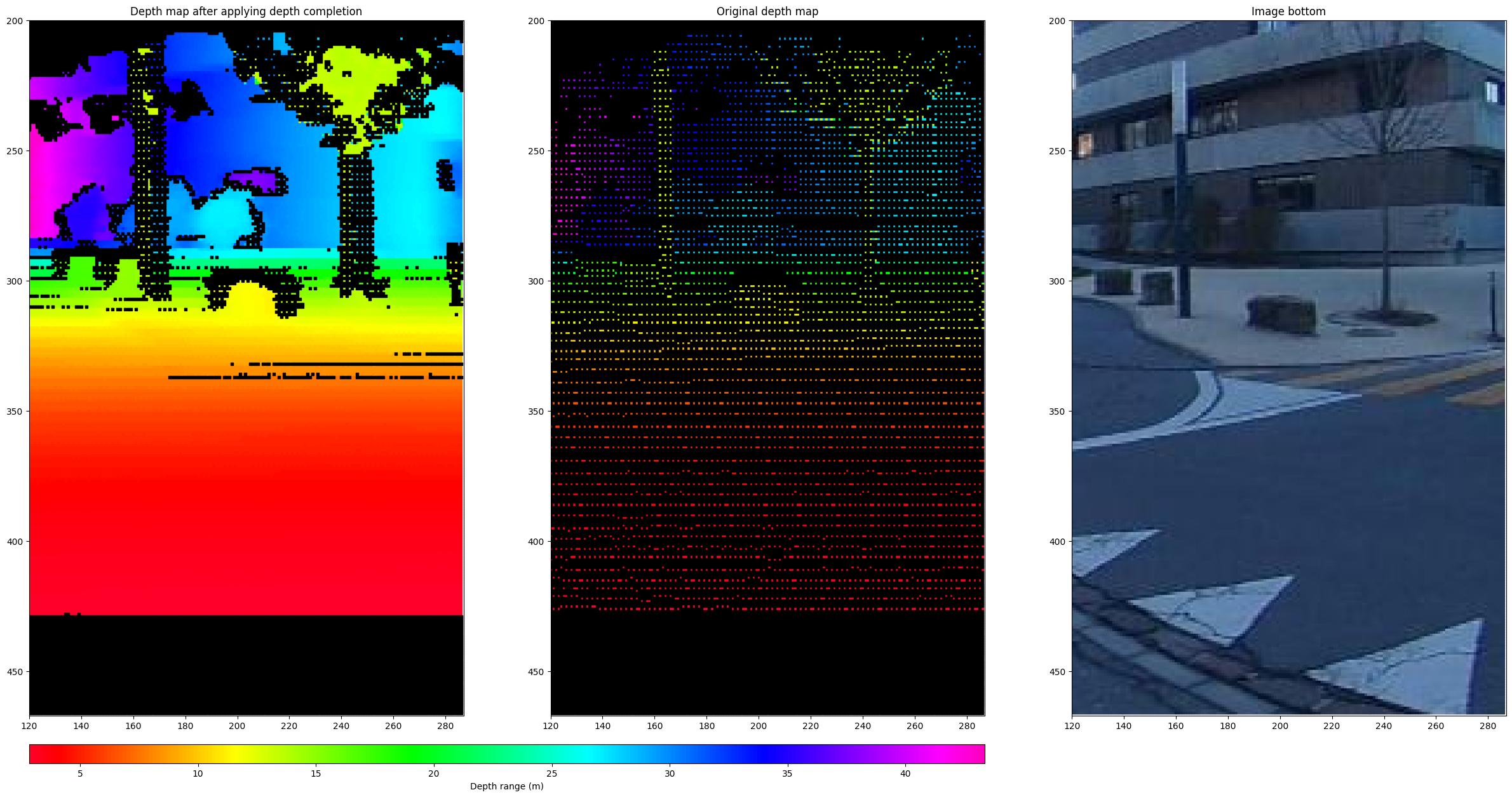}
    \caption{Depth completed depth map, original depth map and bottom image in detail view from an \textit{outdoor} sequence.}
    \label{appx_fig:depth_completion_detail}
\end{figure*}%
\begin{figure*}
    \centering
    \includegraphics[width=\textwidth]{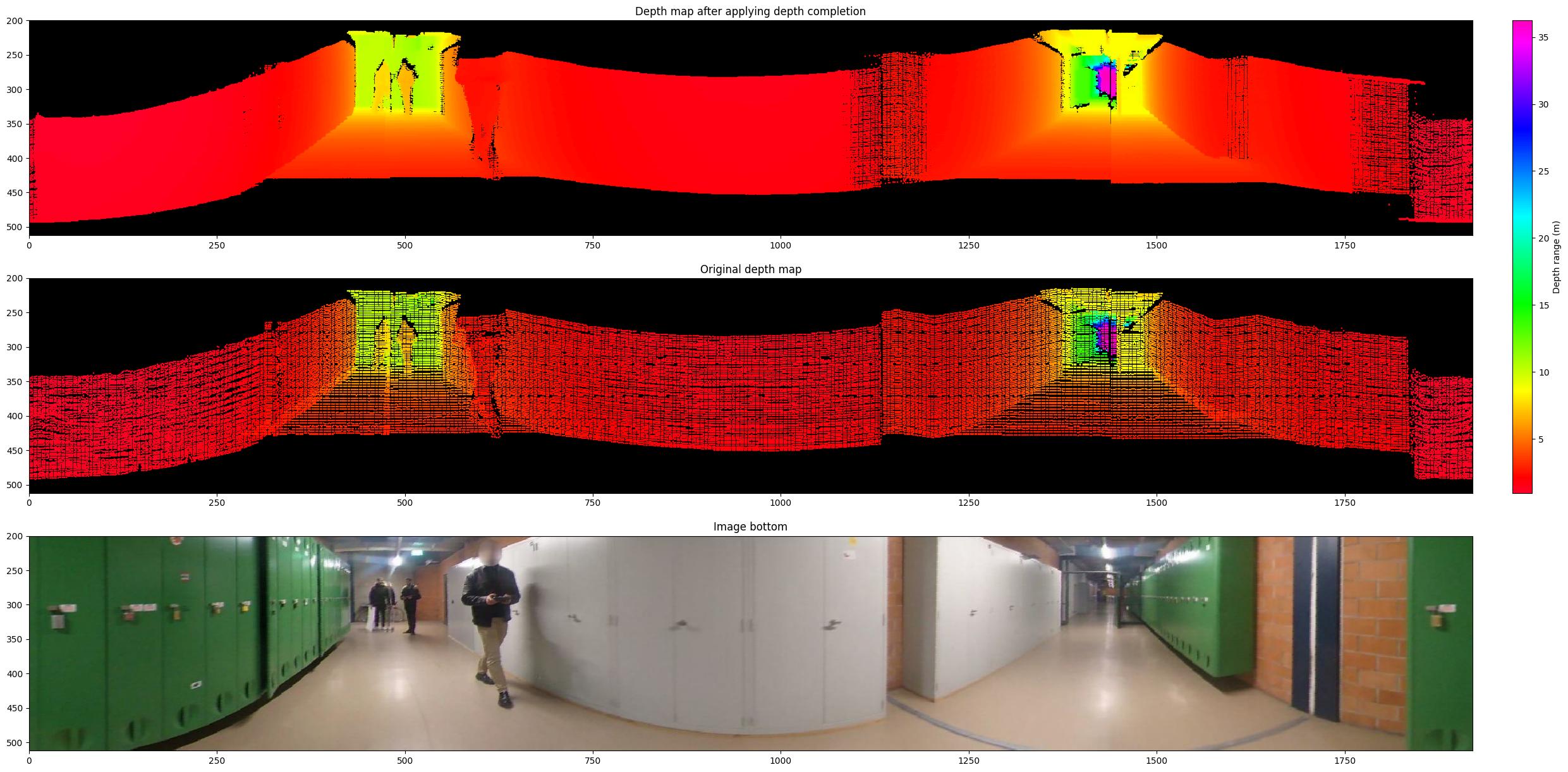}
    \caption{Depth completed depth map, original depth map and bottom image in complete view from an \textit{indoor} sequence.}
    \label{appx_fig:depth_completion_complete}
\end{figure*}%
\cref{appx_fig:depth_completion_detail} depicts a completed depth map and an original depth map together with the corresponding image of a patch from an \textit{outdoor} sequence.
In areas of homogeneous depth within the LiDAR's field of view, the completed depth map provides dense depth labels.
It is evident that the depth completion method does not provide labels for pixels at object boundaries. For instance, this can be observed at the street lamp located in the upper half of the image, around column 165, as well as at the boundaries of the tree in the upper half of the image, between columns 190 and 270.
This is an understandable limitation, as it is challenging to ascertain the precise locations of such boundaries based on depth data alone.
Even in the original depth map, it is evident that the boundaries of the tree are not clearly defined, and also some measurements at its boundaries appear to be erroneous.
Consequently, minor discrepancies may also be present in the measured point clouds.

The comprehensive representation of the \textit{indoor} scene in \cref{appx_fig:depth_completion_complete} substantiates the favorable visual impression of the depth completion when the original depth map exhibits no abrupt depth transitions.
Instead of providing labels with high errors in ambiguous regions, the original labels are retained due to the filtering of regions with high uncertainty.

Overall, the majority of pixels are labeled when applying depth completion, and no substantial errors are discernible visually.

%=================================================================================%
\section{Benchmark Specifications}
\label{appx_sec:benchmark}
%=================================================================================%

This section outlines the evaluation framework and benchmark specifications for assessing model performance on the \textsc{Helvipad} dataset. Furthermore, it details the architecture of 360-IGEV-Stereo adaptations.
%=--------------------------------------------------------------------------------%
\subsection{Evaluation Metrics}
\label{appx_subsec:metrics}
%=--------------------------------------------------------------------------------%

To assess the performance of models, we rely on several metrics, each providing insights into different aspects of the model's disparity and depth estimation accuracy. Given the sparse nature of our ground-truth data for disparity and depth, we apply a masking technique to evaluate models' predictions only in areas with available ground truth values. The metrics are computed by summing over all pixels for which ground truth is available. \\

\noindent
More formally, let us define $\mathcal{I}$ as the set of test set images and $p_{ij}$ a pixel $j$ within. We denote the depth and disparity ground truth values for a pixel $j$ in image $i \in \mathcal{I}$ as $r_{ij}$ and $d_{ij}$ respectively. Similarly, the corresponding values of this pixel $j$ in image $i$ predicted by the model are denoted respectively as $\hat{r}_{ij}$ and $\hat{d}_{ij}$. Among all pixels of the image $i$, we denote $\mathcal{A}_i$ the subset of pixels with available ground truth values in the image. \\

\begin{itemize}
\item \textbf{Mean Absolute Error (MAE):} MAE measure the average magnitude of errors between the predicted and actual disparity in degrees (and depth in meters), offering a direct assessment of overall error. For disparity and depth respectively, the MAE is defined as:
\begin{equation}
    \text{MAE} = \frac{1}{|\mathcal{I}|}\sum_{i\in \mathcal{I}} \frac{1}{|\mathcal{A}_i|} \sum_{j \in \mathcal{A}_i} \left\vert y_{ij} - \hat{y}_{ij}\right\vert,
\end{equation}
where $y$ can represent either the depth ($r$) or disparity ($d$) values. \\

\item \textbf{Root Mean Square Error (RMSE):} RMSE measures the square root of the average squared differences, emphasizing larger errors. It is defined as:
\begin{equation}
    \text{RMSE} = \frac{1}{|\mathcal{I}|}\sum_{i\in \mathcal{I}} \sqrt{\frac{1}{|\mathcal{A}_i|} \sum_{j \in \mathcal{A}_i} \vert\vert y_{ij} - \hat{y}_{ij} \vert\vert^2},
\end{equation}
where $y$ can represent either the depth ($r$) or disparity ($d$) values. \\

\item \textbf{Mean Absolute Relative Error (MARE):} Considering the varying range of disparity (and depth) values, MARE is crucial. The metrics normalizes the error against the actual depth values, offering a nuanced measure of accuracy. The MARE is defined as:
\begin{equation}
    \text{MARE} = \frac{1}{|\mathcal{I}|}\sum_{i\in \mathcal{I}} \frac{1}{|\mathcal{A}_i|} \sum_{j \in \mathcal{A}_i} \left\vert\frac{y_{ij} - \hat{y}_{ij}}{y_{ij}}\right\vert,
\end{equation}
where $y$ can represent either the depth ($r$) or disparity ($d$) values.

\item \textbf{Left-Right Consistency Error (LRCE):} This metrics evaluates the consistency at the left and right boundaries of 360° images by measuring the discrepancy between predicted depth values across the image edges. In the original work introducing LRCE~\citep{shen2022panoformer}, the metrics also accounts for left-right discrepancies in ground-truth data to address extreme cases where object edges align exactly with the image boundaries. Due to the sparsity of the LiDAR depth maps in the test set, there are very few valid points simultaneously at both image edges for computing LRCE metric (3 pixel pairs per image in average). As an alternative, we use the depth-completed tests maps (136 pixels in average) and compute LRCE with this augmented ground-truth. Given $\mathcal{B}_i$ the subset of valid pixel pairs in image $i$ where ground-truth labels exist for both the leftmost and rightmost columns ($\mathcal{B}_i \subset \mathcal{A}_i$), LRCE is defined as the sum of absolute differences between left and right edges for both predicted and ground-truth disparity values:
\begin{equation}
    \text{LRCE} = \frac{1}{|\mathcal{I}|} \sum_{i \in \mathcal{I}} \frac{1}{|\mathcal{B}i|} \sum_{j \in \mathcal{B}i} \left| e^{\text{gt}}_{i,j} - e^{\text{pred}}_{i,j} \right|,
\end{equation}
where $e_{i,j} = \left| d_{\text{left},i,j} - d_{\text{right},i,j} \right|$ is the left-right discrepancy term in image $i$ for pixel pair $j$, computed for predicted disparity error ($e^{\text{pred}}_{i,j}$) and ground-truth disparity error ($e^{\text{gt}}_{i,j}$).

\end{itemize}

%=--------------------------------------------------------------------------------%
\subsection{Implementation Details}
\label{appx_subsec:implementation}
%=--------------------------------------------------------------------------------%

In the following, we elaborate on the implementation and training details of each model included in the benchmark. All our experiments are conducted using Nvidia A100-SXM4-80GB GPUs.

\paragraph{PSMNet.} Despite its age, PSMNet is a robust and popular method for conventional stereo depth estimation that we included in our benchmark. Our implementation is based on the code provided by the authors\footnote{\url{https://github.com/JiaRenChang/PSMNet}}. We initialize our model with weights from a SceneFlow-pretrained model and fine-tune it on our dataset for 24 epochs. The model is trained with a batch size of 20 images, with an Adam optimizer, an initial learning rate of 0.0001, and no weight decay.

\paragraph{360SD-Net.} The model is trained from scratch for 40 epochs using an Adam optimizer with an initial learning rate of 0.001, no weight decay, and a batch size of 16. It undergoes further fine-tuning for 10 epochs at a reduced learning rate of 0.0001 to enhance performance. Our implementation is based on the code provided by the authors\footnote{\url{https://github.com/albert100121/360SD-Net}}.

\paragraph{IGEV-Stereo.} Again, our implementation is based on the code provided by the authors\footnote{\url{https://github.com/gangweiX/IGEV/tree/main/IGEV-Stereo}}. We initialize the model from SceneFlow-pretrained weights and fine-tune it on our dataset employing an AdamW optimizer and a one-cycle learning rate schedule with a maximum learning rate of $3e^{-5}$, alongside a weight decay of $1e^{-5}$. The training spans 200k steps with a batch size of 16, equivalent to approximately 92 epochs.

\paragraph{360-IGEV-Stereo.} We adapt the IGEV-Stereo code to include our architecture modification stated in the paper and further detailed in \cref{appx_subsec:architecture}.
The implementation details of 360-IGEV-Stereo are similar to IGEV-Stereo, with some small modifications.

We convert the disparity, given in degree, to pixels to be able to warp the top image appropriately for constructing the cost volume:
\begin{equation}
    d_\text{pix} = \frac{960~\text{px} \times d_\text{deg}}{180^\circ}.
\end{equation}
Hereby, $960~\text{px}$ is the height of the downsampled image before cropping.

During training 360-IGEV-Stereo had some instabilities. To mitigate this issue, we clamp the disparity $d_\text{deg}$ in each step to $d_\text{deg} \in [d_\text{deg, min}, d_\text{deg, max}]$.
According to the statistics of the dataset, the minimum and maximum disparity are set to $d_\text{deg, min} = 0.048^\circ$ and $d_\text{deg, max} = 23^\circ$.

To enable a better understanding of the context, we use the the full image size of 512 x 1920 for training.
Except of a common photometric data augmentation, we do not apply any data augmentations.

All weights that have not been modified are initialized with the original IGEV-Stereo weights created with pretraining on Sceneflow by the authors.
The model is trained with a batch size of 4 for 20 epochs which corresponds to around 130k steps.
Furthermore, the maximum disparity for constructing the cost volumes is set to 128 which is the smallest number that is divisible by 32 and larger than the maximum disparity in pixels.

%=--------------------------------------------------------------------------------%
\subsection{360-IGEV-Stereo Architecture}
\label{appx_subsec:architecture}
%=--------------------------------------------------------------------------------%

\begin{table}[htbp]
    \centering
    \setlength{\tabcolsep}{5pt}
    \resizebox{\columnwidth}{!}{%
    \begin{tabular}{lcccc} 
        \toprule
        \textbf{Layer} & \multicolumn{2}{c}{~~~~\textbf{Channels}} & \textbf{Scaling} & \textbf{Input} \\ 
        & in & out &  &  \\ 
        \midrule
        \multicolumn{5}{c}{\textit{1. Image feature extractor}} \\
        \midrule
        \texttt{img\_conv} & 3 & 32 &  1/2 &  top / bottom image \\    
        \texttt{img\_bottleneck1} & 32 & 16 & 1 & \texttt{img\_conv} \\
        \texttt{img\_bottleneck2} & 16 & 24 & 1/2 & \texttt{img\_bottleneck1} \\
        \texttt{img\_bottleneck3} & 24 & 32 & 1/2 & \texttt{img\_bottleneck2} \\
        \texttt{img\_bottleneck4} & 32 & 96 & 1/2 & \texttt{img\_bottleneck3} \\
        \texttt{img\_bottleneck5} & 96 & 160 & 1/2 & \texttt{img\_bottleneck4} \\
        \midrule
        \multicolumn{5}{c}{\textit{2. Feature concatenation}} \\
        \midrule
        \texttt{concat\_img\_pm} & (160+32) & 192 & 1 &  (\texttt{img\_bottleneck4}, \texttt{pm\_bottleneck4}) \\
        \texttt{concat\_conv} & 192 & 160 & 1 &  \texttt{concat\_img\_pm} \\
        \midrule
        \multicolumn{5}{c}{\textit{3. Upsampling layers}} \\
        \midrule
        \texttt{up\_bottleneck6} & 160 & 192 & 2 & (\texttt{concat\_conv}, \texttt{img\_bottleneck4}) \\
        \texttt{up\_bottleneck7} & 192 & 64 & 2 & (\texttt{up\_bottleneck6}, \texttt{img\_bottleneck3}) \\
        \texttt{up\_bottleneck8} & 64 & 48 & 2 & (\texttt{up\_bottleneck7}, \texttt{img\_bottleneck2}) \\
        \texttt{final\_conv3x3} & 48 & 48 & 1 & \texttt{up\_bottleneck8} \\
        \midrule
        \multicolumn{5}{c}{\textit{4. Stem part}} \\
        \midrule
        \texttt{stem2pre} & 3 & 32 &  1/2 &  top / bottom image \\
        \texttt{stemconcat} & (32+32) & 64 &  1 & (\texttt{stem2pre}, \texttt{pm\_coord2}) \\
        \texttt{stem2post} & 64 & 32 & 1 & \texttt{stemconcat} \\
        \texttt{stem4} & 32 & 48 & 1/2 & \texttt{stem2post} \\
        \bottomrule
    \end{tabular}
    }
    \caption{\textbf{Architecture of 360-IGEV-Stereo's feature network}.
    The steps 1 to 3 are part of the main feature network whose features at the scales 1/4, 1/8, 1/16, and 1/32 are used to build the CGEV.
    Its features are concatenated with the encoded polar map at its bottleneck in 1/32 of the original image size.
    The orange part in the bottom of the feature network in \cref{fig:architecture_360_igev_stereo} is called stem.
    At scale 1/4 it is used for the construction of the CGEV and at scale 1/2 the feature map obtained for the bottom image is used for spatial upsampling.
    The encoded polar map is concatenated with stem at 1/2 of the original image size.
    }
    \label{appx_tab:360igev_feature_network}
\end{table}

\begin{table}[htbp]
    \centering
    \setlength{\tabcolsep}{5pt}
    \resizebox{\columnwidth}{!}{%
    \begin{tabular}{lcccc} 
        \toprule
        \textbf{Layer} & \multicolumn{2}{c}{~~~~\textbf{Channels}} & \textbf{Scaling} & \textbf{Input} \\ 
        & in & out &  &  \\ 
        \midrule
        \multicolumn{5}{c}{\textit{1. Image feature extractor}} \\
        \midrule
        \texttt{img\_conv7x7} & 3 & 64 &  1 &  bottom image \\    
        \texttt{img\_resblock1} & 64 & 64 & 1 & \texttt{img\_conv7x7} \\
        \texttt{img\_resblock2} & 64 & 96 & 1/2 & \texttt{img\_resblock1} \\
        \texttt{img\_resblock3} & 96 & 128 & 1/2 & \texttt{img\_resblock2} \\
        \midrule
        \multicolumn{5}{c}{\textit{2. Feature concatenation}} \\
        \midrule
        \texttt{concat\_img\_pm} & (128+32) & 160 & 1 &  (\texttt{img\_resblock3}, \texttt{pm\_coord4}) \\
        \texttt{concat\_conv} & 160 & 128 & 1 &  \texttt{concat\_img\_pm} \\
        \midrule
        \multicolumn{5}{c}{\textit{3. Multi-scale outputs}} \\
        \midrule
        \texttt{output04\_conv} & 128 & 128 & 1 & \texttt{concat\_conv} \\
        \texttt{output04\_resblock4} & 128 & 128 & 1/2 & \texttt{concat\_conv} \\
        \texttt{output08\_conv} & 128 & 128 & 1 & \texttt{output04\_resblock4} \\
        \texttt{output08\_resblock5} & 128 & 128 & 1/2 & \texttt{output04\_resblock4} \\
        \texttt{output16\_conv} & 128 & 128 & 1 & \texttt{output04\_resblock5} \\
        \bottomrule
    \end{tabular}
    }
    \caption{\textbf{Architecture of 360-IGEV-Stereo's context network}.
    The features of the image are concatenated with the encoded polar map at 1/4 of the original image size.
    Context features at the scales 1/4, 1/8, and 1/16 are used by the ConvGRU block.}
    \label{appx_tab:360igev_context_network}
\end{table}

\begin{table}[htbp]
    \centering
    \setlength{\tabcolsep}{5pt}
    \resizebox{\columnwidth}{!}{%
    \begin{tabular}{lcccc} 
        \toprule
        \textbf{Layer} & \multicolumn{2}{c}{~~~~\textbf{Channels}} & \textbf{Scaling} & \textbf{Input} \\ 
        & in & out &  &  \\
        \midrule
        \texttt{pm\_coord2} & 1 & 32 & 1/2 &  polar map \\    
        \texttt{pm\_coord4} & 32 & 32 & 1/2 & \texttt{pm\_coord2} \\    
        \texttt{pm\_coord8} & 32 & 32 & 1/2 & \texttt{pm\_coord4} \\    
        \texttt{pm\_coord16} & 32 & 32 & 1/2 & \texttt{pm\_coord8} \\    
        \texttt{pm\_coord32} & 32 & 32 & 1/2 & \texttt{pm\_coord16} \\
        \bottomrule
    \end{tabular}
    }
    \caption{\textbf{Architecture of 360-IGEV-Stereo's polar encoder}.
    The polar encoder's at 1/2, 1/4, and 1/32 of the original polar map size are used.
    All layers are convolutional with a kernel size of 3, a stride of 2.
    After each layer batch normalization and the Leaky ReLU activation function are applied.}
    \label{appx_tab:360igev_polar_encoder}
\end{table}

To construct 360-IGEV-Stereo, we introduce three key enhancements to the IGEV-Stereo architecture.

Firstly, the polar map is added as an additional input to the network.
It has the same size as the input image and consists of repeated columns within a range of $[48^\circ, 144^\circ]$, corresponding to the vertical field of view of the input image.
Its encoder is shared between feature and context network.
The encoder contains convolutional layers with a stride of 2 to decrease the polar map size gradually.
For fusing the encoded polar map with the feature we concatenate the features at the lowest possible resolution before producing multi-scale outputs, followed by a convolution that recreates the number of channels.
The overall architecture is outlined more in detail in \cref{appx_tab:360igev_feature_network}, \cref{appx_tab:360igev_context_network} and \cref{appx_tab:360igev_polar_encoder}.

Secondly, we build the cost volume based on vertical instead of horizontal warping.
This means that in the construction of the geometry encoding volume the group-wise correlation volume is calculated by shifting the top image about the corresponding disparity index vertically.
Similarly, for building the all-pairs correlation volume the top image is warped downwards according to the disparity index.

Lastly, we apply circular padding at evaluation time.
Assuming the original image $I$ has height $H$ and width $W$, the value of the pixel with the row index $i$ and the column index $j$ of the circular padded image $I^\text{cp}$ can be calculated in dependence of the amount of padding $P$ with the following formula:
\begin{equation}
    I^\text{cp}_{i,j} = 
        \begin{cases}
            I_{i, j + W - P} & \text{if } 0 \leq j < P \\
            I_{i, j - P} & \text{if } P \leq j < W + P \\
            I_{i, j - W - P} & \text{if } W + P \leq j < W + 2P
        \end{cases}
\end{equation}
Circular padding is omitted during training to reduce computations and enable larger batch sizes.

%=================================================================================%
\section{Additional Results}
\label{appx_sec:add_results}
%=================================================================================%

In this section, we further study the impact of pretrained weight initialization and cross-dataset generalization.

%=--------------------------------------------------------------------------------%
\subsection{Effect of Pretrained Weights}
\label{appx_subsec:pretraining}
%=--------------------------------------------------------------------------------%

In addition to the ablative studies detailed in the main paper, we report in \cref{appx_tab:pretraining} a detailed comparison of each model performances using randomly initialization versus fine-tuning from pretrained weights. 

Both 360-IGEV-Stereo and IGEV-Stereo show significant improvements when initialized with Scene Flow pretrained weights, outperforming their randomly initialized counterparts across all depth and disparity metrics. For example, 360-IGEV-Stereo achieves a reduced depth MAE of 1.77m and RMSE of 4.36m, demonstrating the effectiveness of leveraging models trained on standard images like Scene Flow for omnidirectional image training. Surprisingly, this pattern does not hold for PSMNet, where the model initialized randomly achieves better performance than using Scene Flow pretrained weights. In contrast, 360SD-Net shows minimal improvement with Stereo-MP3D pretraining, despite the dataset being omnidirectional. This discrepancy could be attributed to Stereo-MP3D’s limitation to indoor scenes, whereas \textsc{Helvipad} has a broader range of scenes.

%=--------------------------------------------------------------------------------%
\subsection{Left-Right Consistency}
\label{appx_subsec:LRCE}
%=--------------------------------------------------------------------------------%

In addition to the main results available in \cref{tab:comparative_results}, we provide a more detailed analysis of Left-Right Consistency Error (LRCE) across different scene types in \cref{appx_tab:LRCE_augmented}.

\begin{table}[h]
    \centering
    \setlength{\tabcolsep}{10pt}
    \resizebox{\linewidth}{!}{%
    	\begin{tabular}{lcccc}
    		\toprule
    		Model & All & Indoor & Outdoor & Night Outdoor \\
    		\midrule
    		PSMNet & 1.80 & 0.93 & 1.31 & 1.16 \\
            360SD-Net & 0.90 & 0.52 & 1.02 & 1.01  \\
            IGEV-Stereo & 1.20 & 0.79 & 1.21 & 1.55  \\
            360-IGEV-Stereo & $\bm{0.38}$ & $\bm{0.17}$ & $\bm{0.38}$ & $\bm{0.46}$ \\
    		\bottomrule
    	\end{tabular}
    }
    \label{appx_tab:LRCE_augmented}
    \caption{\textbf{Depth-LRCE with augmented ground-truth across different scene types.} Results are reported in meters.} 
\end{table}

The results indicate that left-right consistency is more challenging to maintain in outdoor scenes, with the highest errors observed in night outdoor conditions due to low-light environments and reduced texture details. In contrast, indoor scenes achieve the lowest LRCE, likely due to more structured environments with well-defined depth boundaries and fewer extreme lighting variations.

%=--------------------------------------------------------------------------------%
\subsection{Cross-Dataset Generalization}
\label{appx_subsec:cross-dataset}
%=--------------------------------------------------------------------------------%

We study the cross-dataset generalization of 360SD-Net by first training the model on the \textsc{Helvipad} dataset and then fine-tuning it on Stereo-MP3D and Stereo-SF3D datasets~\citep{wang20icra} , which share a similar top-bottom camera configuration. Results are available in \cref{appx_tab:cross-dataset}. Compared to a baseline trained from random initialization on Stereo-MP3D, fine-tuning on \textsc{Helvipad} significantly improved all depth and disparity metrics. On Stereo-MP3D, fine-tuning on \textsc{Helvipad} reduces depth MAE from 0.087m (random initialization) to 0.072m, along with consistent improvements in all disparity metrics. A similar trend is observed on Stereo-SF3D, where \textsc{Helvipad} pretraining improves depth MAE from 0.029m to 0.027m and disparity MAE from 0.105° to 0.099°, outperforming models pretrained on Stereo-MP3D. Note that although we used the authors' provided code\footnote{\url{https://github.com/albert100121/360SD-Net}} and the reported hyperparameters, our reproduced results differ from those reported in the original paper. We present our outcomes for a fair comparison. 

Fine-tuning on \textsc{Helvipad} offers a broader diversity of scenes, particularly outdoor environments. These results suggest that \textsc{Helvipad} not only captures a wider range of scenarios but also provides robust features that enhance generalization to datasets with overlapping characteristics.

\begin{table*}[t]
    \centering
    \setlength{\tabcolsep}{10pt}
    \resizebox{\textwidth}{!}{%
        \begin{tabular}{ll ccc ccc } 
            \toprule
            \textbf{Method} & \textbf{Initialization} & \multicolumn{3}{c}{\textbf{Disparity} (°)} & \multicolumn{3}{c}{\textbf{Depth} (m)} \\
            \cmidrule(lr){3-5} \cmidrule(lr){6-8}
            & & MAE $\downarrow$ & RMSE $\downarrow$ & MARE $\downarrow$ & MAE $\downarrow$ & RMSE $\downarrow$ & MARE $\downarrow$ \\
            \midrule
            \multirow{2}{*}{\centering PSMNet~\citep{chang2018pyramid}} & random & 0.29 & 0.50 & 0.25 & 2.51 & 5.67 & 0.18 \\
            & Scene Flow & 0.33 & 0.54 & 0.29 & 2.78 & 6.17 & 0.19 \\
            \midrule
            \multirow{2}{*}{\centering
            360SD-Net~\citep{wang20icra}} & random & 0.22 & 0.42 & 0.19 & 2.12 & 5.08 & 0.18 \\
            & Stereo-MP3D & 0.23 & 0.44 & 0.20 & 2.31 & 5.41 & 0.16 \\
            \midrule
            \multirow{2}{*}{\centering
            IGEV-Stereo~\citep{xu2023iterative}} & random & 0.23 & 0.44 & 0.18 & 2.10 & 5.30 & 0.17 \\
            & Scene Flow & 0.23 & 0.42 & 0.17 & 1.86 & 4.47 & 0.15 \\
            \midrule
            \multirow{2}{*}{\centering
            360-IGEV-Stereo} & random & 0.20 & 0.40 & 0.16 & 1.91 & 4.60 & 0.14 \\
            & Scene Flow & 0.19 & 0.40 & 0.15 & 1.72 & 4.30 & 0.13 \\
            \bottomrule
        \end{tabular}
    }
    \caption{Comparison of model performance with random initialization vs. fine-tuned from pretrained weights.}
    \label{appx_tab:pretraining}
\end{table*}

\begin{table*}[t]
    \centering
    \setlength{\tabcolsep}{10pt}
    \resizebox{\textwidth}{!}{%
        \begin{tabular}{ll ccc ccc }
            \toprule
            \textbf{Dataset} & \textbf{Initialization} & \multicolumn{3}{c}{\textbf{Disparity} (°)} & \multicolumn{3}{c}{\textbf{Depth} (m)} \\
            \cmidrule(lr){3-5} \cmidrule(lr){6-8}
            & & MAE $\downarrow$ & RMSE $\downarrow$ & MARE $\downarrow$ & MAE $\downarrow$ & RMSE $\downarrow$ & MARE $\downarrow$ \\
            \midrule
            \multirow{2}{*}{\centering
            Stereo-MP3D} & \textcolor{gray}{reported~\citep{wang20icra}} & \textcolor{gray}{0.145} & \textcolor{gray}{0.693} & \textcolor{gray}{--} & \textcolor{gray}{0.059} & \textcolor{gray}{0.218} & \textcolor{gray}{--} \\
            & random & 0.148 & 0.994 & 0.074 & 0.087 & 0.294 & 0.050 \\
            & \textsc{Helvipad} & $\bm{0.129}$ & $\bm{0.930}$ & $\bm{0.063}$ & $\bm{0.072}$ & $\bm{0.252}$ & $\bm{0.039}$ \\
            \midrule
            \multirow{4}{*}{\centering
            Stereo-SF3D} & \textcolor{gray}{reported~\citep{wang20icra}} & \textcolor{gray}{0.103} & \textcolor{gray}{0.369} & \textcolor{gray}{--} & \textcolor{gray}{0.003} & \textcolor{gray}{0.091} & \textcolor{gray}{--} \\
            & random & 0.105 & $\bm{0.468}$ & 0.020 & 0.029 & 0.071 & 0.016 \\
            & Stereo-MP3D & 0.121 & 0.505 & 0.023 & 0.035 & 0.079 & 0.019 \\
            & \textsc{Helvipad} & $\bm{0.099}$ & 0.469 & $\bm{0.018}$ & $\bm{0.027}$ & $\bm{0.069}$ & $\bm{0.015}$ \\
            %\midrule
            %\multirow{3}{*}{\centering
            %360-IGEV-Stereo} & random & 0.33 & 1.45 & 0.05 & 0.06 & 0.26 & 0.04 \\
            %& Scene Flow$^**$ & 0.33 & 1.37 & 0.05 & 0.06 & 0.24 & 0.04 \\
            %& \textsc{Helvipad} & 0.38 & 1.54 & 0.05 & 0.07 & 0.26 & 0.04 \\
            \bottomrule
        \end{tabular}
    }
    \caption{Cross-dataset generalization results by fine-tuning 360SD-Net~\citep{wang20icra}.}
    \label{appx_tab:cross-dataset}
\end{table*}

\textbf{Real-world representation.} While collected on a university campus, the dataset captures many common urban environments, such as parking lots, roads, underpasses, pedestrian squares, footpaths, corridors and crowded halls. To further demonstrate transferability, we present below the qualitative result of 360-IGEV-Stereo on a real-world image without labels from the 360SD-Net paper:

\begin{figure}[ht]
    \centering
    \includegraphics[width=0.48\textwidth]{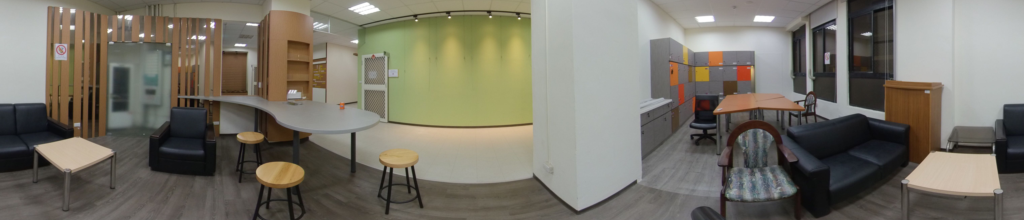}
    \includegraphics[width=0.48\textwidth]{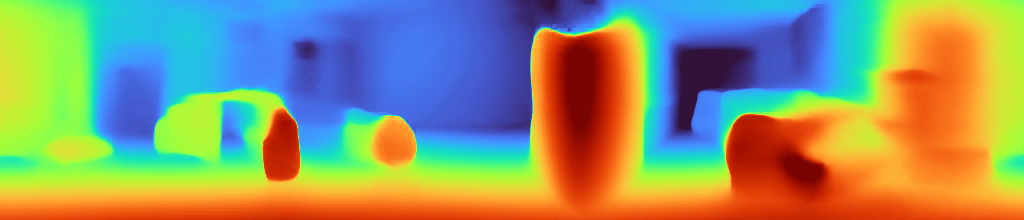}
    \caption{\textbf{Qualitative result of 360-IGEV-Stereo on a real-world image.} The top image shows the bottom image of the input, while the bottom image displays the predicted disparity map.}
    \label{fig:real_world_representation}
\end{figure}

Trained solely on \textsc{Helvipad}, the model demonstrates zero-shot capabilities in this environment.

\end{document}